%% file: main_tip.tex
\documentclass[lettersize,journal]{IEEEtran}
\usepackage{amsmath,amsfonts}
\usepackage{algorithmic}
\usepackage{algorithm}
\usepackage{array}
\usepackage{textcomp}
\usepackage{stfloats}
\usepackage{url}
\usepackage{verbatim}
\usepackage{graphicx}
\usepackage{cite}
\usepackage{color}
\usepackage{multirow}
\usepackage{subcaption}
\usepackage[font={small}]{caption}
\usepackage{booktabs, dcolumn}
\usepackage{pifont} 
\newcommand{\cmark}{\ding{51}}%
\newcommand{\xmark}{\ding{55}}%
\usepackage{adjustbox}
\usepackage{gensymb}

\makeatletter
\let\NAT@parse\undefined
\makeatother
\usepackage[numbers,sort,compress]{natbib}
\usepackage{hyperref}

\newcommand{\imgext}{jpg}

\begin{document}

\title{HIVE: HIerarchical Volume Encoding for Neural Implicit Surface Reconstruction}


\author{Xiaodong Gu$^1$*, Weihao Yuan$^1$*, Heng Li$^{1 2}$, Zilong Dong$^{1}$ and Ping Tan$^{1 2}$
\thanks{The authors are with $^1$ Alibaba Group, Hangzhou 310000, China and $^2$ Hong Kong University of Science and Technology, Hong Kong 999077, China.}
\thanks{*Equal contribution. Emails:
        {\tt\footnotesize dadong.gxd@alibaba-inc.com, qianmu.ywh@alibaba-inc.com}}
}




\maketitle




\input{includes/abstract}


\input{includes_v2/introduction}

\input{includes_v2/related_work.tex}

\input{includes_v2/method}

\input{includes_v2/experiments}
\input{includes_v2/conclusion}

{
\bibliographystyle{IEEEtranN}
\bibliography{reference}
}


\end{document}

%% file: includes/abstract.tex
\begin{abstract}
Neural implicit surface reconstruction has become a new trend in reconstructing a detailed 3D shape from images.
In previous methods, however, the 3D scene is only encoded by the MLPs which do not have an explicit 3D structure.
To better represent 3D shapes, we introduce a volume encoding to explicitly encode the spatial information.
We further design hierarchical volumes to encode the scene structures in multiple scales.
The high-resolution volumes capture the high-frequency geometry details since spatially varying features could be learned from different 3D points, while the low-resolution volumes enforce the spatial consistency to keep the shape smooth since adjacent locations possess the same low-resolution feature.
In addition, we adopt a sparse structure to reduce the memory consumption at high-resolution volumes, and two regularization terms to enhance results smoothness.
This hierarchical volume encoding could be appended to any implicit surface reconstruction method as a plug-and-play module, and can generate a smooth and clean reconstruction with more details.
Superior performance is demonstrated in DTU, EPFL, and BlendedMVS datasets with significant improvement on the standard metrics.

\end{abstract}

%% file: includes_v2/introduction.tex
\section{INTRODUCTION}

\IEEEPARstart{S}{urface} reconstruction from multi-view images is a heavily studied classic task in computer vision and is essential for numerous applications such as robotics, autonomous driving, augmented reality, and 3D modeling.
Given multiple images from different views of an object as well as the corresponding camera parameters, this task aims to recover the accurate 3D surface of the target object. Traditional methods~\citep{Fuhrmann2014} usually solve a depth map for each input image and then fuse~\citep{kazhdan2006poisson} those depth images to build a complete surface model. 
After the arising of deep networks, many methods try to exploit the neural networks to directly learn the mapping from 2D images to 3D surfaces~\citep{murez2020atlas,sun2021neuralrecon}.
These learning-based methods skip the intermediate depth map estimation and are highly efficient even on unseen objects and scenes. But they usually only recover coarse scale geometry and miss most of the high-frequency surface details.


Recently, many methods~\citep{idr, wang2021neus, yariv2021volsdf, darmon2022neuralwarp} have achieved highly accurate reconstruction results based on neural implicit surface. 
These methods represent the 3D shape with an implicit function, such as occupancy~\citep{oechsle2021unisurf} or signed distance fields (SDF)~\citep{wang2021neus}, and then leverage the neural radiance field (NeRF)~\citep{mildenhall2020nerf} to render the implicit geometry into color images.
Thus, the difference between the rendered images and the input images could optimize the neural radiance field as well as the implicit geometry. Strong results have been demonstrated. 
However, in these methods, the 3D shape is implicitly encoded in the multi-layer perceptrons (MLPs).
Although MLPs are compact and memory efficient, they do not have an explicit 3D structure, which may cause difficulties in optimizing the target 3D shape as is observed in mesh and point-cloud processing~\citep{Chibane2020,Peng2020}.
Furthermore, it is also known~\citep{sun2021neuralrecon} that compact MLPs are hard to encode all the geometry details, such that the recovered surface is prone to be over-smooth.

To solve this problem, some methods employ feature volumes or feature hash tables to facilitate MLPs to encode the 3D space~\cite{fridovich2022plenoxels, sun2022DVGO, Chen2022TensoRF, mueller2022instant}, which could directly encode the geometry of each 3D position faithfully and unambiguously. 
However, there are some problems in existing methods.
For the volume encoding methods~\cite{fridovich2022plenoxels, sun2022DVGO, Chen2022TensoRF}, there is usually only one high-resolution volume in their frameworks, in which case each voxel is updated in isolation. Due to the high degree of freedom in optimizing one voxel, it is hard to maintain a globally coherent shape, as is shown in Figure~\ref{fig:vis_compare} (c). 
For the hash-table encoding methods~\cite{mueller2022instant}, there are hash collisions which could cause some geometry defects, as is shown in Figure~\ref{fig:vis_compare} (a) (b).
To address these, we introduce a hierarchical volume encoding to encode the 3D space.
This hierarchical structure has three advantages.
First, the high-resolution feature volume helps to reason high-frequency geometry details in the corresponding locations.
Secondly, the voxels in the low-resolution volumes contain the information of large space, which could enforce spatial consistency to keep the shape smooth.
Thirdly, this hierarchical structure allows us to use low-dimensional features in the high-resolution volume, which helps to reduce memory consumption.
To further reduce memory consumption, we sparsify high-resolution volumes with the preliminary surface reconstruction computed from low-resolution volumes. We only keep voxels nearby the surface of the preliminary results.  
Finally, we design two smoothness terms to facilitate the optimization to make the reconstructed shape clean. 

In the experiments, we demonstrate that by simply adding the proposed hierarchical volume encoding, the performance of different methods are all improved significantly.
Specifically, the error of NeuS~\citep{wang2021neus} is reduced by 25\% from $0.84$ mm to $0.63$ mm, the error of VolSDF~\citep{yariv2021volsdf} is reduced by 23\% from $0.86$ mm to $0.66$ mm, and the error of NeuralWarp~\citep{darmon2022neuralwarp} is reduced by 10\% from $0.68$ mm to $0.61$ mm on the DTU~\citep{jensen2014dtudataset} dataset.
More than that, the error of NeuralWarp is reduced by 31\% on the EPFL~\citep{strecha2008epfl} dataset with the ``full" metric. 
Figure~\ref{fig:vis normal} shows an example from the DTU dataset. The color coded normal map clearly demonstrates our method can significantly boost NeuS~\citep{wang2021neus} to recover more geometry details while keeping the surface smooth and clean. 

The main contributions of this work are summarized in the following:

$\bullet$ We propose a hierarchical volume encoding, which can significantly boost the performance of neural implicit surface reconstruction as a plug-and-play module.

$\bullet$ The hierarchical volume encoding is further improved by employing a sparse structure which reduces the memory consumption, and by enforcing two regularization terms that keep the reconstructed surface smooth and clean.

$\bullet$ We demonstrate superior performance in three datasets.

%% file: includes_v2/related_work.tex
\section{RELATED WORK}

\input{tab_fig/vis_normal}

\input{tab_fig/vis_compare.tex}

\textbf{Traditional multi-view surface reconstruction}
Traditional methods typically follow a long pipeline of structure-from-motion (i.e. camera calibration)~\citep{colmap_sfm,tip3fu2021fast, tip4zhang2022cvids}, depth map estimation~\citep{barnes2009patchmatch}, and multi-depth fusion~\citep{kazhdan2006poisson} to reconstruct surface models from images.
Many advanced geometric methods~\citep{Jiang2013,Lhuillier2005,furukawa2009accurate,galliani2015massively,colmap_mvs} have been developed to enhance these different steps in the last two decades. 

After the arising of the deep networks, almost all steps in the conventional pipeline have been revolutionized, including feature extraction and matching~\citep{detone2018superpoint, sarlin2020superglue,tip5migliorati2020learnable, Li_match}, structure-from-motion~\citep{ummenhofer2017demon,tang2018ba}, and depth map estimation \citep{yao2018mvsnet,gu2020cascade,yuan2022neural,tip2yang2023single,gu2023dro,tip1li2023nrmvsnet,yang_depth,chen_depth,Tong_depth}. 


To skip the intermediate depth map estimation, given multiple images with known camera poses, some methods try to directly regress a volumetric prediction end-to-end like an occupancy volume~\citep{ji2017surfacenet} or a TSDF volume~\citep{murez2020atlas,sun2021neuralrecon,yuan2022former3d}.
While these methods simplify the overall pipeline and can be generalized to unseen objects and scenes, they often only learn to recover coarse scale geometry and miss many surface details.



\textbf{Implicit surface reconstruction}
A neural radiance field (NeRF) encodes a 3D scene~\citep{mildenhall2020nerf} implicitly in a neural network. 
The network is optimized to match the rendered images to the input images such that it can generate high quality novel view synthesis results. 
However, since there is no constraint imposed on the 3D geometry, the surface extracted from the implicit network usually has some defects due to tuning of the density threshold~\citep{oechsle2021unisurf}.
To address this problem, some methods first represent the 3D shape with an implicit geometry network, like occupancy grid~\citep{niemeyer2020dvr, oechsle2021unisurf} or signed distance fields~\citep{idr, wang2021neus, yariv2021volsdf, darmon2022neuralwarp, sun2022neural3dinthewild, long2022sparseneus, fu2022geoneus, yu2022monosdf}, and then transfer the output of the geometry network to a density function, with which the radiance network could render color images.
In this way, the radiance network as well as the geometry network can be optimized together by matching the rendered and input images.
{
In these methods, the 3D geometry is directly encoded in the MLPs without any explicit 3D spatial information.
To facilitate MLPs, some methods propose to encode the 3D space with a single geometric volume~\cite{fridovich2022plenoxels, sun2022DVGO, Chen2022TensoRF, martel2021acorn} or hash tables~\cite{mueller2022instant}.
However, single-volume may cause noise due to the high degree of freedom in optimizing one voxel, and hash tables may cause hash collisions which leads to defects, as displayed in Figure~\ref{fig:vis_compare}.
In this paper, we introduce a hierarchical volume encoding to explicitly encode the 3D spatial information, thus our method can reconstruct more surface details while keeping the shape globally coherent.
A similar structure is proposed in \cite{takikawa2021neural} for the SDF encoding task, which sums up the features from a large-feature-channel octree, while we concatenate the features from multiple small-feature-channel volumes, in which case our method consumes less memory.
}



%% file: tab_fig/vis_normal.tex



\begin{figure*}[]
\vspace{-5mm}
\begin{center}

\begin{subfigure}{0.67\columnwidth}
  \centering
  \includegraphics[width=1\columnwidth, trim={11cm 7cm 7cm 0cm}, clip]{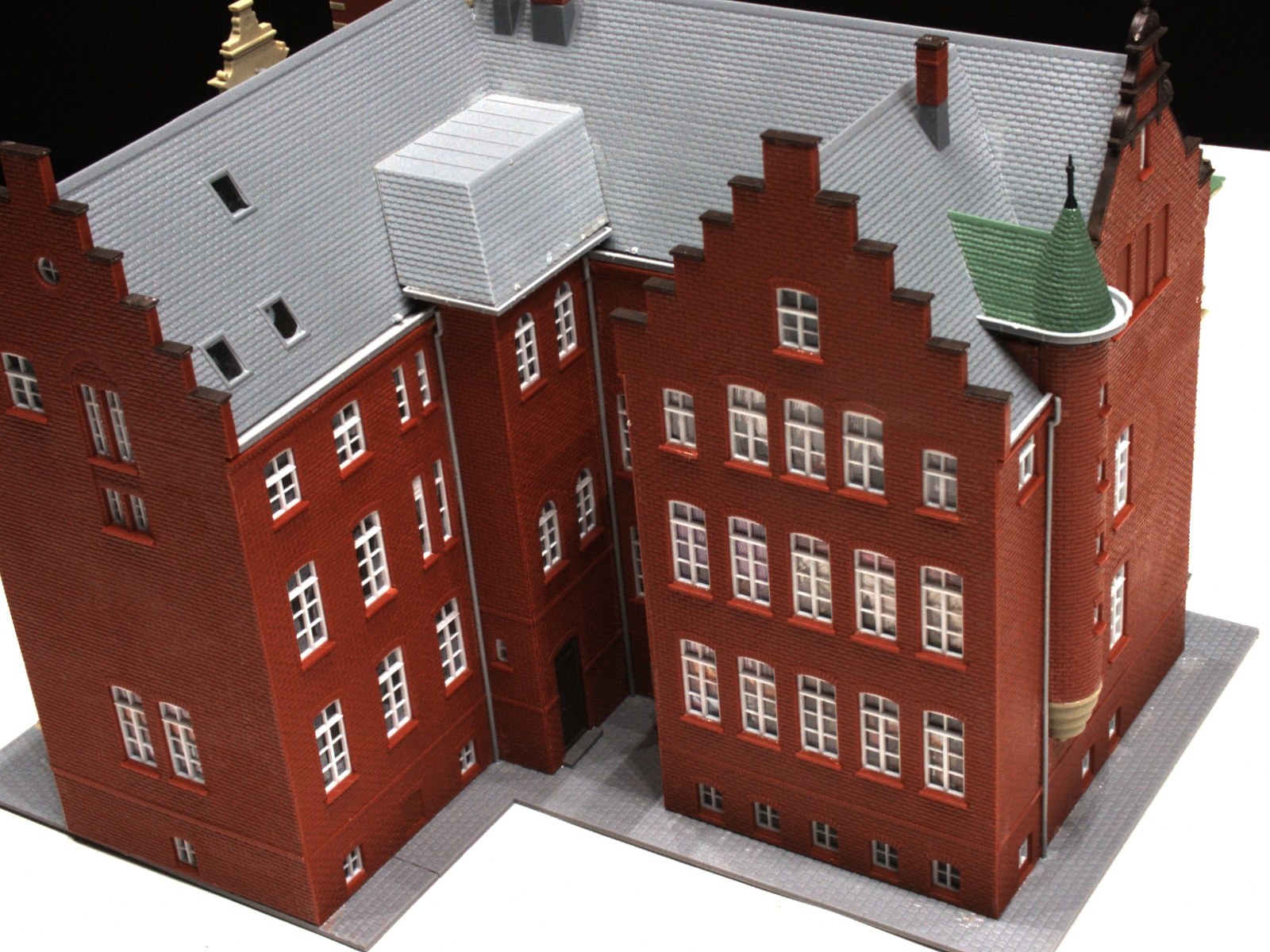}
  \caption*{Image}
\end{subfigure}
\begin{subfigure}{0.67\columnwidth}
  \centering
  \includegraphics[width=1\columnwidth, trim={11cm 7cm 7cm 0cm}, clip]{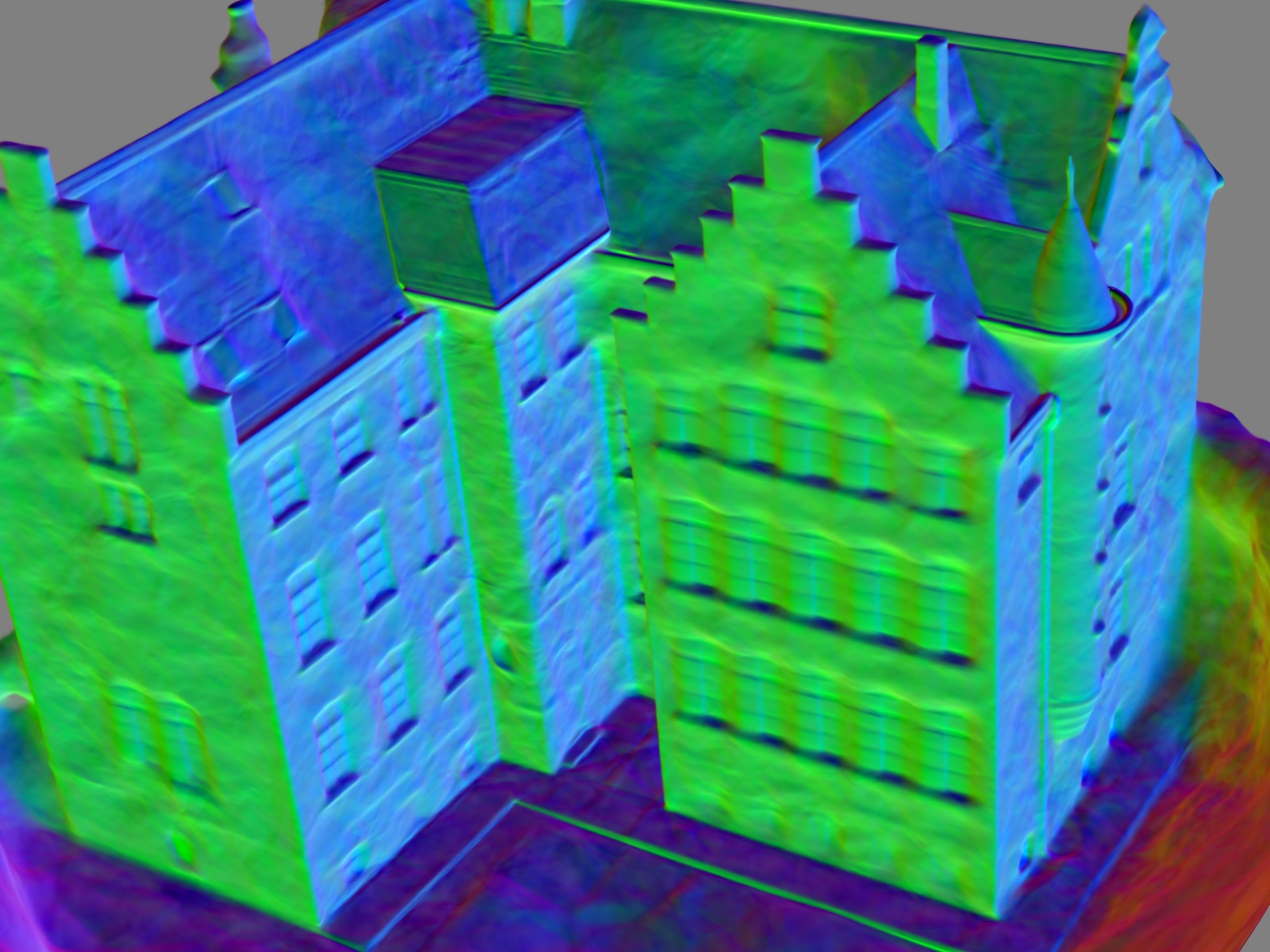}
  \caption*{NeuS}
\end{subfigure}
\begin{subfigure}{0.67\columnwidth}
  \centering
  \includegraphics[width=1\columnwidth, trim={11cm 7cm 7cm 0cm}, clip]{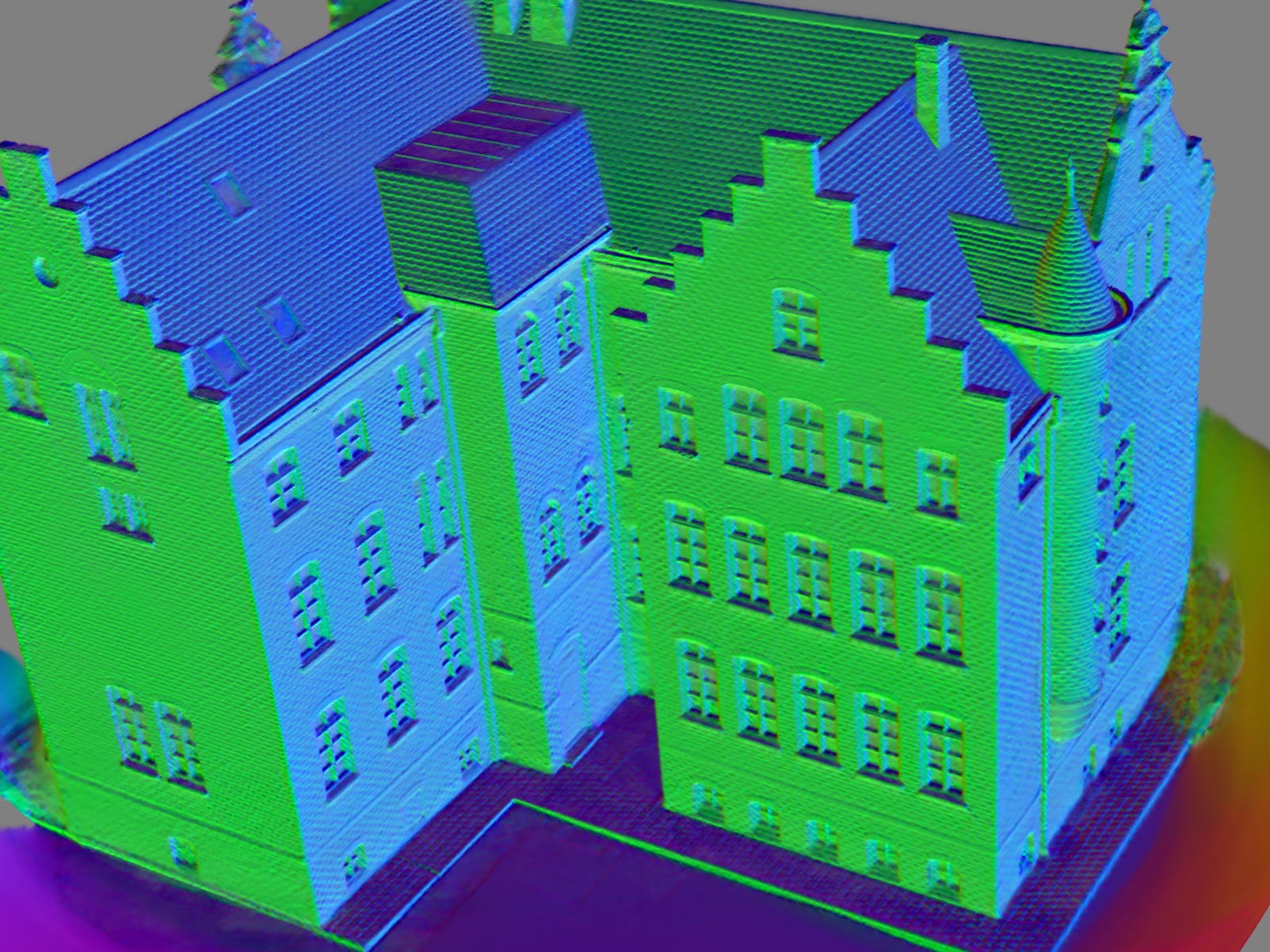}
  \caption*{NeuS+Ours}
\end{subfigure}

\end{center}
 \caption{Visualization of normal maps to highlight our advantages in recovering shape details.}
\label{fig:vis normal}
\end{figure*}

%% file: tab_fig/vis_compare.tex
\begin{figure*}[t]
\centering

\begin{subfigure}{0.49\columnwidth}
  \centering
  \includegraphics[width=1\columnwidth, trim={6cm 0cm 0cm 3cm}, clip]{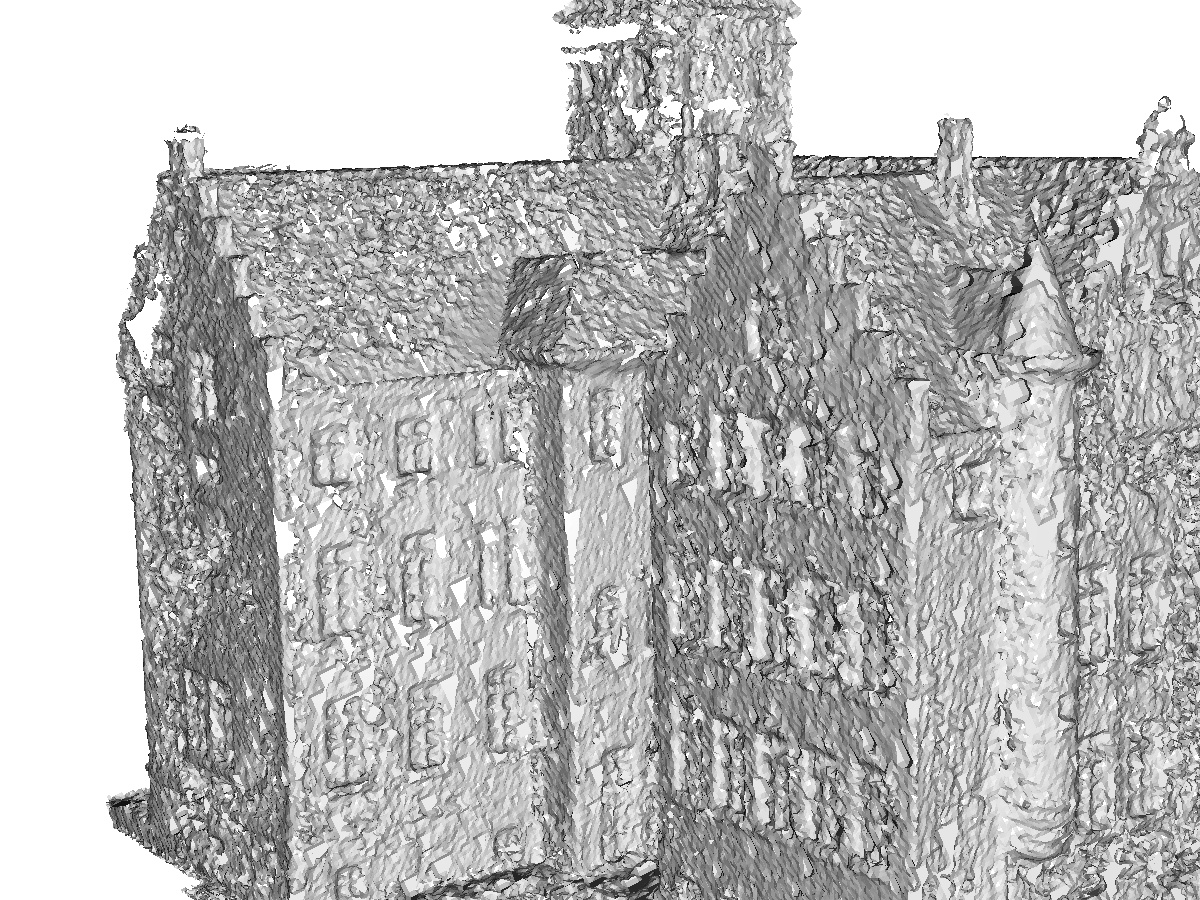}
  \caption*{(a) Instant-NGP}
\end{subfigure}
\begin{subfigure}{0.49\columnwidth}
  \centering
  \includegraphics[width=1\columnwidth, trim={6cm 0cm 0cm 3cm}, clip]{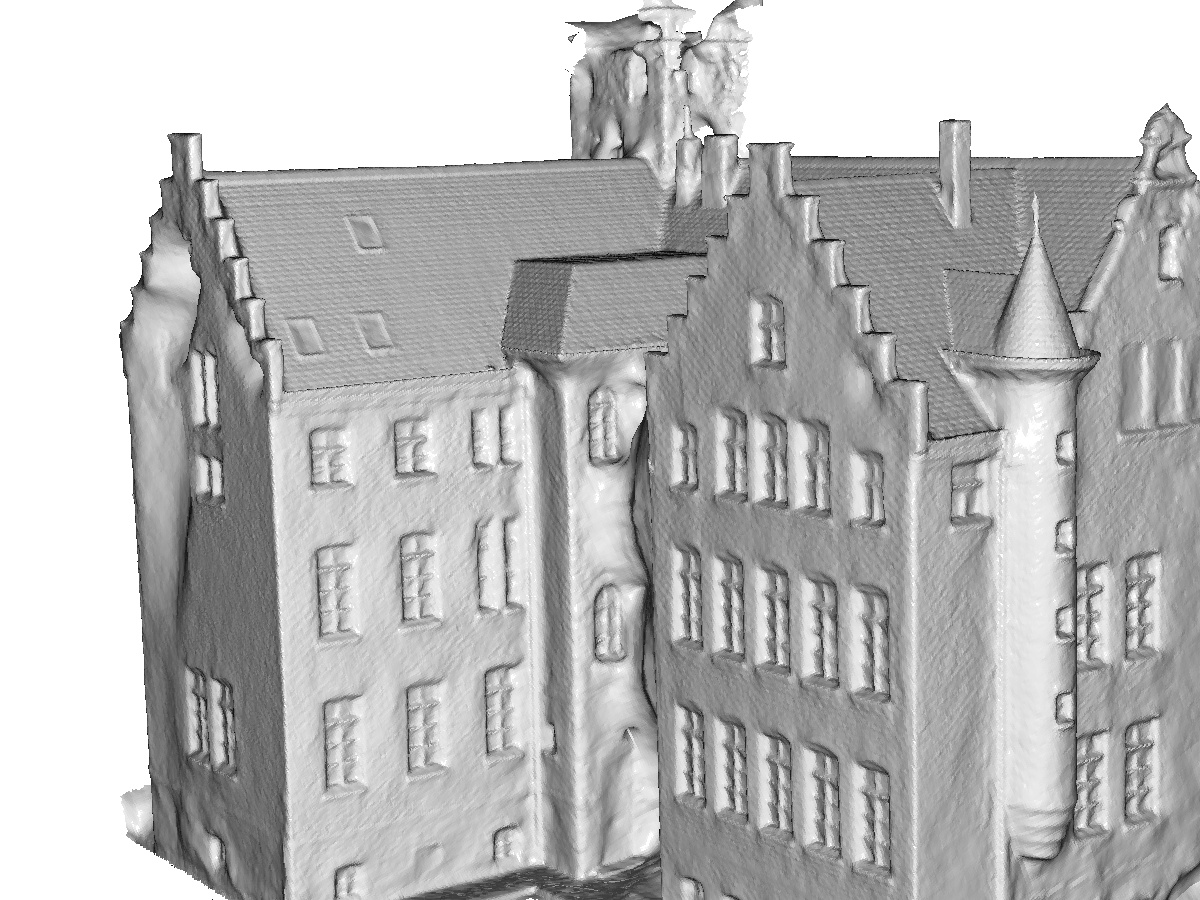}
  \caption*{(b) NeuS+Hash}
\end{subfigure}
\begin{subfigure}{0.49\columnwidth}
  \centering
  \includegraphics[width=1\columnwidth, trim={6cm 0cm 0cm 3cm}, clip]{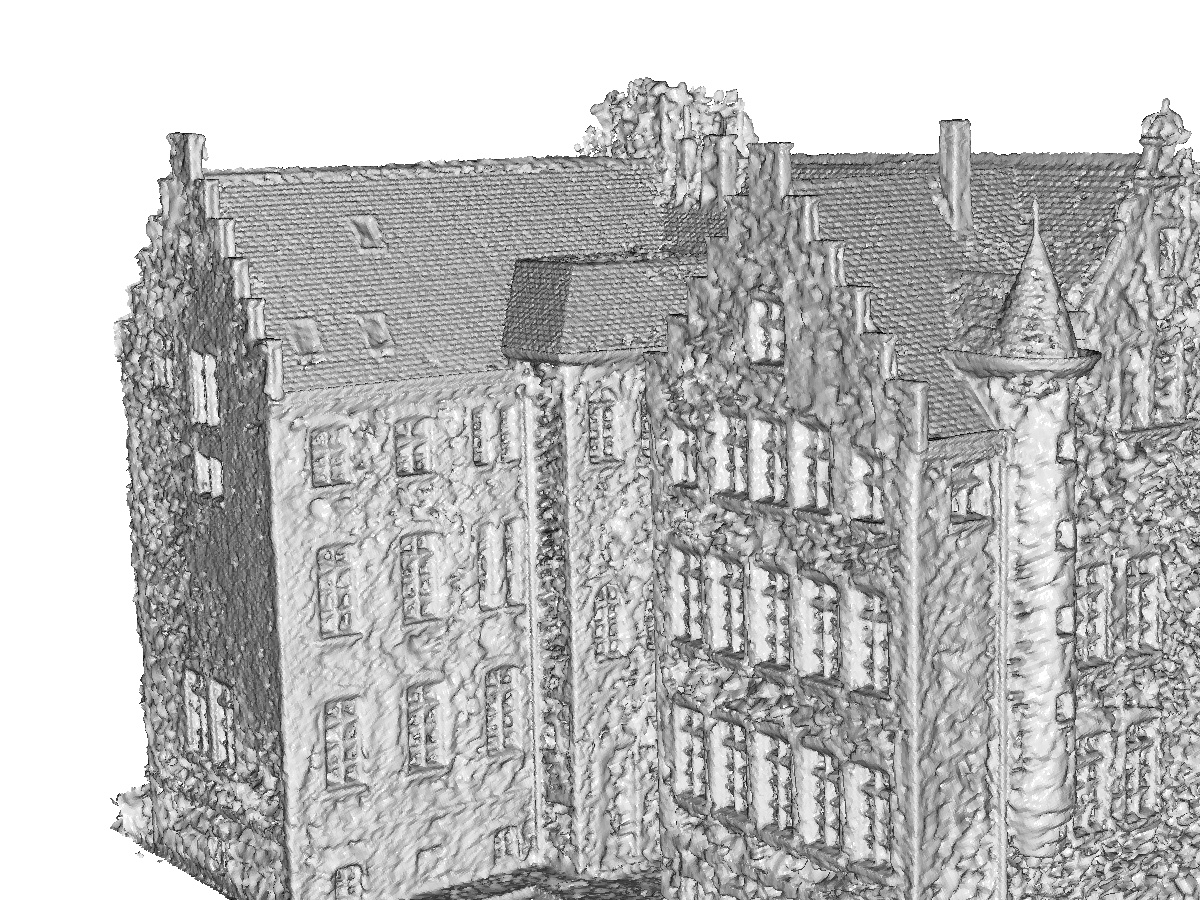}
  \caption*{(c) NeuS+Plenoxel}
\end{subfigure}
\begin{subfigure}{0.49\columnwidth}
  \centering
  \includegraphics[width=1\columnwidth, trim={6cm 0cm 0cm 3cm}, clip]{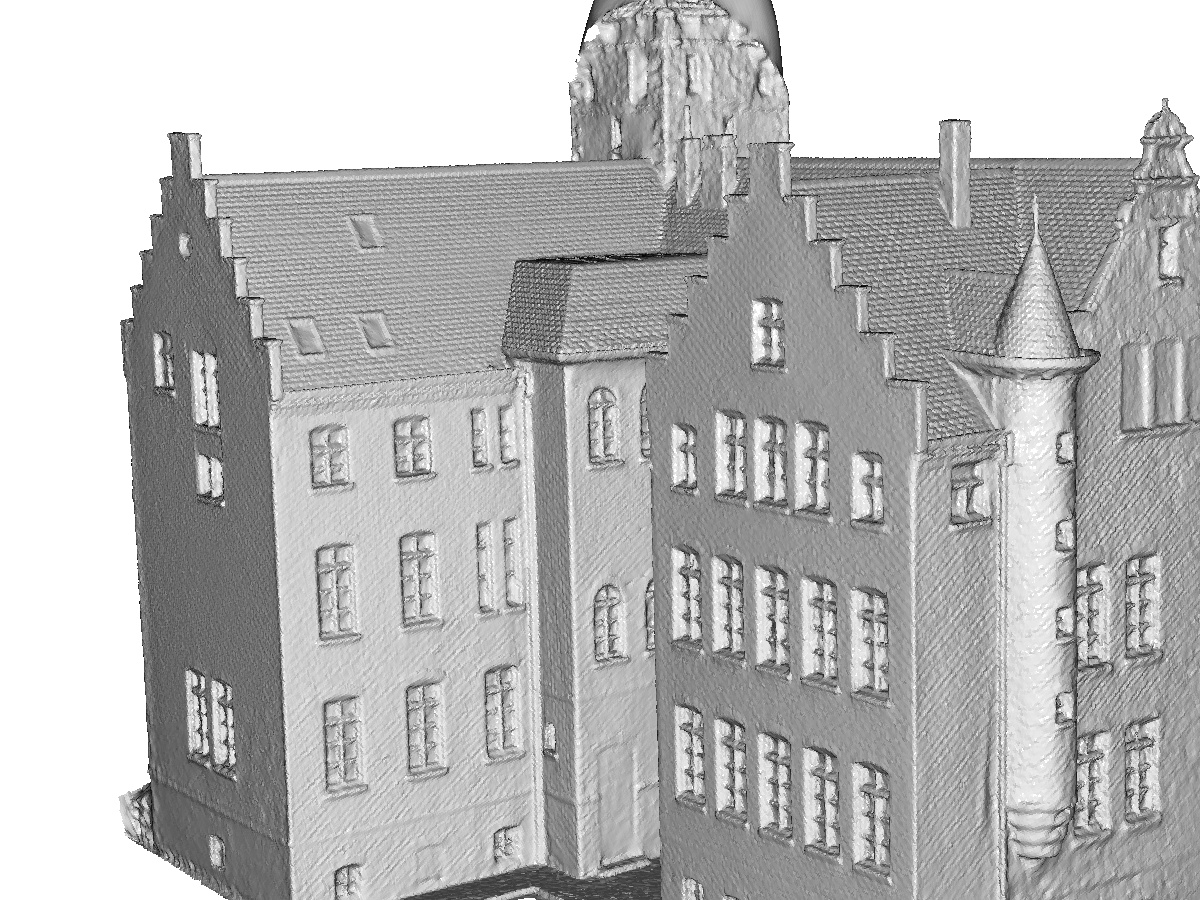}
  \caption*{(d) NeuS+Ours}
\end{subfigure}

 \caption{Visual comparison of different encoding on DTU scan-24.}
\label{fig:vis_compare}
\end{figure*}

%% file: includes_v2/method.tex
\section{METHODS}

\subsection{Overview}

\input{tab_fig/framework}

The overall framework of our method is illustrated in Figure~\ref{fig:pipeline}.
Given multiple images $\{\mathbf{I}_{i}\}_{i=1}^{N}$ of an object and corresponding camera poses, the task is to reconstruct the 3D surface of this object.
The 3D shape is represented by an implicit SDF network, with which another implicit color network renders images using the neural volume rendering.
To enhance the representation ability of the spatial information, we add a hierarchical volume encoding as the input of the SDF network, which can embed the 3D space in multiple scales.
To save the memory consumption, we also introduce a sparse structure for high-resolution volumes.
Finally, two implicit networks as well as the volume encoding are optimized by minimizing the difference between the rendered images and the input images as well as minimizing two regularization terms.

\subsection{Implicit Surface Reconstruction Based on Volume Rendering}

Neural implicit volume rendering is first introduced in \citep{mildenhall2020nerf} for novel view rendering and then used in \citep{wang2021neus, yariv2021volsdf, darmon2022neuralwarp} for surface reconstruction.
Our method could be added to any of them for better reconstruction.
Here we take NeuS~\citep{wang2021neus} as an example, which uses two multi-layer perceptrons (MLPs) to serve as two functions for representing an object: 
one is the SDF network $sdf: \mathbb{R}^3 \rightarrow \mathbb{R}$ that maps a spatial position $\mathbf{x}\in \mathbb{R}^3$ to its SDF value of the object surface, 
and the other one is the color network $c: \mathbb{R}^3 \times \mathbb{S}^2 \rightarrow \mathbb{R}$ that maps the spatial point $\mathbf{x}$ and a viewing direction $\mathbf{v}\in \mathbb{S}^2$ to a color value. 
The surface $\mathcal{S}$ is then represented by the zero-level set of the SDF function as:
\begin{equation}
    \mathcal{S} = \{\mathbf{x}\in \mathbb{R}^3 | {sdf}(\mathbf{x})=0\}.
\end{equation}

To optimize the SDF network, images are rendered from the color network as well as a weighting function computed from the SDF network.
The volume rendering function for generating colors is calculated as
\begin{equation}
    C(\mathbf{o}, \mathbf{v}) = \int_0^{+\infty}w(t)c(\mathbf{p}(t), \mathbf{v}) dt ,
\end{equation}
where $w$ is the weighting function computed by the density from the SDF network.

To make the weighting function unbiased and occlusion-aware, it is calculated as
\begin{equation}
    w(t) = T(t)\rho(t),
    \label{equ:w}
\end{equation}
where 
\begin{equation}
    T(t) = e^{-\int_0^t\rho(u)du}  ,
\end{equation}
and 
\begin{equation}
    \rho(t) = \max (\frac{-\frac{d\Phi_s}{dt}({sdf}(\mathbf{p}(t)))}{\Phi_s({sdf}(\mathbf{p}(t)))}, 0), \ \ 
    \Phi_s(x) = (1+e^{-sx})^{-1}    .
\end{equation}

\subsection{Hierarchical Volume Encoding}

\input{tab_fig/hierarchical_vol}

Previous methods usually encode all the information of a 3D object in the MLPs.
To assist the MLPs,
we build a 3D feature volume as the input of the MLPs, which explicitly encodes the 3D spatial information. 
This feature volume can naturally encode the knowledge about the 3D space of the object, while being optimized as well as the MLPs from the rendering loss.

To enhance the representation ability of the encoding, we also employ a hierarchical mechanism, where different scales of the 3D feature volumes are adopted, as is shown in Figure~\ref{fig:hierarchical_volume}.
In the experiments, we find a combination of multi-scale volumes works better than a single large-width volume.
This is reasonable since in low-resolution volumes, one voxel represents a large space, such that this space can have the same code.
The same code could smooth this space and prevent crushing 3D shapes, which is important in surface reconstruction.

In a basic setting, we employ eight feature volumes for the 3D space encoding, whose resolutions increase from $2 \times 2 \times 2 \times 4$ to $256 \times 256 \times 256 \times 4$.
When a point $\mathbf{x}$ is being rendered, eight features of the corresponding position from these eight volumes are trilinearly interpolated and concatenated to form  $\mathcal{F}(\mathbf{x})$, which works as the input of the MLPs.
Therefore, the SDF function becomes
\begin{equation}
    \mathcal{S} = \{\mathbf{x}\in \mathbb{R}^3 | {sdf}(\mathcal{F}(\mathbf{x}))=0\}.
\end{equation}

\subsection{Sparse High Resolution Volume}
\input{tab_fig/sparse_table}

\input{tab_fig/dtu_table}

\input{tab_fig/epfl_table}

It is evident that the higher the resolution of the volume is, the more details would be recovered, but a volume larger than $256 \times 256 \times 256$ would consume much more memory.
To solve this problem, a multi-stage optimization scheme is adopted with high-resolution sparse volumes.
In general, a lot of voxels in a dense volume are unoccupied and invalid.
Thus, we could save memory consumption a lot by using the shape reconstructed from the coarse volumes to cull voxels far from the surface in the high-resolution volumes.


To do this, we propose a multi-stage optimization framework. 
In the first stage, we use the above-mentioned basic structure, where the resolution of the largest volume is $256$.
After the first stage, we obtain a coarse surface reconstruction $\mathcal{S}_c$. 
In the second stage, we utilize $\mathcal{S}_c$ to cull unnecessary voxels to obtain the valid voxels $\mathbf{V}_h^{valid}$ in the high-resolution volume.
Specifically, we first dilate the surface $\mathcal{S}_c$ to ensure all the valid voxels nearing the surface are included.
Therefore, we only need to optimize the embeddings of $n$ valid voxels in $\mathbf{V}_h^{valid}$ instead of all voxels.
Here, we use a simple data structure, an embedding table $T_e$, to store these embeddings.


Given a floating-point three-dimensional position $\mathbf{x}$, we obtain the surrounding eight integer coordinates through the rounding operation, take out the embedding of these integer points, and then fuse them through trilinear interpolation to serve as the encoding of $\mathbf{x}$.
As shown in Figure~\ref{fig:sparse_table}, in order to efficiently extract the corresponding embedding from the embedding table $T_e$, we construct another index table $T_i$ which stores the indexes of $T_e$.
The values in table $T_i$ are all initialized to $-1$, which corresponds to the last embedding in table $T_e$.
The length of the embedding table $T_e$ is only $n+1$ ($n \ll N^3$), such that the memory consumption is reduced a lot due to the sparse structure of the high-resolution volumes. More details are given in the supplementary materials.


\subsection{Loss Function}
We equip three previous methods~\citep{wang2021neus, yariv2021volsdf, darmon2022neuralwarp} with our hierarchical volume encoding.
To optimize the model, we use the losses in their work without changes, i.e.
a rendering loss $\mathcal{L}_{color}$ which minimizes the difference between the rendered colors and input colors, and an Eikonal loss ${L_{eik}}$~\citep{icml2020eik} which encourages unit norm of  the SDF function gradients.
Besides, for NeuralWarp~\citep{darmon2022neuralwarp}, an additional warping color loss $\mathcal{L}_{warp}$ is adopted, which warps views on each other to enforce multi-view photo-consistency.

In addition, we add two additional regularization terms $\mathcal{L}_{tv}$ and $\mathcal{L}_{normal}$ to make the reconstructed surfaces smooth and clean.
The total variation (TV)~\citep{rudin1994tv} regularization is applied to each embedding volume to make adjacent voxels have similar characteristics, in which case the geometry could be more continuous and compact. It is computed as:
\begin{equation}
\begin{aligned}
    \mathcal{L}_{tv} = \sum_{m} \sum_{i,j,k} \sqrt{(V_{i+1,j,k} - V_{i,j,k})^2} + \\
    \sqrt{(V_{i,j+1,k} - V_{i,j,k})^2} + \sqrt{(V_{i,j,k+1} - V_{i,j,k})^2},
\end{aligned}
\end{equation}
where $m$ is the number of the hierarchical volumes.

Another regularization term $\mathcal{L}_{normal}$ is a smoothness constraint for normal.
For each pixel of the image, we calculate the accumulated normal gradients along the marching ray as 
\begin{equation}
    N_{grad}(\mathbf{o}, \mathbf{v}) = \int_0^{+\infty}w(t)n_{grad}(t) dt ,
\end{equation}
where $n_{grad}(t)$ is the gradient of the normal at $\mathbf{p}(t)$ and computed as 
\begin{equation}
    n_{grad}(t) = \nabla^2 sdf(\mathcal{F}(\mathbf{p}(t)).
    \label{equ:n_grad}
\end{equation}
The normal regularization $\mathcal{L}_{normal}$ is then defined as:
\begin{equation}
    \mathcal{L}_{normal} = \frac{1}{N_{pix}}\sum_{pix}||N_{grad}||_2,
\end{equation}
where $N_{pix}$ is the number of pixels in one optimization. 

Thus, the final loss $\mathcal{L}$ is defined as:
\begin{equation}
    \mathcal{L} = \mathcal{L}_{ori} + \lambda_{tv} * \mathcal{L}_{tv} + \lambda_{normal} * \mathcal{L}_{normal},
\end{equation}
where $\mathcal{L}_{ori}$ is the original loss of each method, $\lambda_{tv}$ and $\lambda_{normal}$ are two weighting factors.

%% file: tab_fig/framework.tex
\begin{figure*}[t]
\begin{center}
\includegraphics[width=2\columnwidth, trim={0cm 0cm 0cm 0cm}, clip]{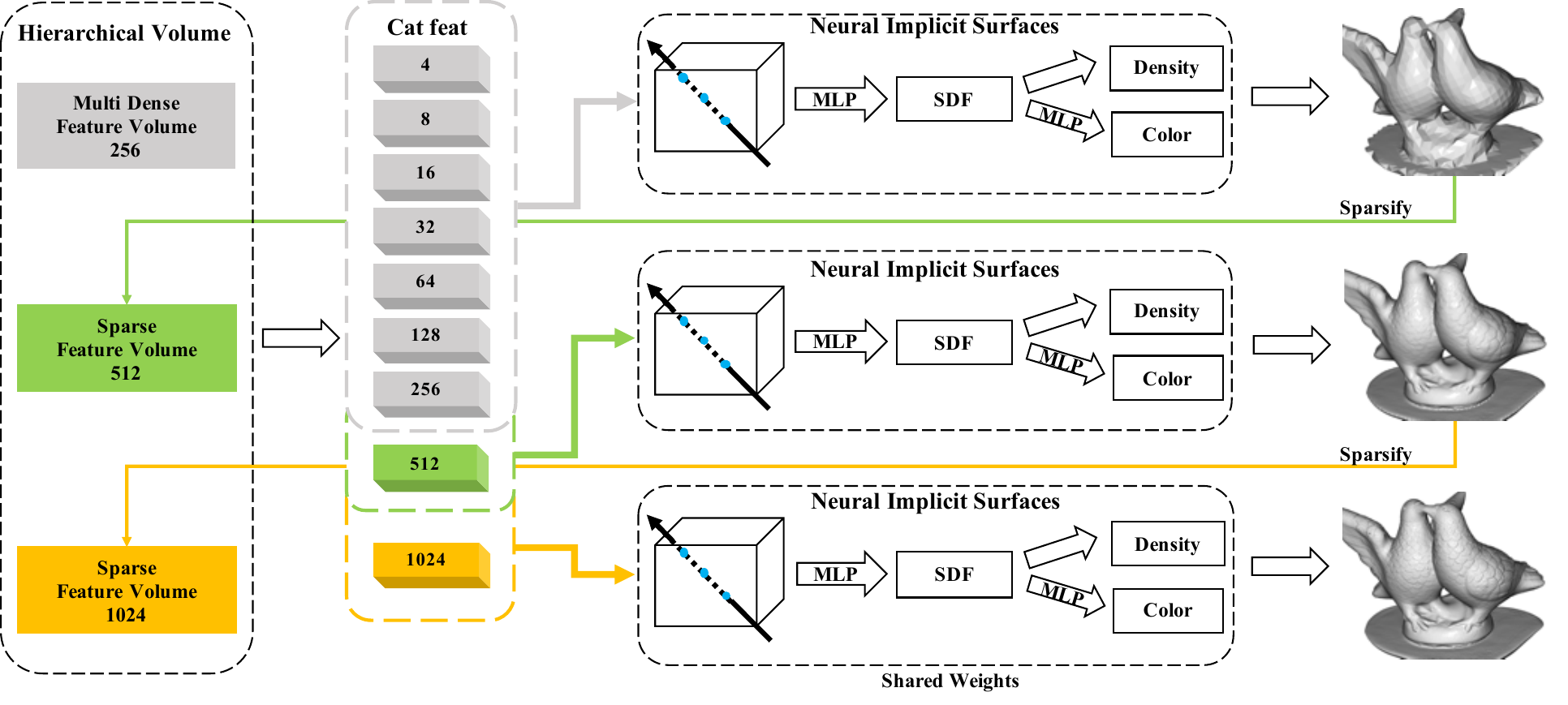}
\end{center}
 \caption{Method overview. In the first stage, we compute an initial result use features from volumes with resolution from $2$ to $256$. In a later stage, we finalize the result use features from sparsified high resolution volumes with a resolution of 512 or 1024.}
\label{fig:pipeline}
\end{figure*}

%% file: tab_fig/hierarchical_vol.tex
\begin{figure*}[t]
\begin{center}

\includegraphics[width=1.95\columnwidth, trim={0cm 0cm 0cm 0cm}, clip]{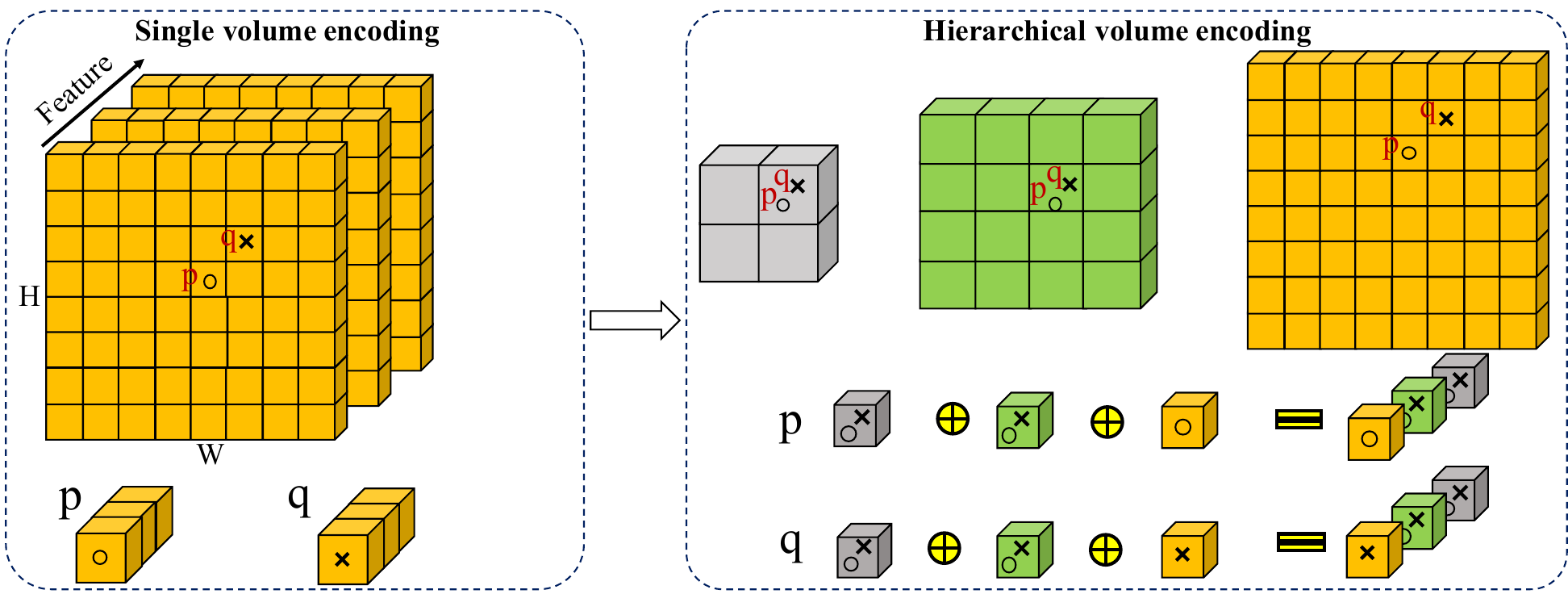}

\end{center}
 \caption{A 2D toy example of the hierarchical volume encoding. Left: a single high resolution volume. Features with  $3$ channels are used to encode two locations $p$ and $q$. Note it only captures spatial variant features. 
 Right: a hierarchical volume with lower dimensionality. Features have just $1$ channel and the memory consumption is much less. The high resolution volume encodes spatial variant features, while the low resolution volume enforces spatial smoothness. 
 }
\label{fig:hierarchical_volume}
\end{figure*}

%% file: tab_fig/sparse_table.tex
\begin{figure*}[]

\begin{center}
\includegraphics[width=1.5\columnwidth, trim={0cm 0cm 0.1cm 0cm}, clip]{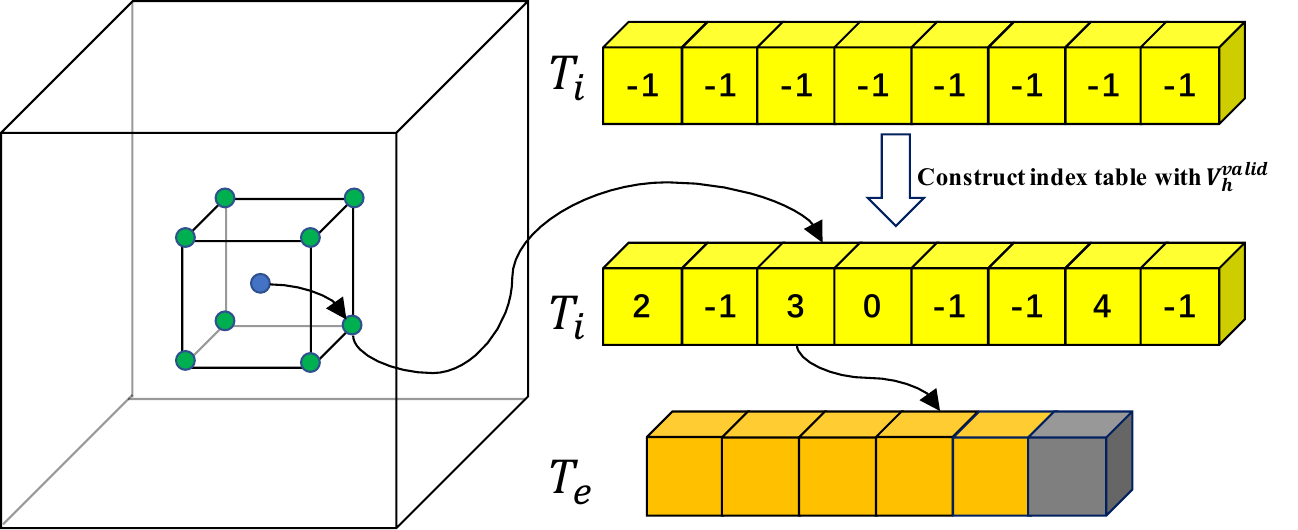}
\end{center}
 \caption{Sparse high-resolution volume. The index of $-1$ fetches the last embedding (in dark gray) in $T_e$.}
\label{fig:sparse_table}
\end{figure*}

%% file: tab_fig/dtu_table.tex
\setlength{\tabcolsep}{5.1pt}
\begin{table*}[]
\centering


\begin{tabular}{l c c c c c c c c c c c c c c c c}
\toprule
Method & 24 & 37 & 40 & 55 & 63 & 65 & 69 & 83 & 97 & 105 & 106 & 110 & 114 & 118 & 122 & Mean \\
\midrule
IDR\citep{idr} & $1.63$ & $1.87$ & $0.63$ & $0.48$ & $1.04$ & $0.79$ & $0.77$ & $1.33$ & $1.16$ & $0.76$ & $0.67$ & $0.90$ & $0.42$ & $0.51$ & $0.53$ & $0.90$ \\ 
MVSDF\citep{mvsdf} & $0.83$ & $1.76$ & $0.88$ & $0.44$ & $1.11$ & $0.90$ & $0.75$ & $1.26$ & $1.02$ & $1.35$ & $0.87$ & $0.84$ & $0.34$ & $0.47$ & $0.46$ & $0.88$ \\ 
COLMAP\citep{colmap_mvs} & $0.45$ & $0.91$ & $0.37$ & $0.37$ & $0.90$ & $1.00$ & $0.54$ & $1.22$ & $1.08$ & $0.64$ & $0.48$ & $0.59$ & $0.32$ & $0.45$ & $0.43$ & $0.65$ \\ 
\midrule    
NeRF\citep{mildenhall2020nerf} & $1.90$ & $1.60$ & $1.85$ & $0.58$ & $2.28$ & $1.27$ & $1.47$ & $1.67$ & $2.05$ & $1.07$ & $0.88$ & $2.53$ & $1.06$ & $1.15$ & $0.96$ & $1.49$ \\ 
UNISURF\citep{oechsle2021unisurf} & $1.32$ & $1.36$ & $1.72$ & $0.44$ & $1.35$ & $0.79$ & $0.80$ & $1.49$ & $1.37$ & $0.89$ & $0.59$ & $1.47$ & $0.46$ & $0.59$ & $0.62$ & $1.02$ \\ 
NeuS\citep{wang2021neus} & $1.00$ & $1.37$ & $0.93$ & $0.43$ & $1.10$ & $0.65$ & $0.57$ & $1.48$ & $1.09$ & $0.83$ & $0.52$ & $1.20$ & $0.35$ & $0.49$ & $0.54$ & $0.84$ \\ 
VolSDF\citep{yariv2021volsdf} & $1.14$ & $1.26$ & $0.81$ & $0.49$ & $1.25$ & $0.70$ & $0.72$ & $1.29$ & $1.18$ & $0.70$ & $0.66$ & $1.08$ & $0.42$ & $0.61$ & $0.55$ & $0.86$ \\ 
NeuralWarp\citep{darmon2022neuralwarp} & $0.49$ & $0.71$ & $0.38$ & $0.38$ & $0.79$ & $0.81$ & $0.82$ & $1.20$ & $1.06$ & $0.68$ & $0.66$ & $0.74$ & $0.41$ & $0.63$ & $0.51$ & $0.68$ \\ 
\midrule
NeuS+Hash & $1.26$ & $1.67$ & $0.84$ & $0.39$ & $1.19$ & $0.78$ & $0.64$ & $1.41$ & $1.24$ & $0.72$ & $0.58$ & $1.49$ & $0.34$ & $0.49$ & $0.52$ & $0.90$ \\ 
NeuS+Ours & $\mathbf{0.41}$ & $0.70$ & $\mathbf{0.35}$ & $\mathbf{0.35}$ & $0.85$ & $\mathbf{0.55}$ & $\mathbf{0.62}$ & $1.33$ & $1.03$ & $0.67$ & $0.53$ & $0.79$ & $\mathbf{0.33}$ & $\mathbf{0.42}$ & $\mathbf{0.45}$ & $0.63$ \\ 
VolSDF+Ours & $0.51$ & $0.77$ & $0.49$ & $0.38$ & $0.75$ & $0.67$ & $0.71$ & $1.16$ & $1.00$ & $\mathbf{0.63}$ & $0.57$ & $0.89$ & $0.38$ & $0.52$ & $0.48$ & $0.66$ \\ 
NeuralWarp+Ours & $0.44$ & $\mathbf{0.69}$ & $0.36$ & $0.38$ & $\mathbf{0.63}$ & $0.64$ & $0.69$ & $\mathbf{1.13}$ & $\mathbf{0.99}$ & $0.66$ & $\mathbf{0.51}$ & $\mathbf{0.61}$ & $0.36$ & $0.53$ & $0.52$ & $\mathbf{0.61}$ \\
\bottomrule
\end{tabular}

\caption{Quantitative results on the DTU dataset.}
\label{tab:dtu}
\end{table*}

%% file: tab_fig/epfl_table.tex
\setlength{\tabcolsep}{4.5pt}
\begin{table}[ht]
    \centering
    
    \footnotesize
    \begin{tabular}{l | c c c c | c c}
    \toprule
    \multirow{2}{*}{Method} & \multicolumn{2}{c}{Fountain-P11} & \multicolumn{2}{c|}{Herzjesu-P7} & \multicolumn{2}{c}{Mean} \\
    & Full & Center & Full & Center & Full & Center \\
    \midrule
    COLMAP & $6.47$ & $2.45$ & $\mathbf{7.95}$ & $2.31$ & $7.21$ & $2.38$ \\
    UNISURF & $26.16$ & $17.72$ & $27.22$ & $13.72$ & $26.69$ & $15.72$ \\
    MVSDF & $6.87$ & $2.26$ & $11.32$ & $2.72$ & $9.10$ & $2.49$ \\
    VolSDF & $12.89$ & $2.99$ & $13.61$ & $4.58$ & $13.25$ & $3.78$ \\
    NeuS &  $7.64$ & $2.25$ & $19.03$ & $14.02$ & $13.33$ & $7.79$ \\
    NeuralWarp & $7.77$ & $1.92$ & $8.88$ & $2.03$ & $8.32$ & $1.97$ \\
    \midrule
    NeuS+Ours & $7.39$ & $1.76$ & $11.23$ & $3.69$ & $9.31$ & $2.73$ \\
    NeualWarp+Ours & $\mathbf{3.31}$ & $\mathbf{1.73}$ & $8.13$ & $\mathbf{1.71}$ & $\mathbf{5.72}$ & $\mathbf{1.72}$ \\
    \bottomrule
    \end{tabular}
    \caption{Quantitative results on the EPFL dataset.}
    \label{tab:epfl}
\end{table}

%% file: includes_v2/experiments.tex
\section{EXPERIMENTS}

\input{tab_fig/vis_dtu_blendedmvs_epfl}

\input{tab_fig/ab_all_vis}

\input{tab_fig/ab_sparse_and_smooth}

\subsection{Implementation Details}
This work is implemented in Pytorch and experimented on Nvidia 2080Ti GPUs.
The Adam optimizer ($0.9, 0.999$) is used to update the network weights.
The learning rate for the MLPs is set to $5e^{-4}$ and decreased to ${1}/{20}$,
while the learning rates for the volumes with resolutions of
$2, 4, 8, 16, 32, 64, 128,$ $256, 512, 1024$ are set to
$1e^{-2},1e^{-2},1e^{-2},1e^{-2},1e^{-2},1e^{-3},$ $1e^{-3},1e^{-4},1e^{-4},1e^{-4}$ and decreased to ${1}/{100}$.



\textbf{Training strategy.}
We adopt a three-stage training strategy by default, and the number of optimization iterations is set to the same as the original framework.
Taking ``NeuS + Ours" as an example, there are $300$K iterations in total. 
In the first stage, eight volumes of resolution from $[2^3,4]$ to $[256^3,4]$ are employed for $80$K-iteration optimization. 
In the second stage, a sparse volume with the resolution of $[512^3,4]$ is added for another $20$K-iteration optimization. 
Finally, a sparse volume of $[1024^3,4]$ is appended for the remaining $200$K-iteration optimization.
The length of the index tables for the second and third stages are set to $256^3$.

\textbf{Details of sparse volume storing.}
We look up the values of $\mathbf{V}_h^{valid}$ from table $T_e$ through $T_i$.
Specifically, we first define a mapping function $f$: 
\begin{equation}
   f(\mathbf{x}) = x + y * N + z * N^2,
\end{equation}
where $N$ is the resolution of the volume, and $x, y, z$ are the coordinates of the position $\mathbf{x}$.
We utilize $f$ to map each voxel $\mathbf{v}_j$ in $\mathbf{V}_h^{valid}$ to the index in table $T_i$, and then $T_i$ further converts it to the index $j$ in $T_e$: 
\begin{equation}
    T_i({f}(\mathbf{x})) = j,\  j = 1...n
    \label{equ:indx}
\end{equation}
where $n$ is the number of valid voxels in $\mathbf{V}_h^{valid}$. 
The length of the embedding table $T_e$ is only $n+1$ ($n \ll N^3$), such that the memory consumption is reduced a lot due to the sparse structure of the high-resolution volumes.

\subsection{Datasets}
DTU~\citep{jensen2014dtudataset} is a well-used dataset for multi-view reconstruction, consisting of $124$ scans of various objects. 
The images are obtained from an RGB camera and a structured light scanner mounted on an industrial robot arm.
Each scene is captured from 49 or 64 views with a resolution of $1600\times 1200$.
For a fair comparison, we follow previous methods and select 15 scenes from DTU for evaluation.
BlendedMVS~\citep{yao2020blendedmvs} is another dataset for multi-view reconstruction composed of 113 scenes including architectures, sculptures, and small objects.
Each scene consists of dozens to hundreds of images with  a resolution of $768 \times 576$.
EPFL~\citep{strecha2008epfl} is a small dataset composed of two outdoor scenes, Fountain and Herzjesu, which contain 11 and 9 high-resolution images, respectively, and the accurate ground truth meshes.

\subsection{Evaluation}
For a fair comparison, we follow previous methods to evaluate our method on DTU~\citep{jensen2014dtudataset}
, EPFL~\citep{strecha2008epfl} and BlendedMVS~\citep{yao2020blendedmvs} benchmark.
The Chamfer L1 distance is used for evaluating the accuracy of the recovered surfaces. 
This metric is the average of the accuracy, which measures the distance from the reconstructed surface to the ground-truth surface, and the completeness, which measures the distance in reverse.

We first evaluate our method on the DTU dataset and report the quantitative results in Table~\ref{tab:dtu}.
{For a fair comparison, the meshes of all methods are extracted by marching cubes with resolution of $512$.}
We follow previous work~\citep{darmon2022neuralwarp, oechsle2021unisurf} to clean the predicted mesh by visibility masks for more reasonable evaluation.
As shown in Table~\ref{tab:dtu}, the accuracy of previous methods is improved significantly by adding the hierarchical volume encoding.
To be specific, the error of NeuS~\citep{wang2021neus} is reduced by $25\% $, from $0.84$ to $0.63$, 
while the error of NeuralWarp~\citep{darmon2022neuralwarp} is reduced by $10\% $, from $0.68$ to $0.61$.
The reconstructed meshes are shown in Figure~\ref{fig:dtu blendedmvs vis}, from where we can see that the surfaces of our method are smooth and clean, while containing accurate geometry details.
For instance, the details of the house in the first scene in Figure~\ref{fig:dtu blendedmvs vis} are recovered, especially the shape of the windows and the bricks of the roof, which is more remarkable in the normal map in Figure~\ref{fig:vis normal}.

We then evaluate our method on the EPFL dataset.
For a fair comparison, we follow NeuralWarp to use both the ``full" chamfer distance and the ``center" chamfer distance to evaluate the reconstructed surfaces. 
The ``center" metric only evaluates the center of the scene cropped by a manually defined box,
which focuses more on the precision of the reconstruction,
while the ``full" metric is also influenced by the ground plane and the rarely seen points, which thus also considers the completeness of the reconstruction.
As shown in Table~\ref{tab:epfl}, the performance of each framework is improved significantly by adding our method.
Especially, the ``full" metric of NeuralWarp is decreased by $31\%$ to $5.72$, which is also a $21\%$ improvement compared to COLMAP~\citep{colmap_mvs}.

Finally, the evaluation is performed on the BlendedMVS dataset, which contains objects of more complex shapes.
As shown in Figure~\ref{fig:dtu blendedmvs vis}, our method can recover more detailed shapes than previous methods.
These illustrate that our hierarchical volume encoding can facilitate MLPs to encode more complex shapes and improve neural implicit surface reconstruction.




\input{tab_fig/ab_num_vol_and_res}

\input{tab_fig/ab_num_mlps}

\subsection{Ablation Study}
We choose ``NeuS + Ours" to perform extensive ablation studies to validate the proposed method.
All the results are reported from the 15 scans of DTU dataset unless otherwise specified.

\textbf{Hierarchical volume encoding.}
To study the effectiveness of the hierarchical volume encoding, we equip the basic set of eight volumes to NeuS~\citep{wang2021neus} without any other change.
As shown in Table~\ref{tab:ab_sparse_smooth}, the error of the recovered surfaces is significantly reduced from $0.84$ to $0.70$.

\textbf{Regularization terms.}
We also ablate the two smoothness terms to study their effectiveness. As shown in  Table~\ref{tab:ab_sparse_smooth}, they help to further reduce the error metric down to $0.66$. Figure~\ref{fig:ab_all_vis} (b) shows a visual comparison without (middle) and with (right) the regularization terms. The reconstructed mesh is smoother and more complete with the regularization terms.

\textbf{Sparse high resolution volume.}
We analyze the effect of the resolution of volumes.
As shown in Table~\ref{tab:ab_sparse_smooth}, sparse high-resolution embedding volume (resolution of 1024 here) further improves the accuracy to $0.63$.
We also visualize the reconstructed surfaces without (left) and with (right) sparse high-resolution embedding in Figure~\ref{fig:ab_all_vis}(a). It is clear that the high-resolution embedding generates more shape details, e.g., clearer facial shapes.

\input{tab_fig/vis_psnr_24.tex}

\input{tab_fig/dtu_normal.tex}

\input{tab_fig/dtu_psnr.tex}

\input{tab_fig/dtu_psnr_24.tex}

\input{tab_fig/vis_ngp.tex}

\input{tab_fig/vis_encoding.tex}

\textbf{Setup of embedding volumes.}
To further inspect the effect of different combinations of the volume encoding, we perform two experiments on the volume setups.
(1) We study the number of the volumes.
We first compare our hierarchical setup to a large volume of $256^3 \times 32$, the number of whose learnable parameters is about $8$ times larger than ours.
However, its performance is much worse than ours, as is presented in the left of Figure~\ref{fig:ab_num_vol_and_com_res}.
Then we keep the number of feature channels fixed at $32$ and increase the volume number. 
The resolution of the first volume is set to $256$, and the resolution of the remaining volumes is decreased by $1/2$ one by one.
As shown in Figure~\ref{fig:ab_num_vol_and_com_res} left, the accuracy of the reconstructed surface is gradually improved with the arising of the volume number.
(2) Next, we study the impact of the low-resolution volume.
We keep the resolution of the largest volume fixed at $256$, the number of the volumes fixed at $8$, and vary the resolution of the smallest volume from $2$ to $64$.
The resolutions of intermediate volumes are sequentially enlarged, and the enlargement factor is $(256/min\_res)^{1/8}$.
From the results shown in Figure~\ref{fig:ab_num_vol_and_com_res} right, we can see the performance is decreased when the low-resolution volume begins from a larger resolution. This is intuitive because the low-resolution volume enforces spatial smoothness. 
Starting from a lower resolution helps to enforce smoothness across larger areas.

\textbf{Layer number of MLPs.}
We perform experiments on the layer number of MLPs in the SDF network to inspect its effect. 
As shown in Table~\ref{tab:ab_num_mlp}, the mean chamfer distance decreases with the increase of the layer number of MLPs.
NeuS~\citep{wang2021neus} adopts $9$ layers of MLPs and gets an error of $0.84$, while the error of our method already decreases to $0.78$ with only $2$ layers.

\subsection{Evaluation on normal consistency metric}

While the Chamfer distance is a good metric to compare the similarity of two point clouds, there have been many questions\cite{wu2021density} about whether the Chamfer distance could capture the high-frequency details or overall distributional similarity between two point clouds. 
As suggested in \cite{mescheder2019occupancy}, a normal consistency measure can indicate whether the method can capture high-frequency surfaces.

We report the normal consistency metric in Table~\ref{tab:dtu_normal}.
This score is obtained by first calculating the mean absolute dot product of the normals in the reconstructed mesh and the normals at the corresponding nearest neighbors in the ground-truth mesh, and then calculating that in reverse.

\vspace{-3mm}

\subsection{Evaluation of novel view synthesis}

To evaluate the effectiveness of our method in novel view synthesis, we perform the experiments and report the results in Table~\ref{tab:dtu_psnr}, Table~\ref{tab:dtu_psnr_24}, and Figure~\ref{fig:vis_psnr_24}.
We select one of each 7 images from the original image set of DTU as the test views, and the remaining images as the training views.
The accuracy of novel view synthesis on 15 scans of DTU is reported in Table~\ref{tab:dtu_psnr}, while the PSNR result with respect to iteration number is reported in Table~\ref{tab:dtu_psnr_24}.
From the results in these two tables and the images displayed in Figure~\ref{fig:vis_psnr_24}, we can see both the accuracy and convergence speed of the novel view synthesis benefit from our hierarchical volume encoding.





\subsection{Comparison to Hash Encoding}

A similar hierarchical feature encoding is adopted in Instant-NGP~\citep{mueller2022instant}, but the features are stored in hash tables such that the hash collision is inevitable.
Although this is efficient in both memory and convergence speed, the hash collision may result in defects in the implicit geometry.
To make a comparison to hash encoding, we perform two experiments: one is the original version of Instant-NGP, and the other one is adding the hash encoding to NeuS framework. 
Some of the results are presented in Figure~\ref{fig:ab_all_vis}(c) and Table~\ref{tab:dtu}.
A visual comparison is displayed in Figure~\ref{fig:vis_ngp},
from where although Instant-NGP or Hash+NeuS could obtain accurate novel view synthesis, there are some defects in their surface. 



\subsection{Visualization of volume encodings}

To inspect how the volumes help encode the 3D space, we visualize the volumes in Figure~\ref{fig:vis_encoding}.
Due to the 4D structure of the volumes, we perform the marching cubes with a threshold to do the visualization.
Following the threshold selection of NeRF, the threshold is set to $0$ since the values of most of the voxels are distributed around $0$, as shown in the histogram of Figure~\ref{fig:vis_encoding}.
From the displayed histograms and volumes, we can see the values in the volumes are set to be distributed randomly ($\mu = 0, \sigma = 0.02$) before the training, and distributed close to the object surface after the training.
Also, the volumes in different resolutions represent different fineness of the scene.

%% file: tab_fig/vis_dtu_blendedmvs_epfl.tex
\begin{figure*}[t]
\centering
\begin{subfigure}{0.03\columnwidth}
  \centering
  \includegraphics[width=1\columnwidth, trim={0cm -1.3cm 0cm 0cm}, clip]{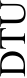}
\end{subfigure}
\begin{subfigure}{0.47\columnwidth}
  \centering
  \includegraphics[width=1\columnwidth, trim={0cm 0cm 0cm 2cm}, clip]{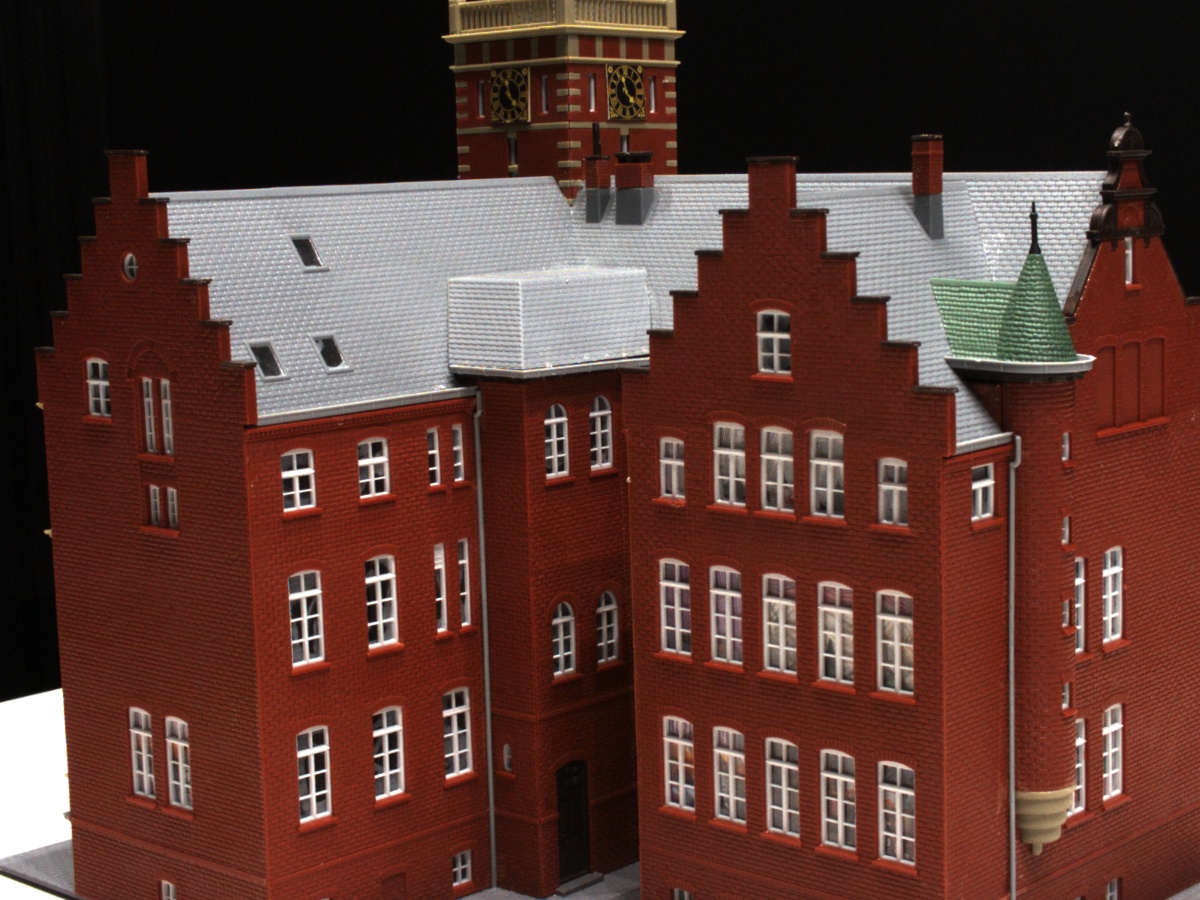}
\end{subfigure}
%
\begin{subfigure}{0.47\columnwidth}
  \centering
  \includegraphics[width=1\columnwidth, trim={0cm 0cm 0cm 2cm}, clip]{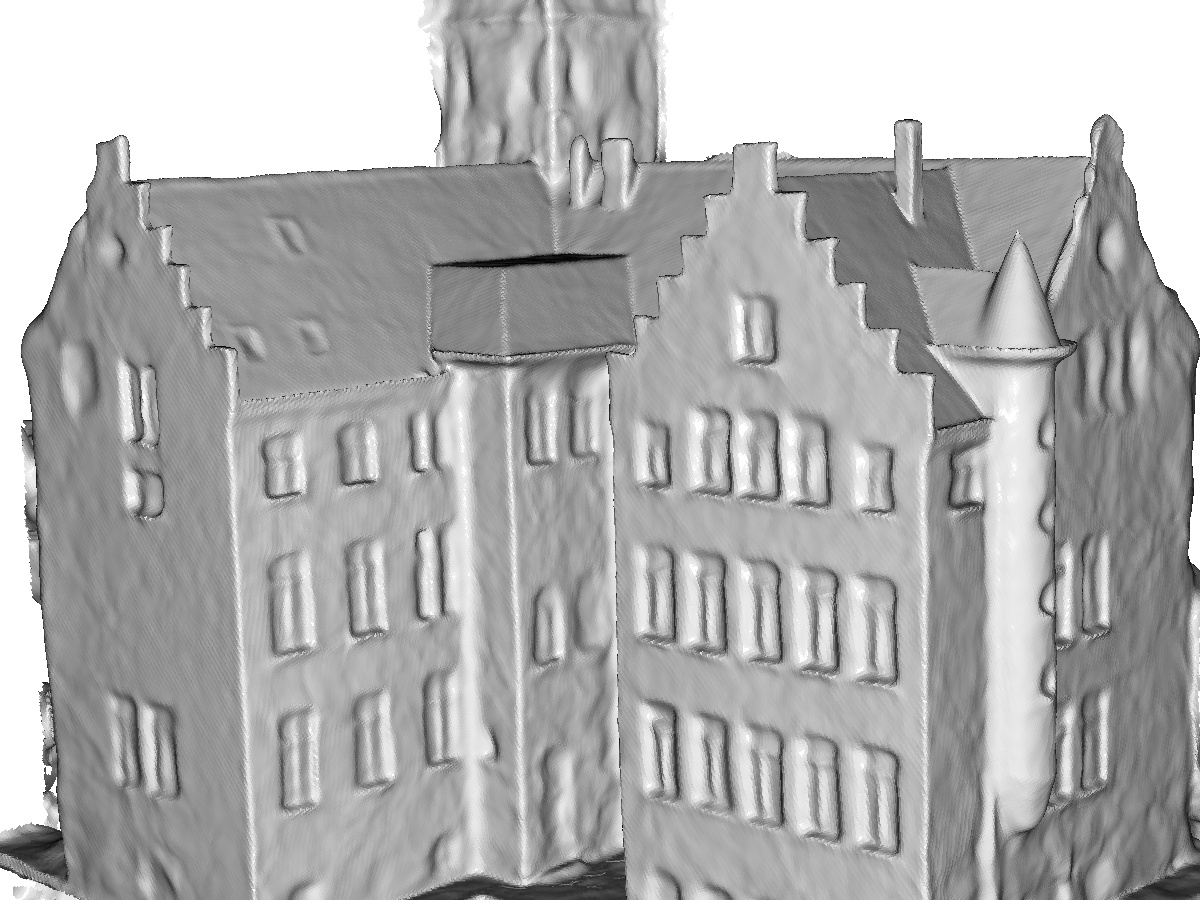}
\end{subfigure}
\begin{subfigure}{0.47\columnwidth}
  \centering
  \includegraphics[width=1\columnwidth, trim={0cm 0cm 0cm 2cm}, clip]{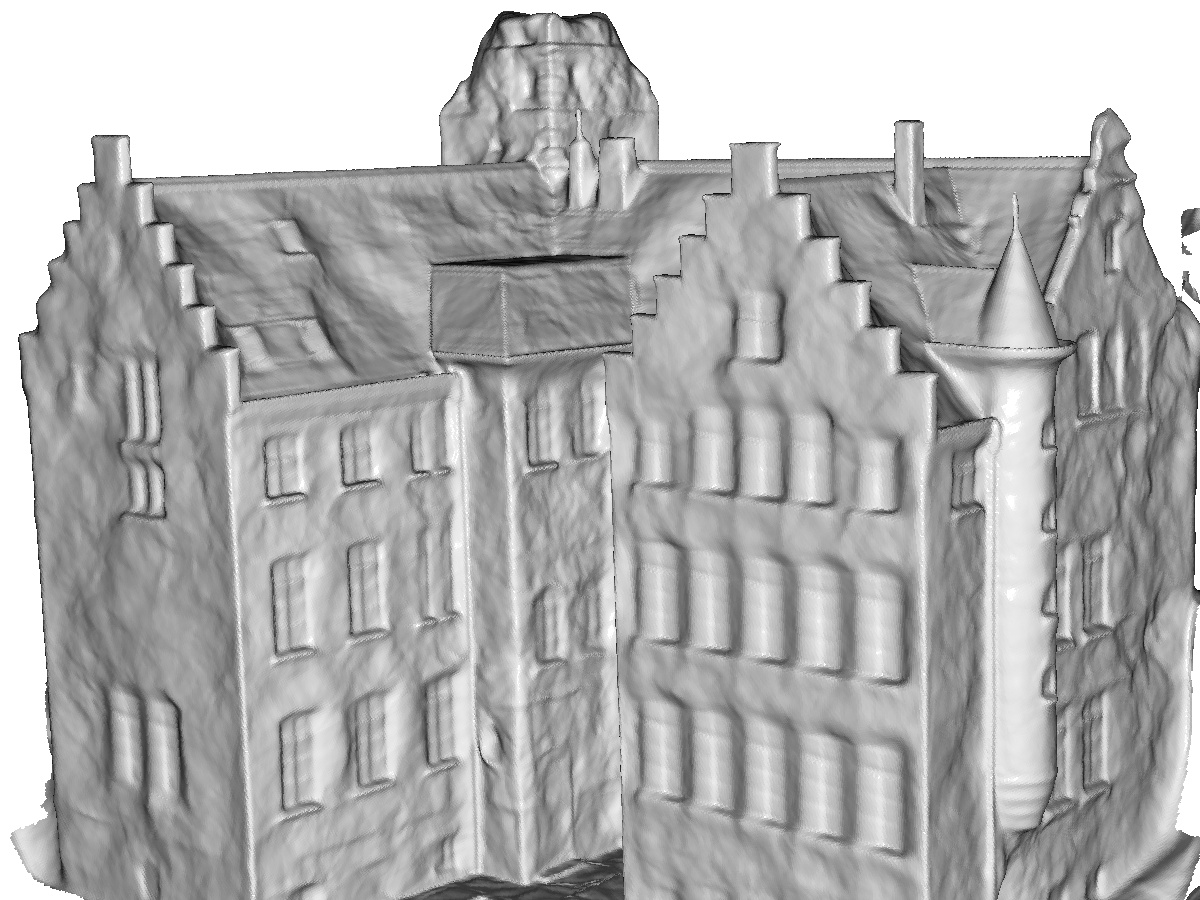}
\end{subfigure}
\begin{subfigure}{0.47\columnwidth}
  \centering
  \includegraphics[width=1\columnwidth, trim={0cm 0cm 0cm 2cm}, clip]{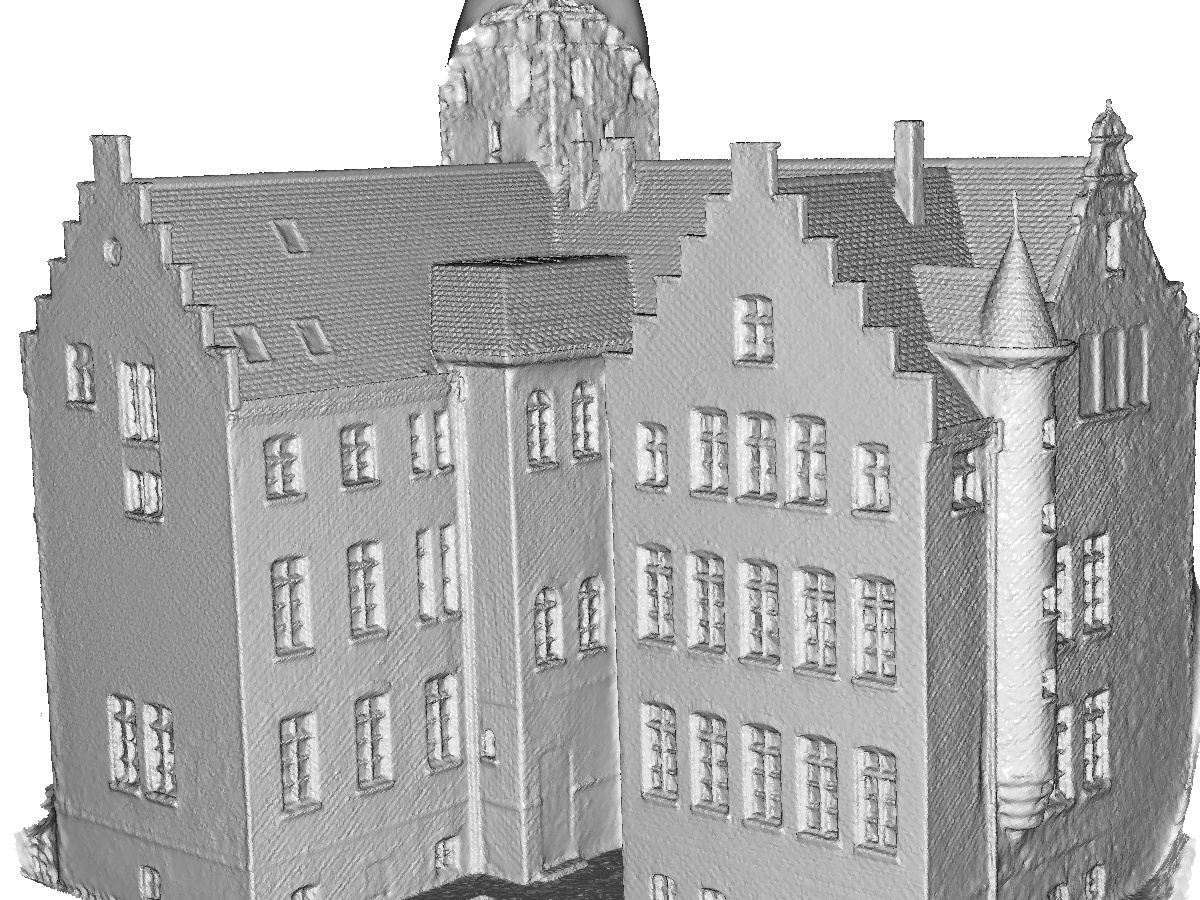}
\end{subfigure}

\begin{subfigure}{0.03\columnwidth}
  \centering
  \includegraphics[width=1\columnwidth, trim={0cm -1.4cm 0cm 0cm}, clip]{figs/compare/dtu_text.pdf}
\end{subfigure}
\begin{subfigure}{0.47\columnwidth}
  \centering
  \includegraphics[width=1\columnwidth, trim={0cm 0cm 0cm 0cm}, clip]{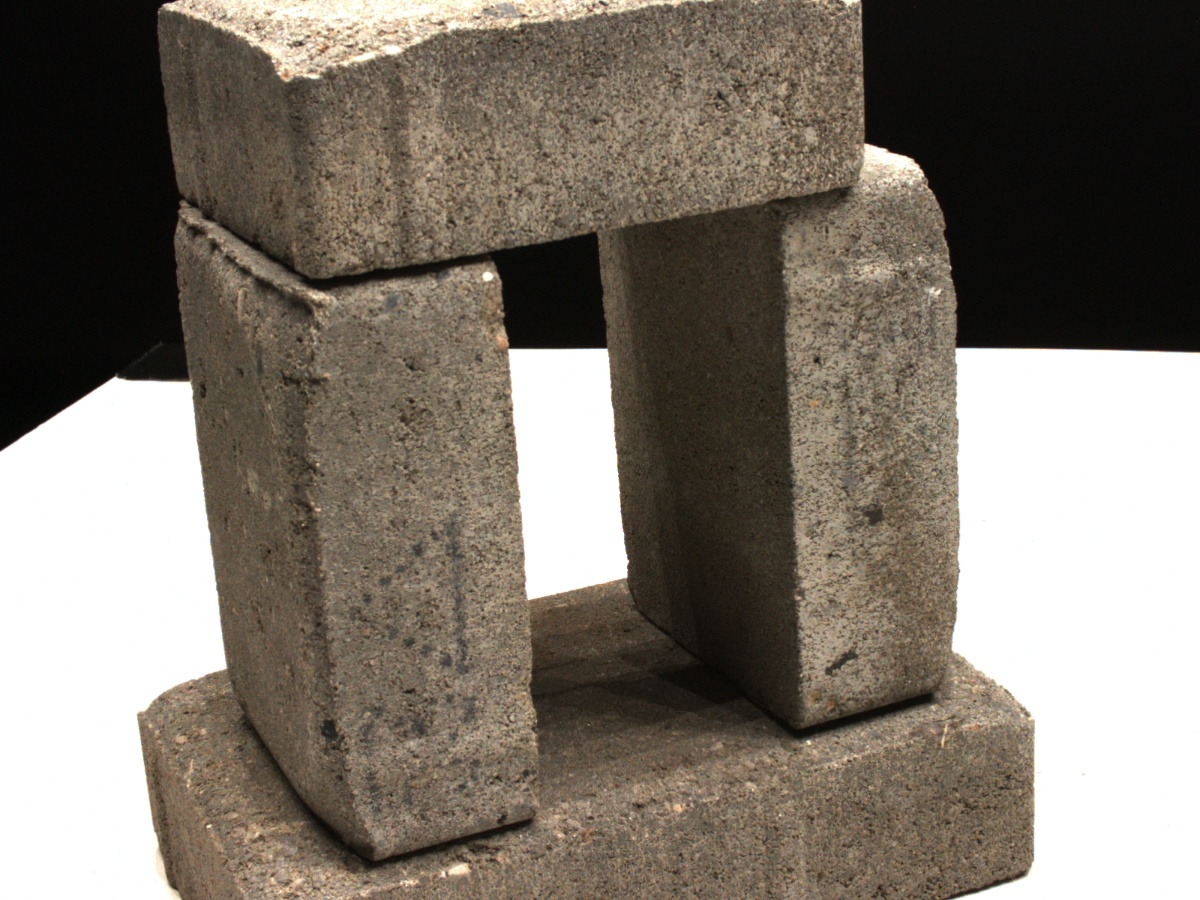}
\end{subfigure}
%
\begin{subfigure}{0.47\columnwidth}
  \centering
  \includegraphics[width=1\columnwidth, trim={0cm 0cm 0cm 0cm}, clip]{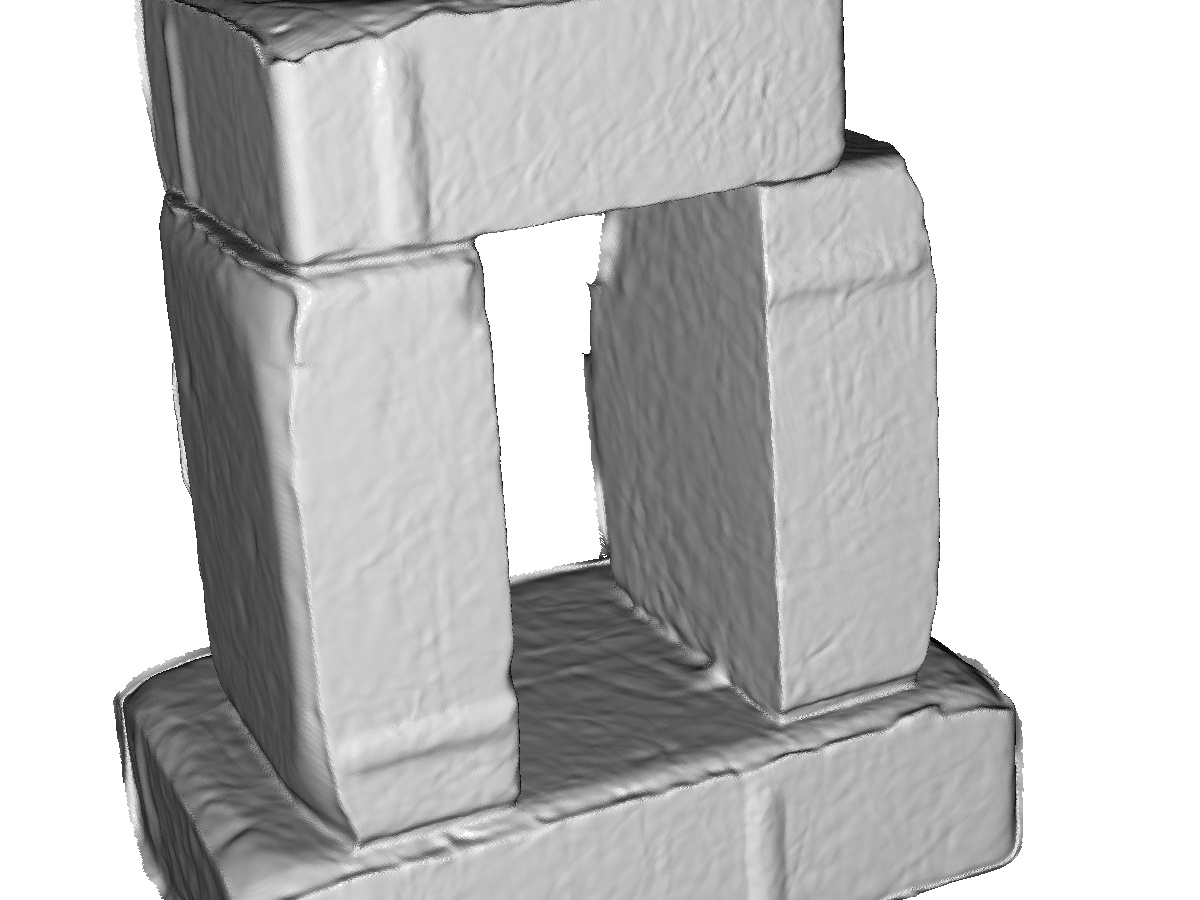}
\end{subfigure}
\begin{subfigure}{0.47\columnwidth}
  \centering
  \includegraphics[width=1\columnwidth, trim={0cm 0cm 0cm 0cm}, clip]{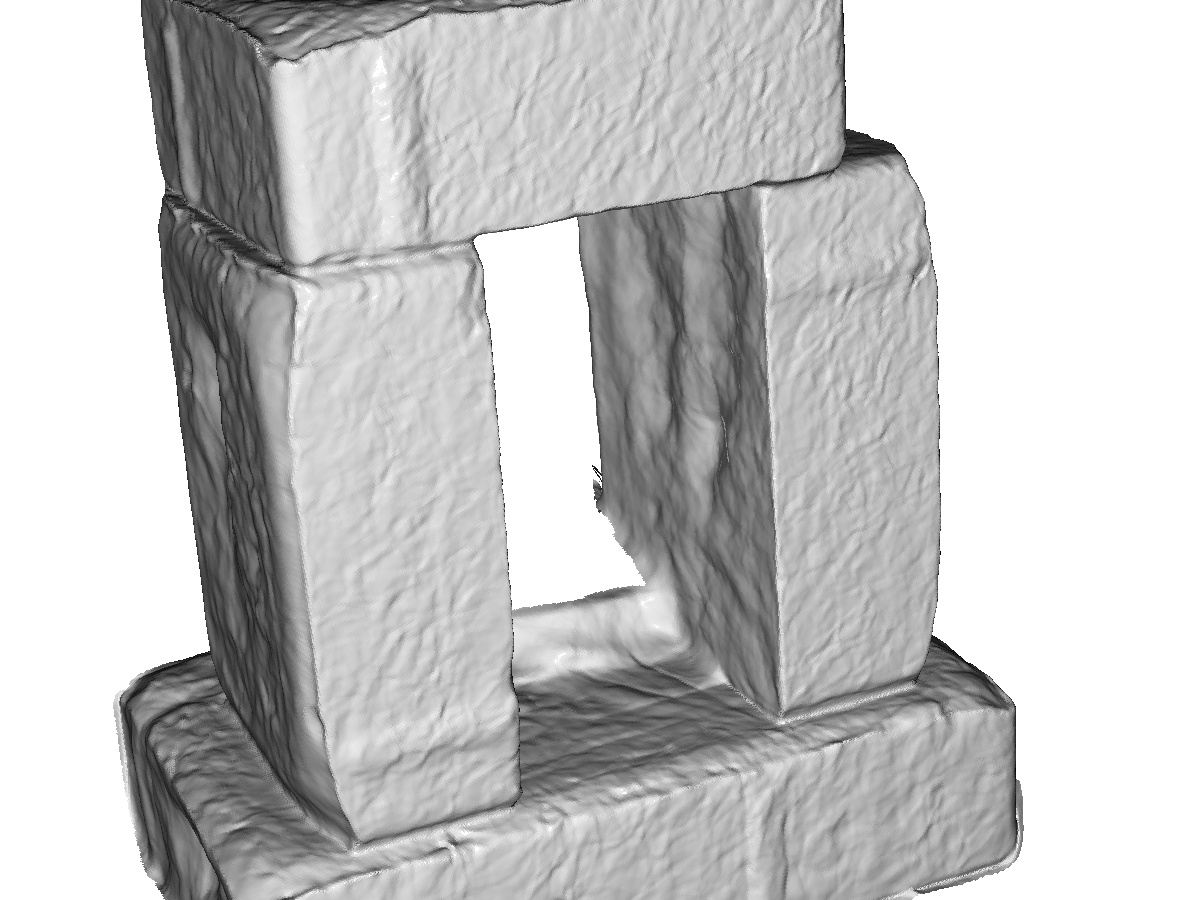}
\end{subfigure}
\begin{subfigure}{0.47\columnwidth}
  \centering
  \includegraphics[width=1\columnwidth, trim={0cm 0cm 0cm 0cm}, clip]{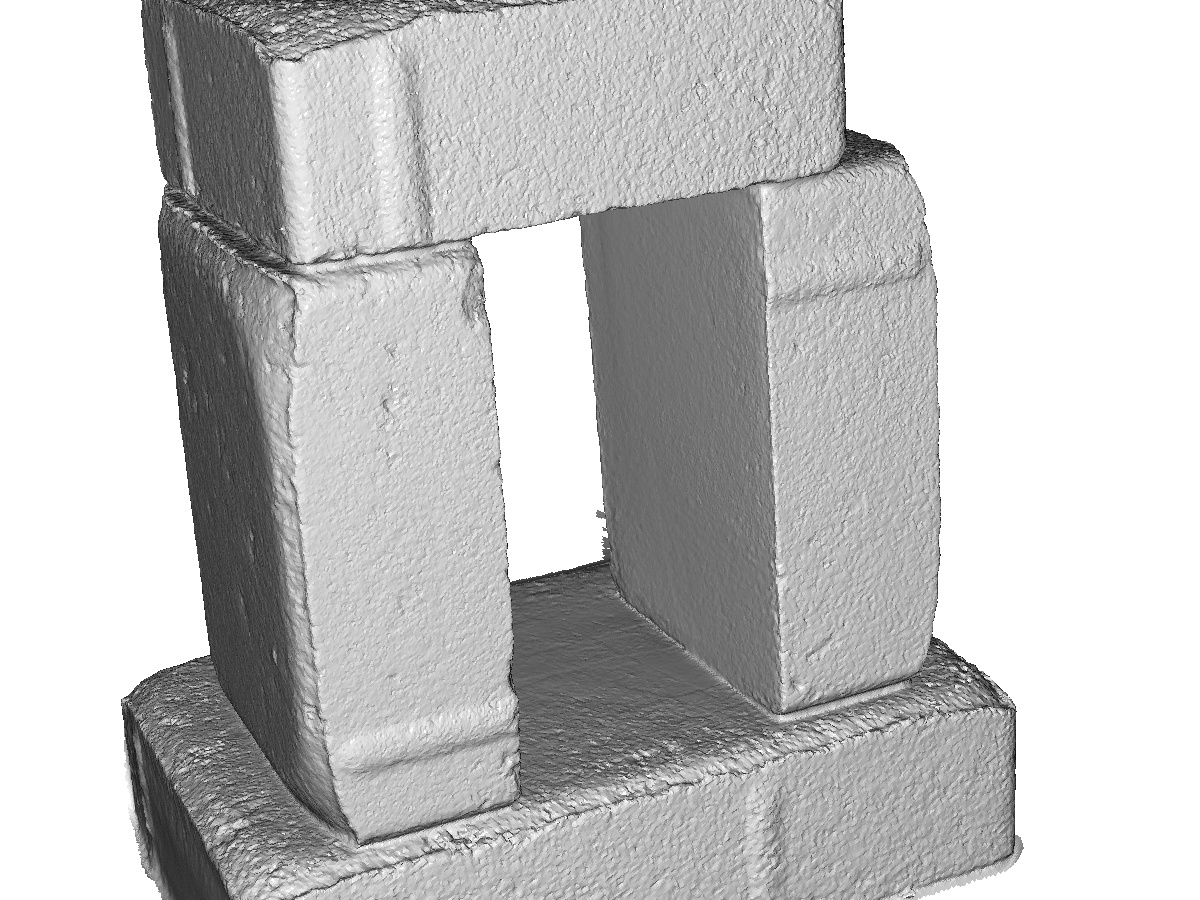}
\end{subfigure}

\begin{subfigure}{0.03\columnwidth}
  \centering
  \includegraphics[width=1\columnwidth, trim={0cm -0.4cm 0cm 0cm}, clip]{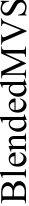}
\end{subfigure}
\begin{subfigure}{0.47\columnwidth}
  \centering
  \includegraphics[width=1\columnwidth, trim={0cm 0cm 0cm 0cm}, clip]{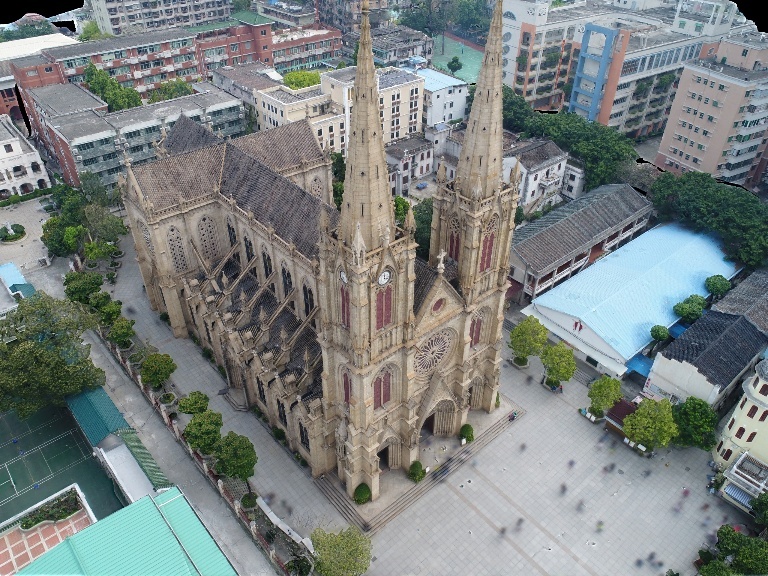}
\end{subfigure}
%
\begin{subfigure}{0.47\columnwidth}
  \centering
  \includegraphics[width=1\columnwidth, trim={0cm 0cm 0cm 0cm}, clip]{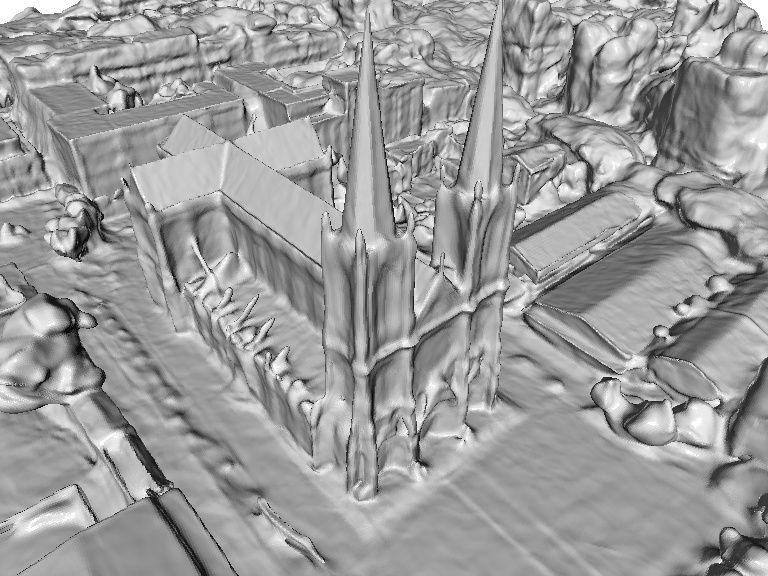}
\end{subfigure}
\begin{subfigure}{0.47\columnwidth}
  \centering
  \includegraphics[width=1\columnwidth, trim={0cm 0cm 0cm 0cm}, clip]{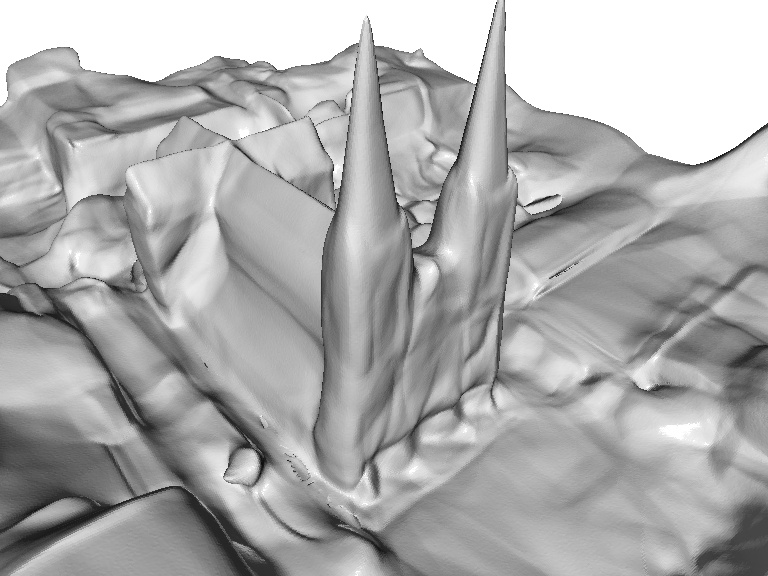}
\end{subfigure}
\begin{subfigure}{0.47\columnwidth}
  \centering
  \includegraphics[width=1\columnwidth, trim={0cm 0cm 0cm 0cm}, clip]{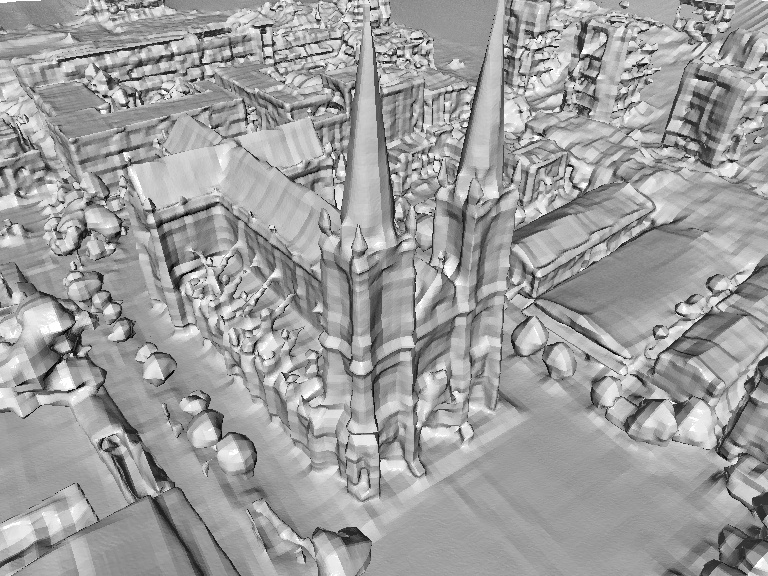}
\end{subfigure}

\begin{subfigure}{0.03\columnwidth}
  \centering
  \includegraphics[width=1\columnwidth, trim={0cm -0.3cm 0cm 0cm}, clip]{figs/compare/blendedmvs_text.pdf}
\end{subfigure}
\begin{subfigure}{0.47\columnwidth}
  \centering
  \includegraphics[width=1\columnwidth, trim={0cm 0cm 0cm 0cm}, clip]{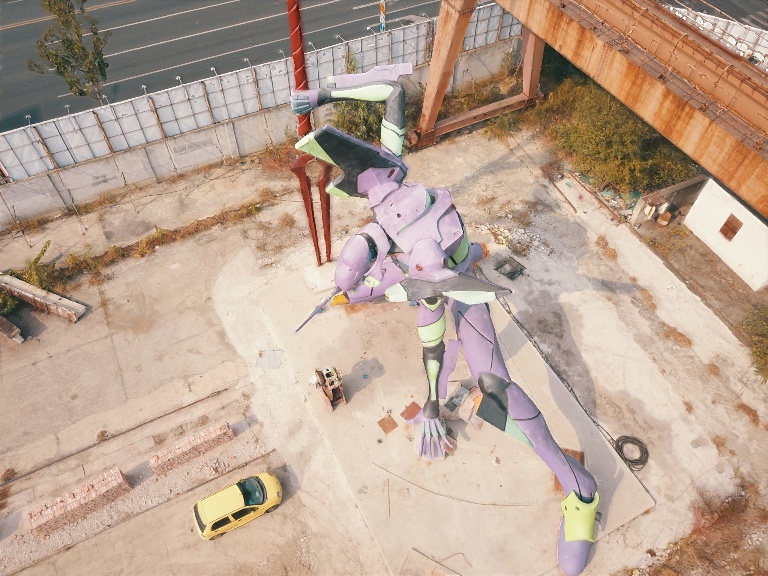}
\end{subfigure}
%
\begin{subfigure}{0.47\columnwidth}
  \centering
  \includegraphics[width=1\columnwidth, trim={0cm 0cm 0cm 0cm}, clip]{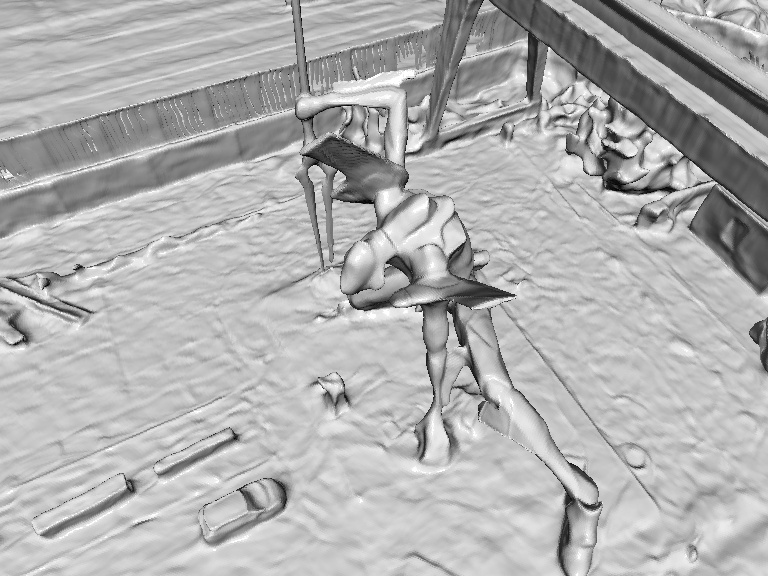}
\end{subfigure}
\begin{subfigure}{0.47\columnwidth}
  \centering
  \includegraphics[width=1\columnwidth, trim={0cm 0cm 0cm 0cm}, clip]{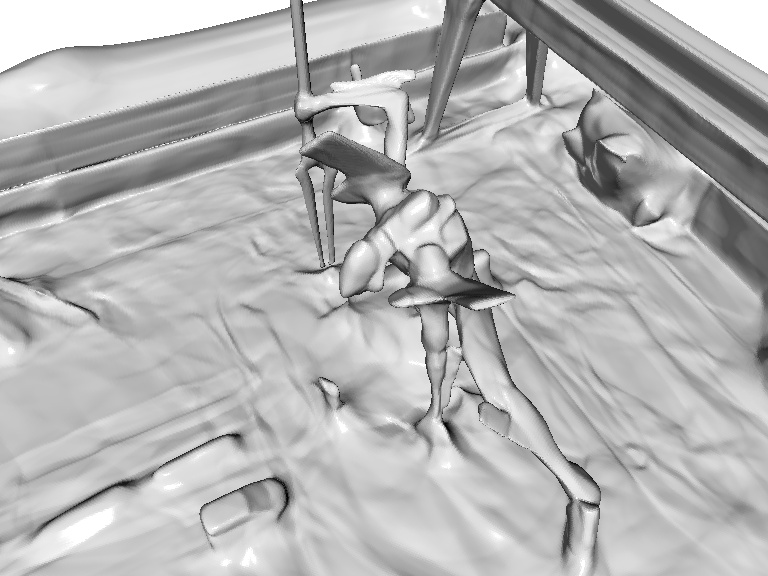}
\end{subfigure}
\begin{subfigure}{0.47\columnwidth}
  \centering
  \includegraphics[width=1\columnwidth, trim={0cm 0cm 0cm 0cm}, clip]{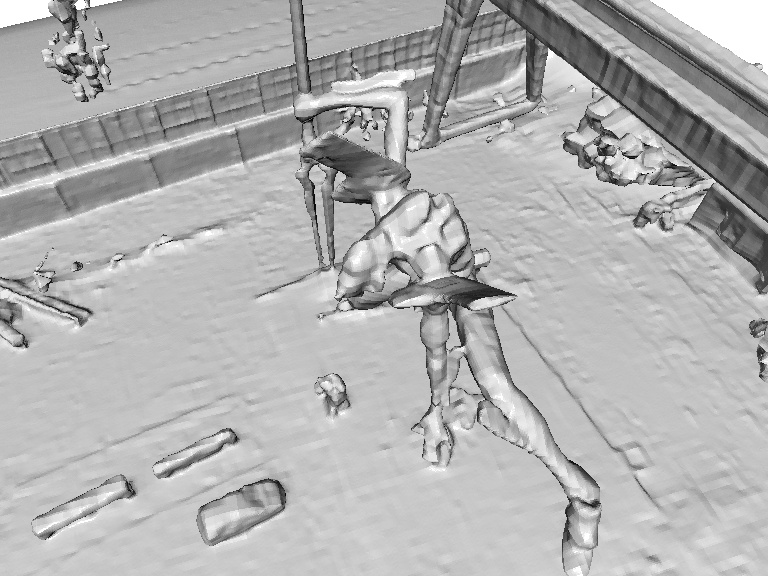}
\end{subfigure}

\begin{subfigure}{0.03\columnwidth}
  \centering
  \includegraphics[width=1\columnwidth, trim={0cm -0.3cm 0cm 0cm}, clip]{figs/compare/blendedmvs_text.pdf}
\end{subfigure}
\begin{subfigure}{0.47\columnwidth}
  \centering
  \includegraphics[width=1\columnwidth, trim={0cm 0cm 0cm 0cm}, clip]{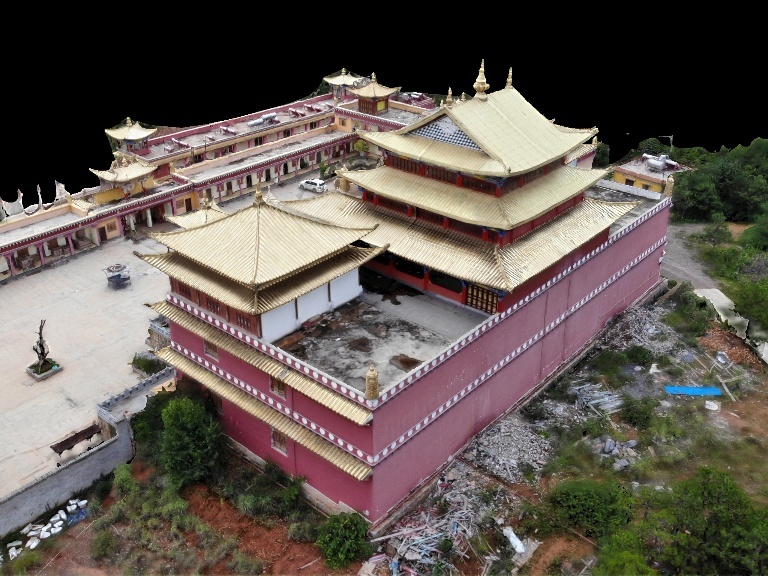}
\end{subfigure}
%
\begin{subfigure}{0.47\columnwidth}
  \centering
  \includegraphics[width=1\columnwidth, trim={0cm 0cm 0cm 0cm}, clip]{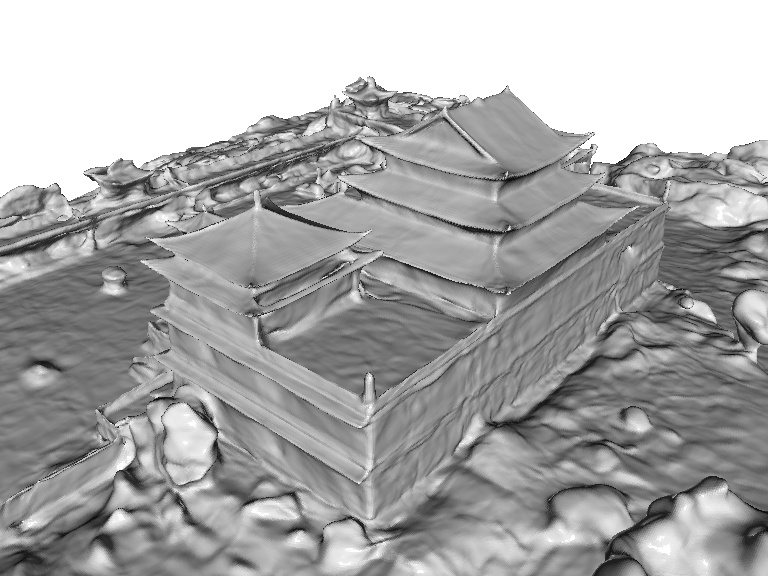}
\end{subfigure}
\begin{subfigure}{0.47\columnwidth}
  \centering
  \includegraphics[width=1\columnwidth, trim={0cm 0cm 0cm 0cm}, clip]{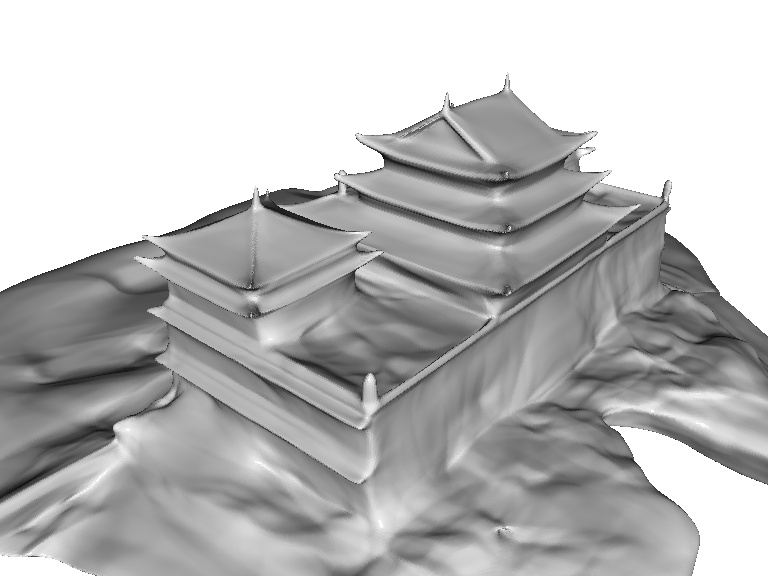}
\end{subfigure}
\begin{subfigure}{0.47\columnwidth}
  \centering
  \includegraphics[width=1\columnwidth, trim={0cm 0cm 0cm 0cm}, clip]{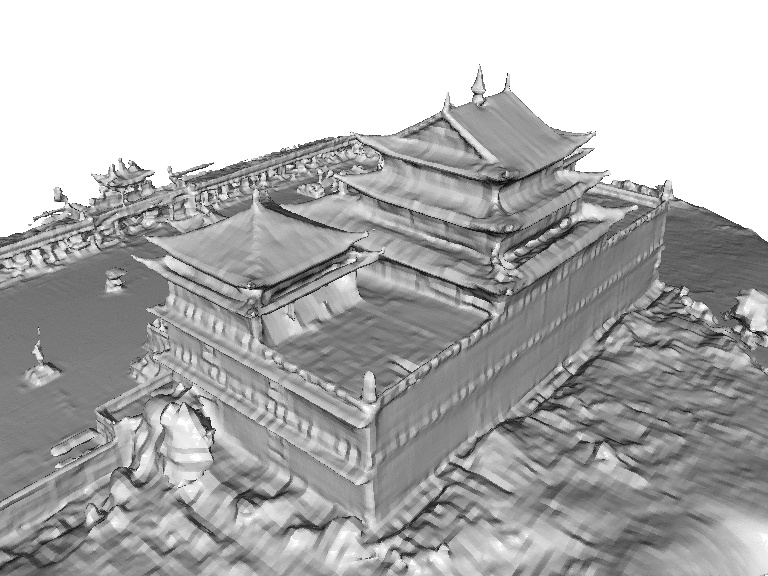}
\end{subfigure}


\begin{subfigure}{0.03\columnwidth}
  \centering
  \includegraphics[width=1\columnwidth, trim={0cm -2.2cm 0cm 0cm}, clip]{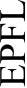}
\end{subfigure}
\begin{subfigure}{0.47\columnwidth}
  \centering
  \includegraphics[width=1\columnwidth, trim={0cm 0cm 0cm 0cm}, clip]{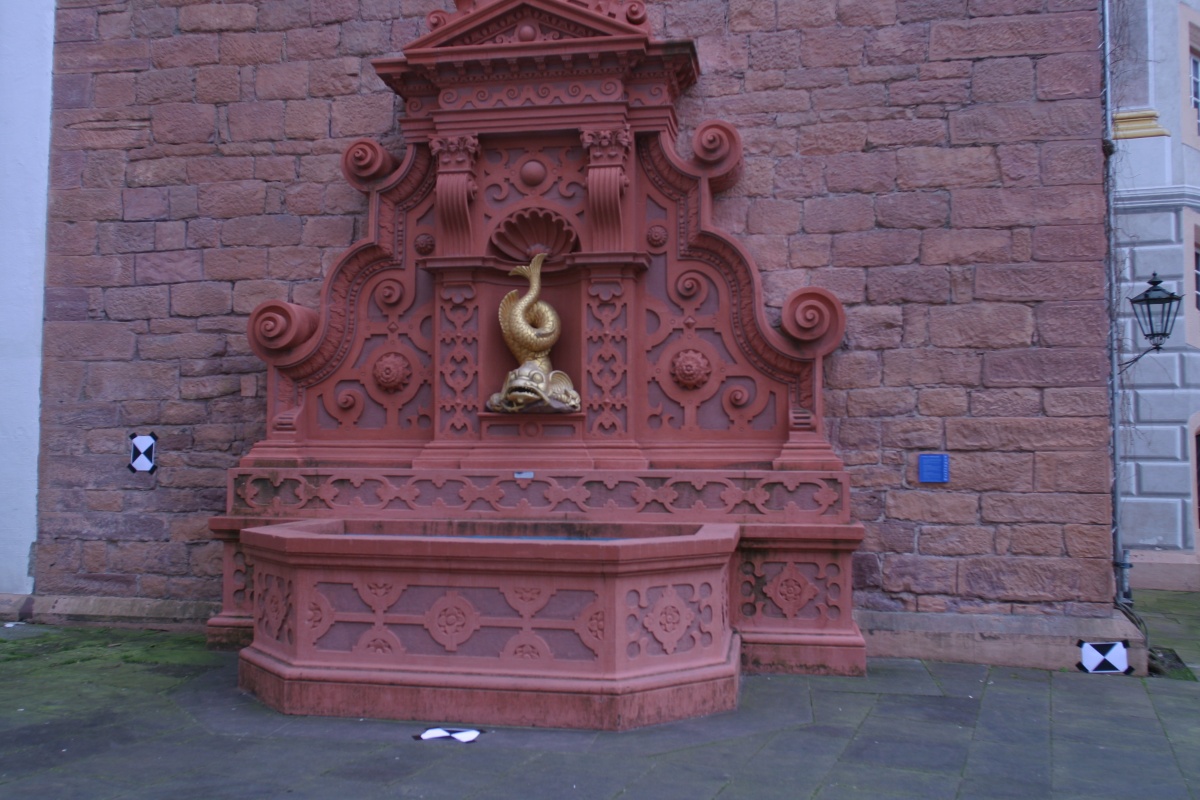}
  \caption*{Image}
\end{subfigure}
%
\begin{subfigure}{0.47\columnwidth}
  \centering
  \includegraphics[width=1\columnwidth, trim={0cm 0cm 0cm 0cm}, clip]{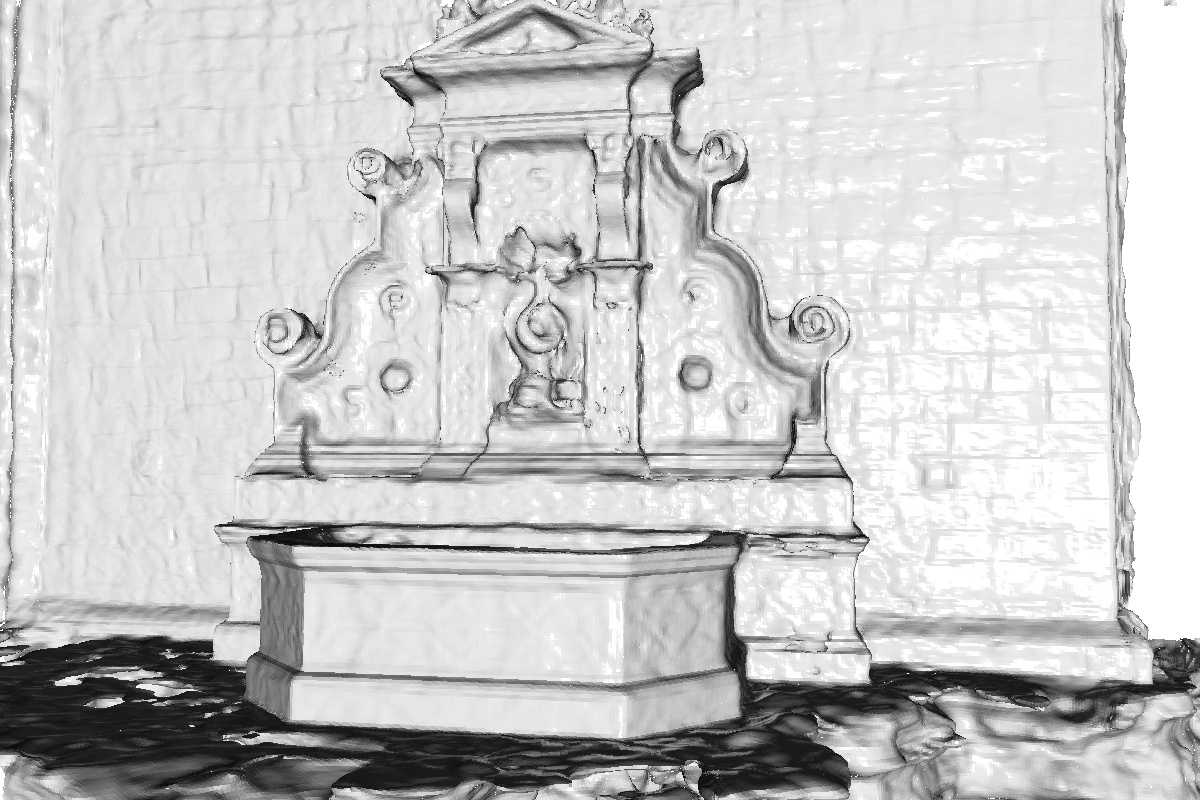}
  \caption*{NeuralWarp}
\end{subfigure}
\begin{subfigure}{0.47\columnwidth}
  \centering
  \includegraphics[width=1\columnwidth, trim={0cm 0cm 0cm 0cm}, clip]{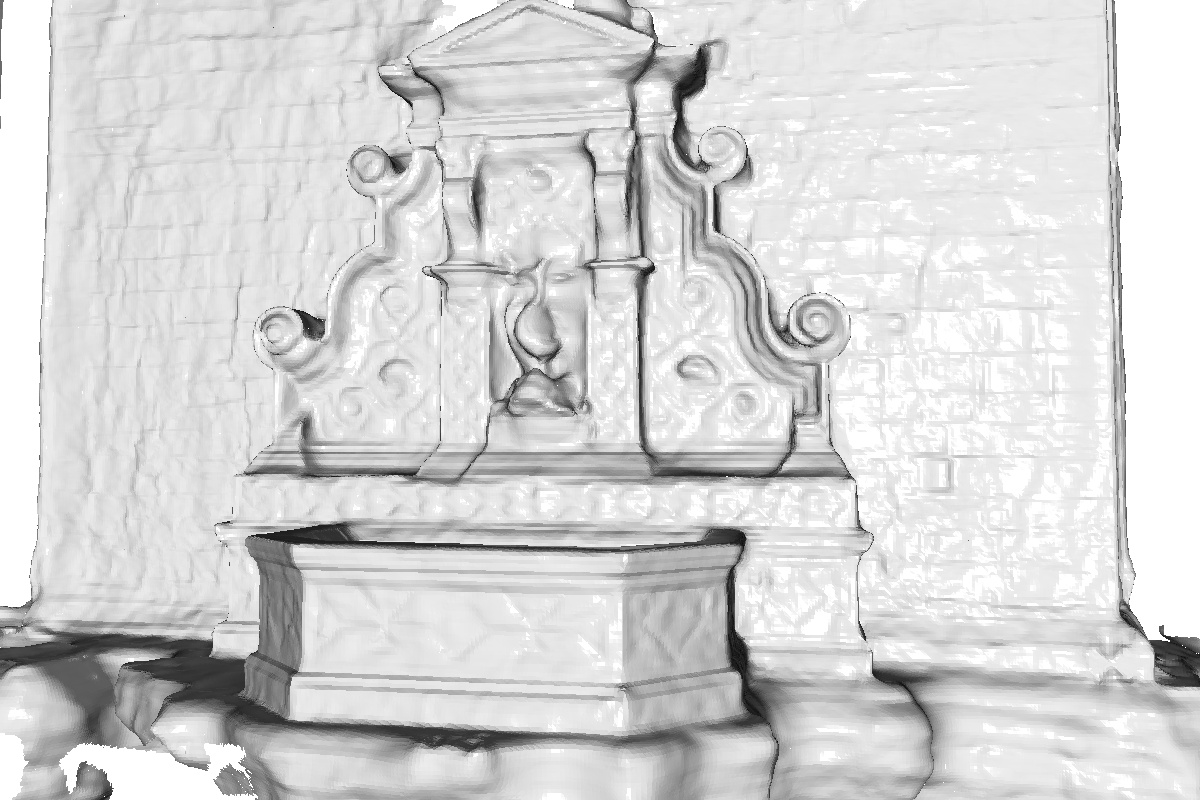}
  \caption*{NeuS}
\end{subfigure}
\begin{subfigure}{0.47\columnwidth}
  \centering
  \includegraphics[width=1\columnwidth, trim={0cm 0cm 0cm 0cm}, clip]{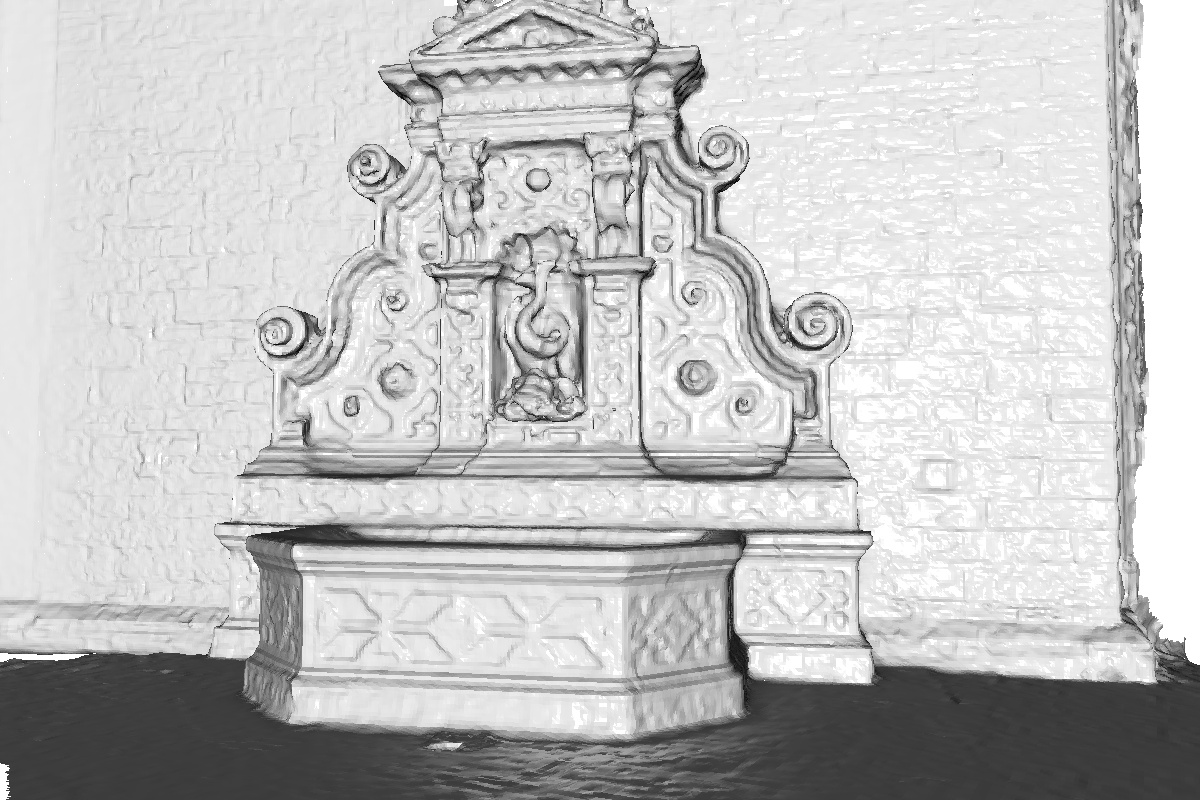}
  \caption*{NeuS+Ours}
\end{subfigure}

 \caption{Visual comparison of the reconstructed meshes.}
\label{fig:dtu blendedmvs vis}
\end{figure*}

%% file: tab_fig/ab_all_vis.tex
\begin{figure*}[t]
\begin{center}

  \includegraphics[width=1.95\columnwidth, trim={0cm 0cm 0cm 0cm}, clip]{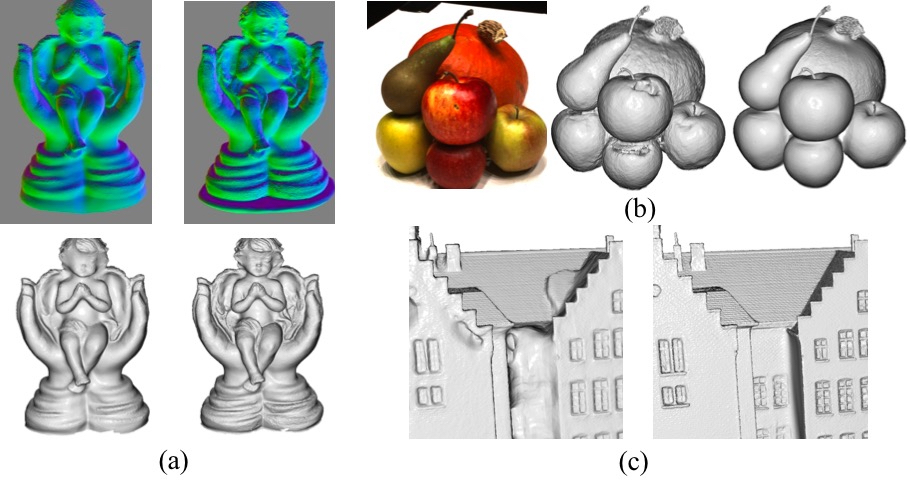}

\end{center}
\vspace{-4mm}
 \caption{Ablation study. (a) Sparse high-res volume. {The results of without and with sparse high-resolution volumes are displayed in the first and second rows, respectively. The top row shows the normal maps and the bottom row shows the reconstructed meshes.} (b) Regularization terms. {The color image, the results of without and with regularization terms are displayed.} (c) NeuS+Hash vs. NeuS+Ours.}
\label{fig:ab_all_vis}
\end{figure*}

%% file: tab_fig/ab_sparse_and_smooth.tex
\setlength{\tabcolsep}{8pt}
\begin{figure}[t]
\centering
\begin{tabular}{c c c |c}
\toprule
Hierarchical
& Regular
& Sparse
& \multirow{2}{*}{Mean} \\
Volume & Terms & High-Res &   \\
\midrule
\xmark & \xmark & \xmark & $0.84$ \\
\cmark & \xmark & \xmark & $0.70$ \\ 
\cmark & \cmark & \xmark & $0.66$ \\
\cmark & \cmark & \cmark & $0.63$ \\ 
\bottomrule
\end{tabular}
\caption{Ablation study on the DTU dataset.}
\label{tab:ab_sparse_smooth}
\end{figure}

%% file: tab_fig/ab_num_vol_and_res.tex
\begin{figure*}[t]
    \centering
    
    \begin{subfigure}{0.98\columnwidth}
      \centering
      \includegraphics[width=1\columnwidth, trim={0.1cm 0.5cm 0.1cm 0.1cm}, clip]{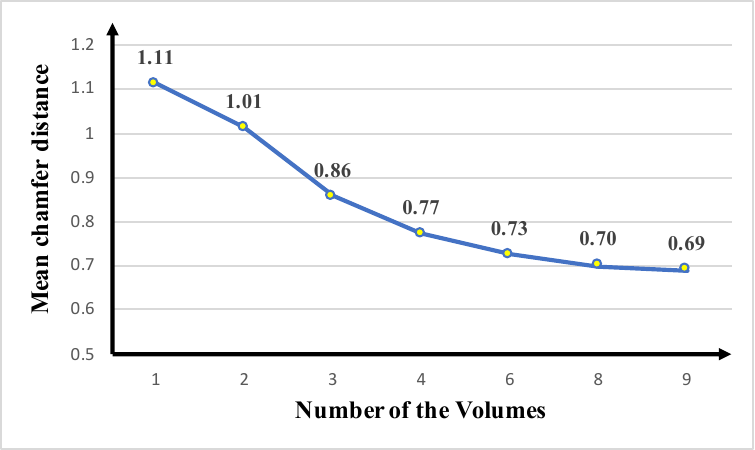} 
    \end{subfigure}
    \begin{subfigure}{0.98\columnwidth}
      \centering
    \includegraphics[width=1\columnwidth, trim={0.1cm 0.5cm 0.1cm 0.1cm}, clip]{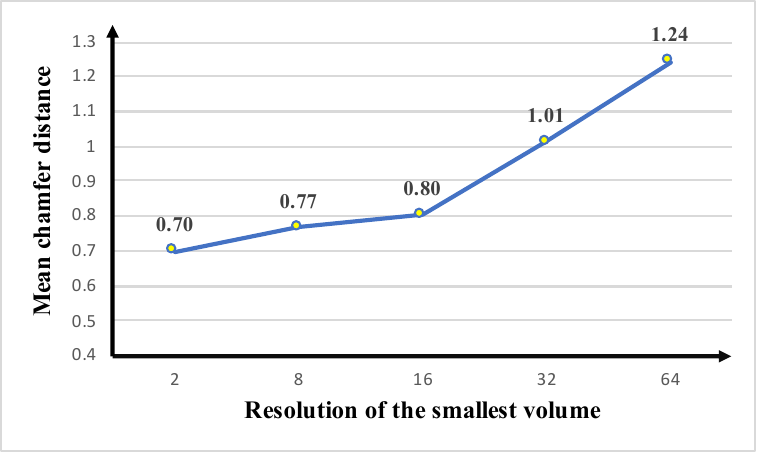} 
    \end{subfigure}
    
    \caption{Ablation study about the number of the volumes and the resolution combination of volumes.}
    \label{fig:ab_num_vol_and_com_res}
\end{figure*}

%% file: tab_fig/ab_num_mlps.tex
\setlength{\tabcolsep}{6pt}
\begin{table*}[ht]
\centering


\begin{tabular}{c c | c c c c c c c c c c c c c c c c}
\toprule
Method & Num. & 24 & 37 & 40 & 55 & 63 & 65 & 69 & 83 & 97 & 105 & 106 & 110 & 114 & 118 & 122 & Mean \\
\midrule
NeuS & 9 & $1.00$ & $1.37$ & $0.93$ & $0.43$ & $1.10$ & $0.65$ & $0.57$ & $1.48$ & $1.09$ & $0.83$ & $0.52$ & $1.20$ & $0.35$ & $0.49$ & $0.54$ & $0.84$ \\
\midrule
\multirow{5}{*}{NeuS+Ours} & 1 & $1.59$ & $1.67$ & $0.56$ & $0.94$ & $0.96$ & $0.73$ & $0.97$ & $1.95$ & $1.82$ & $0.95$ & $1.48$ & $1.67$ & $1.73$ & $0.83$ & $0.78$ & $1.24$ \\ 
& 2 & $0.53$ & $0.78$ & $0.73$ & $0.36$ & $0.90$ & $0.66$ & $0.74$ & $1.48$ & $1.10$ & $0.72$ & $0.55$ & $1.59$ & $0.36$ & $0.57$ & $0.64$ & $0.78$ \\ 
& 3 & $0.49$ & $0.75$ & $0.72$ & $0.37$ & $0.87$ & $0.58$ & $0.63$ & $1.34$ & $1.11$ & $0.74$ & $0.49$ & $1.58$ & $0.34$ & $0.42$ & $0.46$ & $0.73$ \\ 
& 5 & $0.51$ & $0.74$ & $0.58$ & $0.36$ & $0.89$ & $0.58$ & $0.66$ & $1.39$ & $1.05$ & $0.78$ & $0.51$ & $1.47$ & $0.36$ & $0.41$ & $0.42$ & $0.71$ \\ 
& 9 & $0.42$ & $0.77$ & $0.38$ & $0.34$ & $0.81$ & $0.61$ & $0.65$ & $1.26$ & $1.10$ & $0.65$ & $0.49$ & $1.60$ & $0.32$ & $0.58$ & $0.51$ & $0.70$ \\ 
\bottomrule
\end{tabular}

\caption{Quantitative results for the different layer number of MLPs on the DTU dataset.}
\label{tab:ab_num_mlp}
\end{table*}

%% file: tab_fig/vis_psnr_24.tex
\begin{figure*}[ht]
\centering

%
\begin{subfigure}{0.6\columnwidth}
  \centering
  \includegraphics[width=1\columnwidth, trim={6cm 0cm 0cm 3cm}, clip]{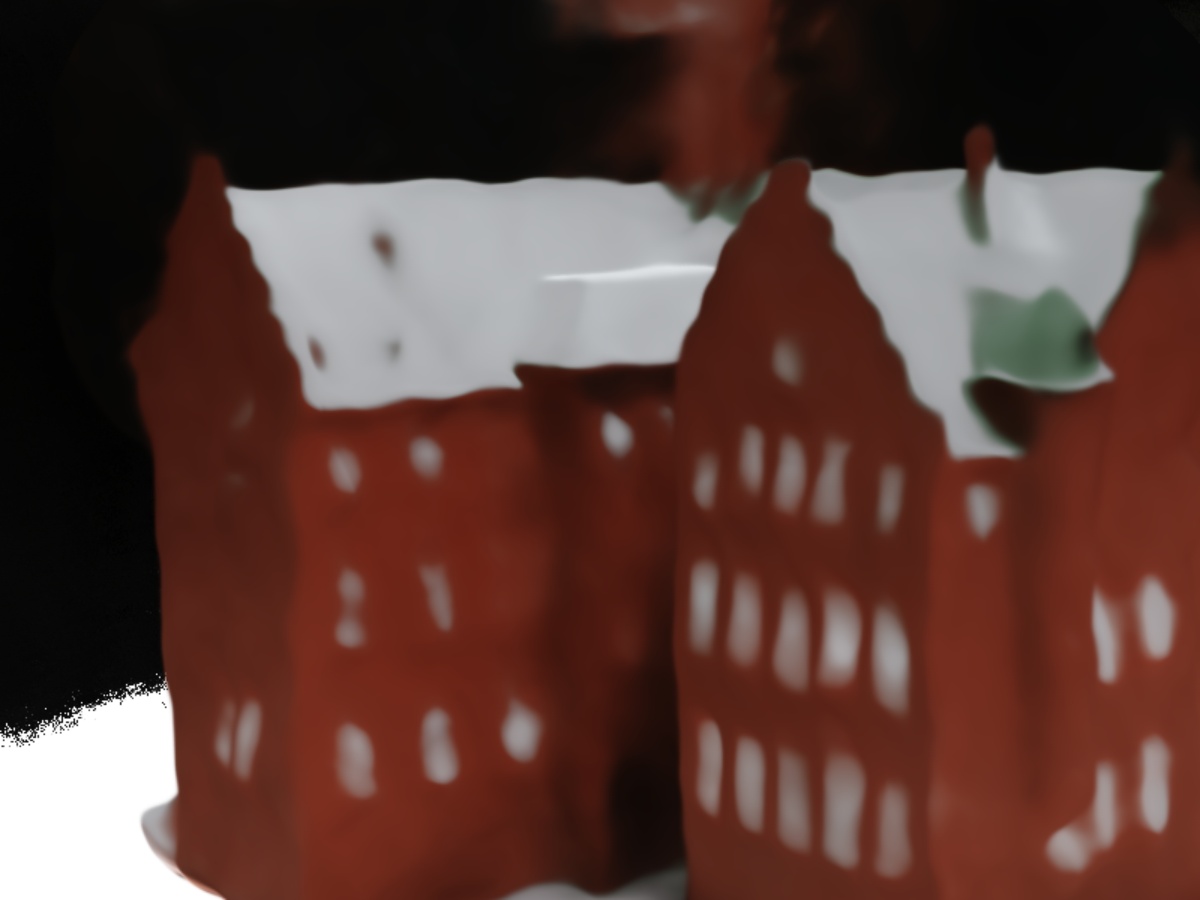}
  \caption*{PSNR: $18.51$}
\end{subfigure}
\begin{subfigure}{0.6\columnwidth}
  \centering
  \includegraphics[width=1\columnwidth, trim={6cm 0cm 0cm 3cm}, clip]{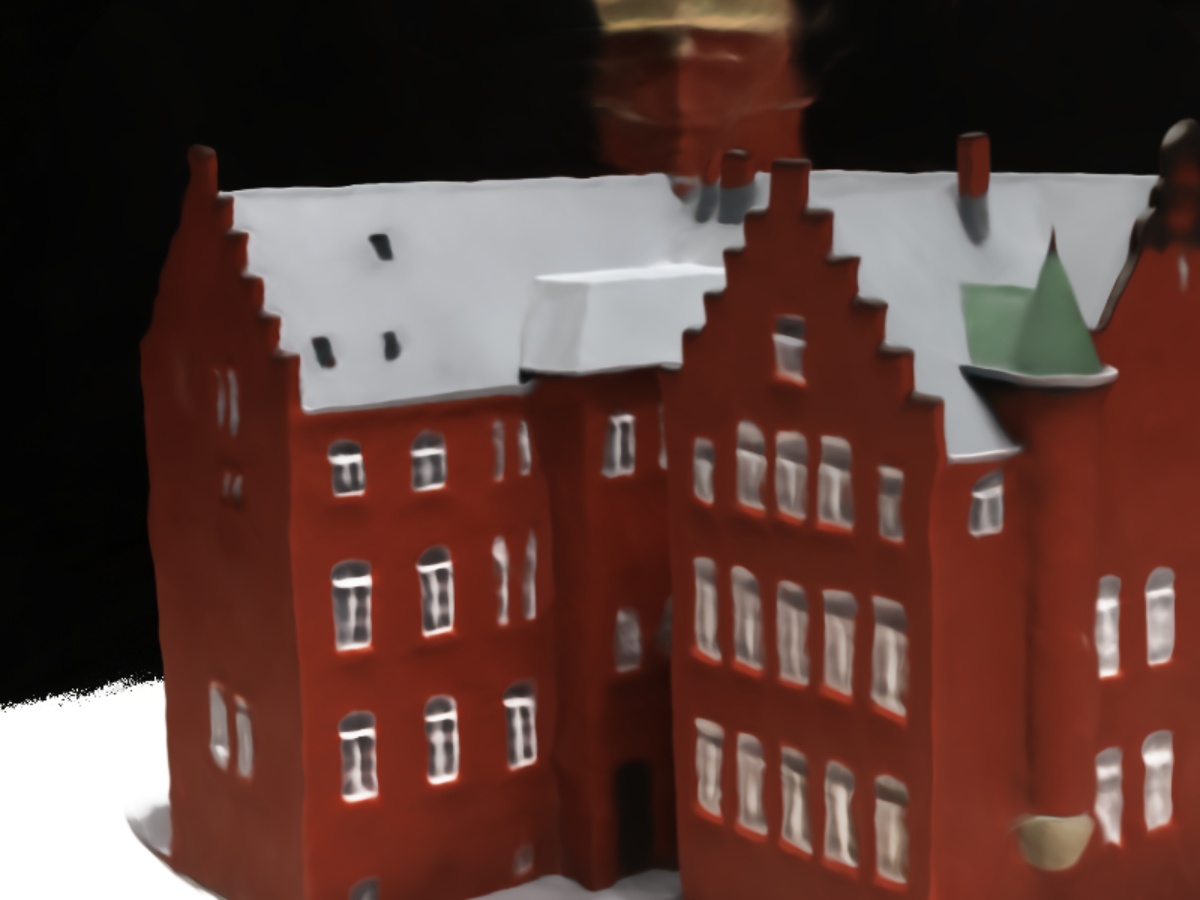}
  \caption*{PSNR: $21.93$}
\end{subfigure}
\begin{subfigure}{0.6\columnwidth}
  \centering
  \includegraphics[width=1\columnwidth, trim={6cm 0cm 0cm 3cm}, clip]{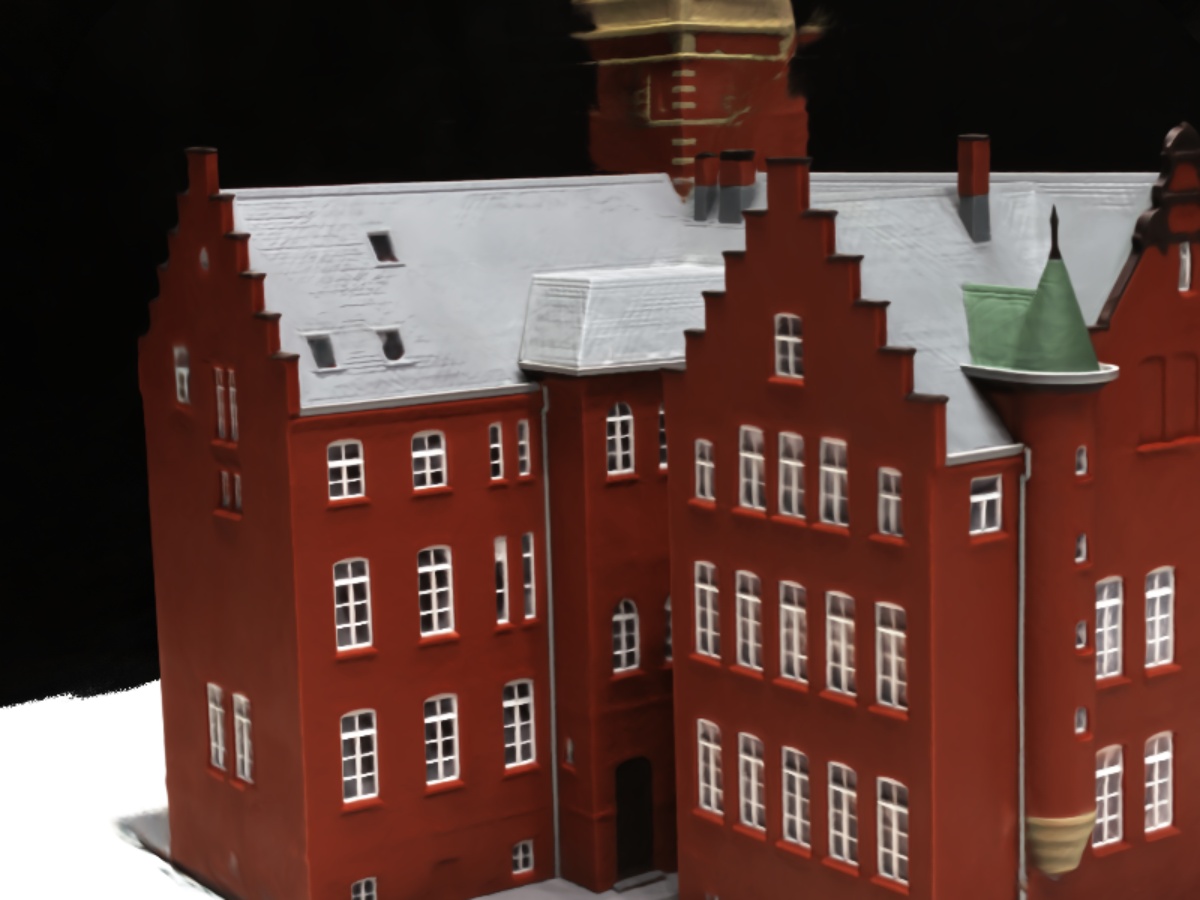}
  \caption*{PSNR: $24.31$}
\end{subfigure}

%
\begin{subfigure}{0.6\columnwidth}
  \centering
  \includegraphics[width=1\columnwidth, trim={6cm 0cm 0cm 3cm}, clip]{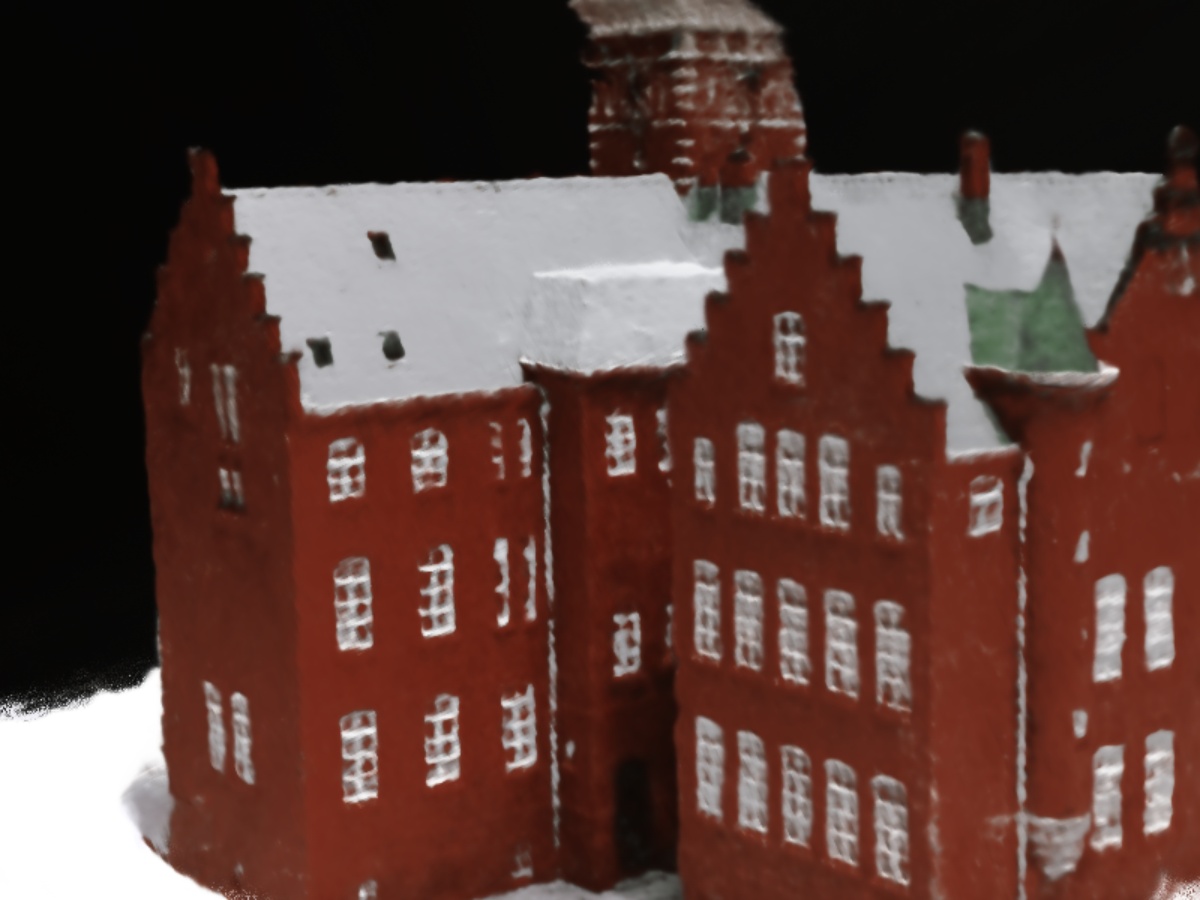}
  \caption*{PSNR: $21.60$}
  \caption*{10K}
\end{subfigure}
\begin{subfigure}{0.6\columnwidth}
  \centering
  \includegraphics[width=1\columnwidth, trim={6cm 0cm 0cm 3cm}, clip]{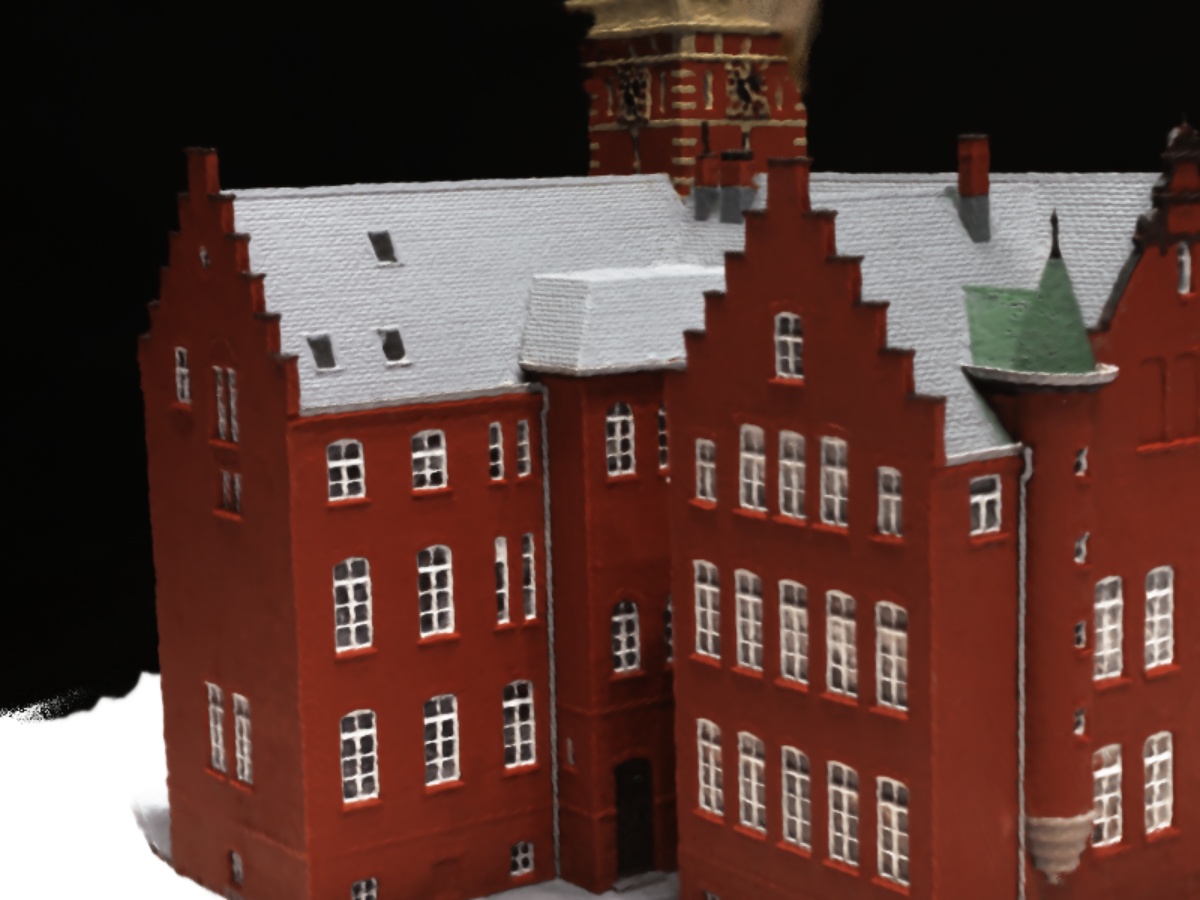}
  \caption*{PSNR: $24.20$}
  \caption*{50K}
\end{subfigure}
\begin{subfigure}{0.6\columnwidth}
  \centering
  \includegraphics[width=1\columnwidth, trim={6cm 0cm 0cm 3cm}, clip]{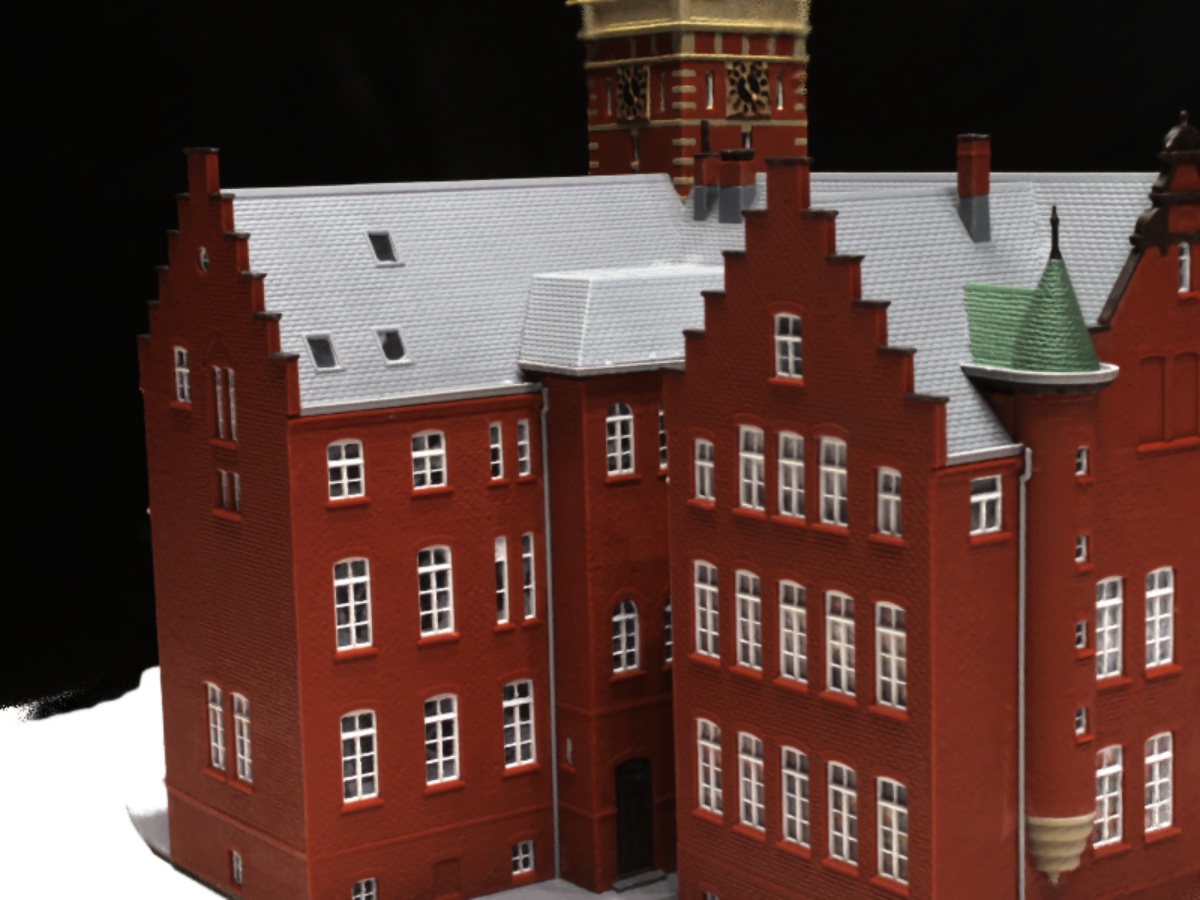}
  \caption*{PSNR: $26.09$}
  \caption*{300K}
\end{subfigure}

\caption{Novel view synthesis on DTU scan-24. The top row is the results of NeuS while the bottom row is the results of NeuS+Ours.}
\label{fig:vis_psnr_24}
\end{figure*}

%% file: tab_fig/dtu_normal.tex
\setlength{\tabcolsep}{4.3pt}
\begin{table*}[h]
\centering


\begin{tabular}{c | c c c c c c c c c c c c c c c c}
\toprule
Method & 24 & 37 & 40 & 55 & 63 & 65 & 69 & 83 & 97 & 105 & 106 & 110 & 114 & 118 & 122 & Mean \\
\midrule
NeuS & $0.895$ & $0.887$ & $0.928$ & $0.955$ & $0.935$ & $0.947$ & $0.819$ & $0.822$ & $0.846$ & $0.816$ & $0.949$ & $0.886$ & $0.942$ & $0.947$ & $0.918$ & $0.899$ \\
NeuralWarp & $0.917$ & $0.908$ & $0.964$ & $0.959$ & $0.931$ & $0.944$ & $0.898$ & $0.826$ & $0.840$ & $0.825$ & $0.947$ & $0.896$ & $0.942$ & $0.942$ & $0.940$ & $0.912$ \\
\midrule
NeuS+Ours & $0.942$ & $0.895$ & $0.968$ & $0.957$ & $0.936$ & $0.953$ & $0.906$ & $0.817$ & $0.856$ & $0.817$ & $0.956$ & $0.897$ & $0.943$ & $0.951$ & $0.927$ & $0.915$ \\ 
NeuralWarp+Ours & $0.920$ & $0.909$ & $0.968$ & $0.955$ & $0.947$ & $0.944$ & $0.910$ & $0.829$ & $0.852$ & $0.822$ & $0.953$ & $0.901$ & $0.940$ & $0.943$ & $0.938$ & $0.915$ \\ 
\bottomrule
\end{tabular}

\caption{Quantitative results for the normal consistency metric on the DTU dataset.}
\label{tab:dtu_normal}
\end{table*}

%% file: tab_fig/dtu_psnr.tex
\setlength{\tabcolsep}{5pt}
\begin{table*}[!h]
\centering


\begin{tabular}{c | c c c c c c c c c c c c c c c c}
\toprule
Method & 24 & 37 & 40 & 55 & 63 & 65 & 69 & 83 & 97 & 105 & 106 & 110 & 114 & 118 & 122 & Mean \\
\midrule
NeuS & $24.54$ & $24.67$ & $25.04$ & $26.83$ & $27.97$ & $29.34$ & $17.90$ & $27.2$ & $25.22$ & $16.3$ & $31.71$ & $31.44$ & $28.86$ & $33.14$ & $21.60$ & $26.12$ \\

NeuS+Ours & $26.00$ & $24.98$ & $26.76$ & $26.94$ & $27.06$ & $30.08$ & $28.44$ & $27.1$ & $25.53$ & $27.68$ & $32.37$ & $31.81$ & $29.73$ & $33.70$ & $34.45$ & $28.84$ \\ 
\bottomrule
\end{tabular}

\vspace{-2mm}
\caption{PSNR results for the novel view synthesis on the test images of DTU dataset.}
\label{tab:dtu_psnr}
\end{table*}

%% file: tab_fig/dtu_psnr_24.tex
\setlength{\tabcolsep}{6pt}
\begin{table*}[!h]
\centering


\begin{tabular}{c | c c c c c c c c}
\toprule
Method & 5K & 10K & 20K & 30K & 50K & 100K & 200K & 300K \\
\midrule
NeuS & $18.22$ & $19.01$ & $20.38$ & $20.88$ & $21.49$ & $22.25$ & $23.84$ & $24.54$  \\

NeuS+Ours & $20.03$ & $21.87$ & $23.04$ & $23.67$ & $23.68$ & $24.84$ & $25.84$ & $26.00$ \\ 
\bottomrule
\end{tabular}

\caption{PSNR results for the novel view synthesis with respect to the iteration number on 7 test images of DTU scan-24.}
\label{tab:dtu_psnr_24}
\end{table*}

%% file: tab_fig/vis_ngp.tex
\begin{figure*}[!h]
\centering

\begin{subfigure}{0.04\columnwidth}
    \small{\rotatebox{90}{~~~~~~~~~~Instant-NGP~~~Image}}
\end{subfigure}
\begin{subfigure}{0.50\columnwidth}
  \centering
  \includegraphics[width=1\columnwidth, trim={6cm 0cm 0cm 3cm}, clip]{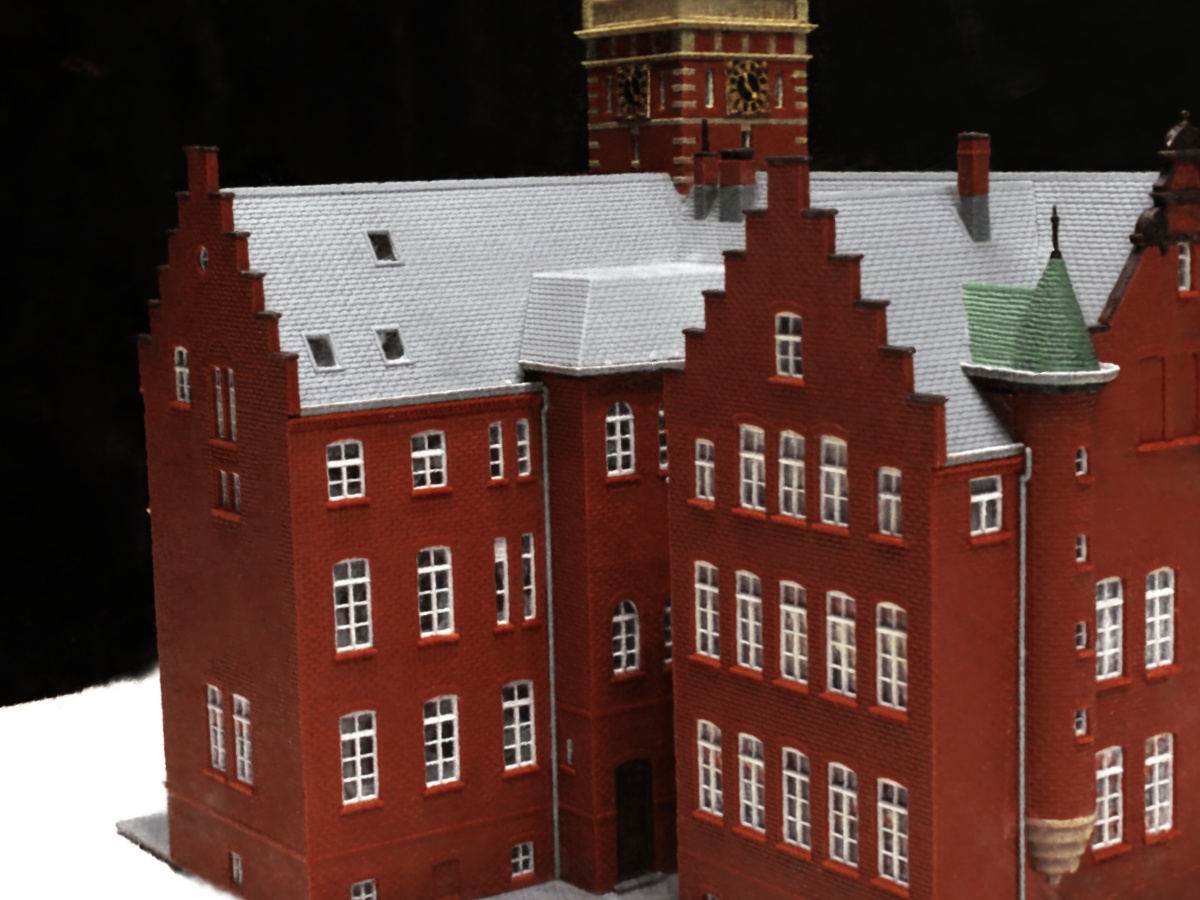}
  \caption*{PSNR: $25.89$}
\end{subfigure}
\begin{subfigure}{0.50\columnwidth}
  \centering
  \includegraphics[width=1\columnwidth, trim={6cm 0cm 0cm 3cm}, clip]{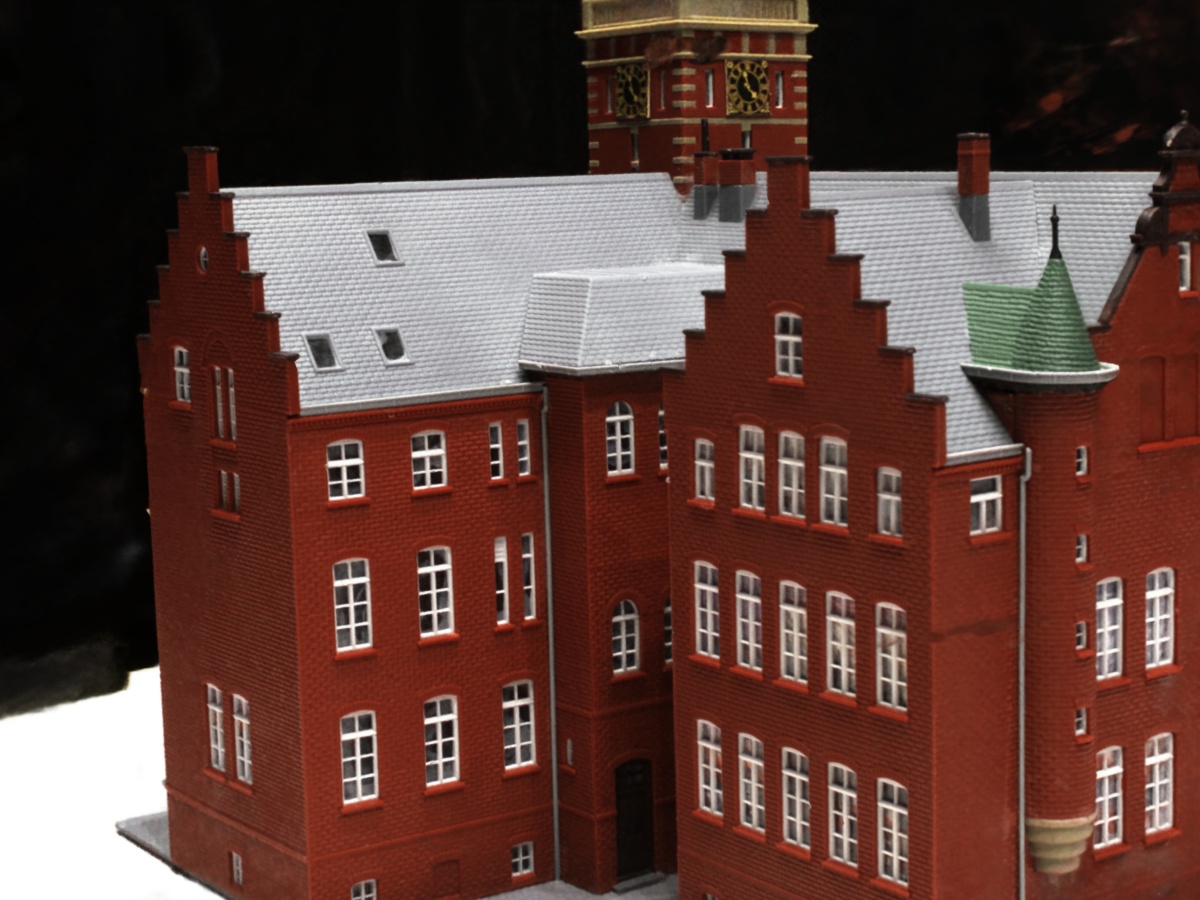}
  \caption*{PSNR: $26.07$}
\end{subfigure}
\begin{subfigure}{0.50\columnwidth}
  \centering
  \includegraphics[width=1\columnwidth, trim={6cm 0cm 0cm 3cm}, clip]{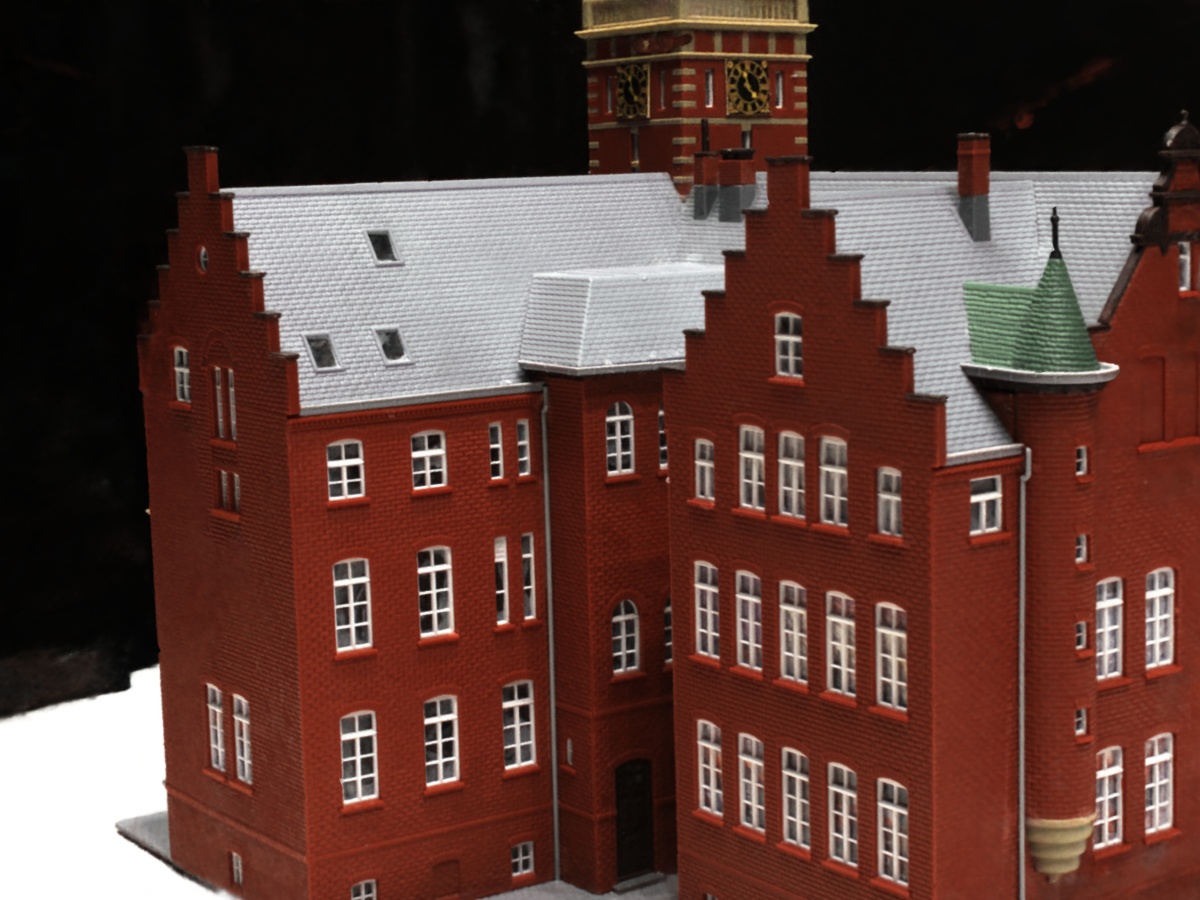}
  \caption*{PSNR: $27.01$}
\end{subfigure}

\begin{subfigure}{0.04\columnwidth}
    \small{\rotatebox{90}{~~~Instant-NGP~~~Surface}}
\end{subfigure}
\begin{subfigure}{0.50\columnwidth}
  \centering
  \includegraphics[width=1\columnwidth, trim={6cm 0cm 0cm 3cm}, clip]{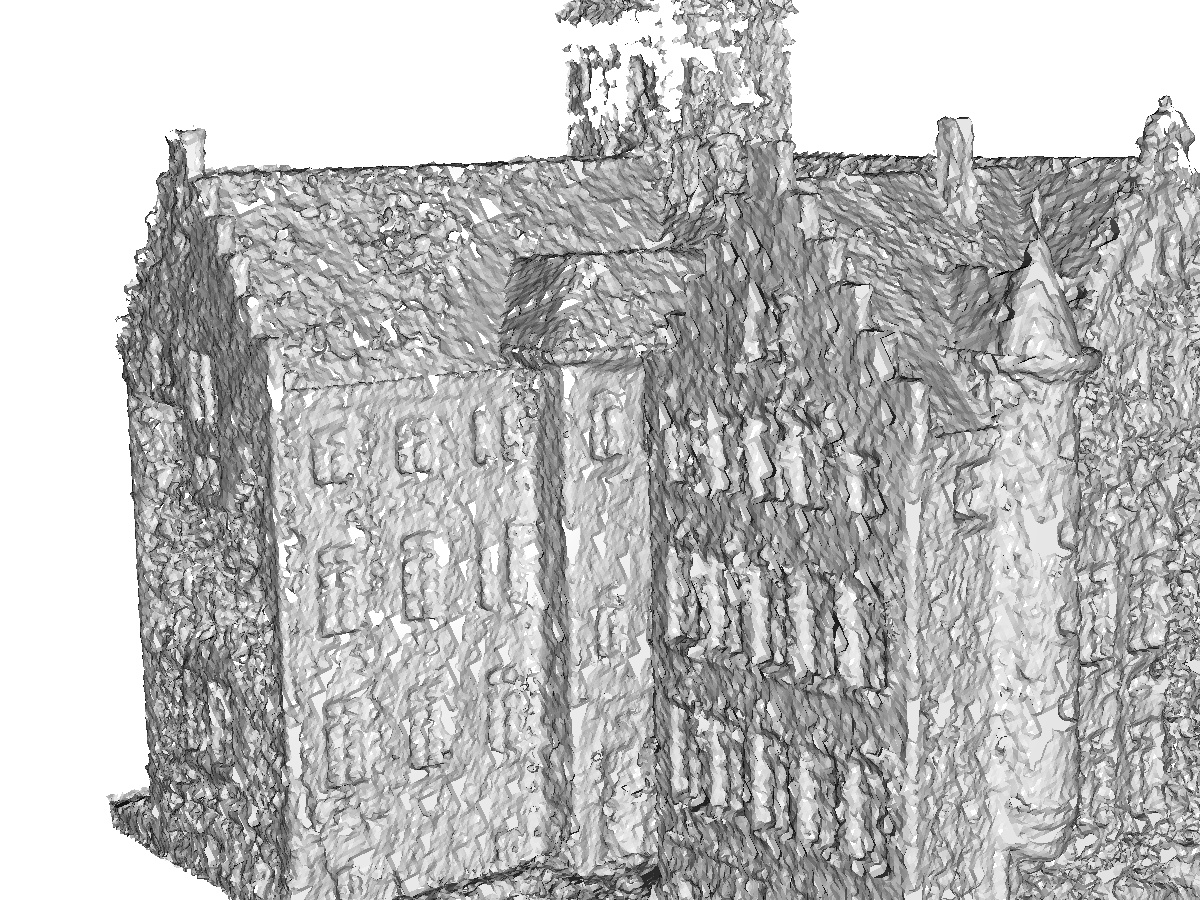}
\end{subfigure}
\begin{subfigure}{0.50\columnwidth}
  \centering
  \includegraphics[width=1\columnwidth, trim={6cm 0cm 0cm 3cm}, clip]{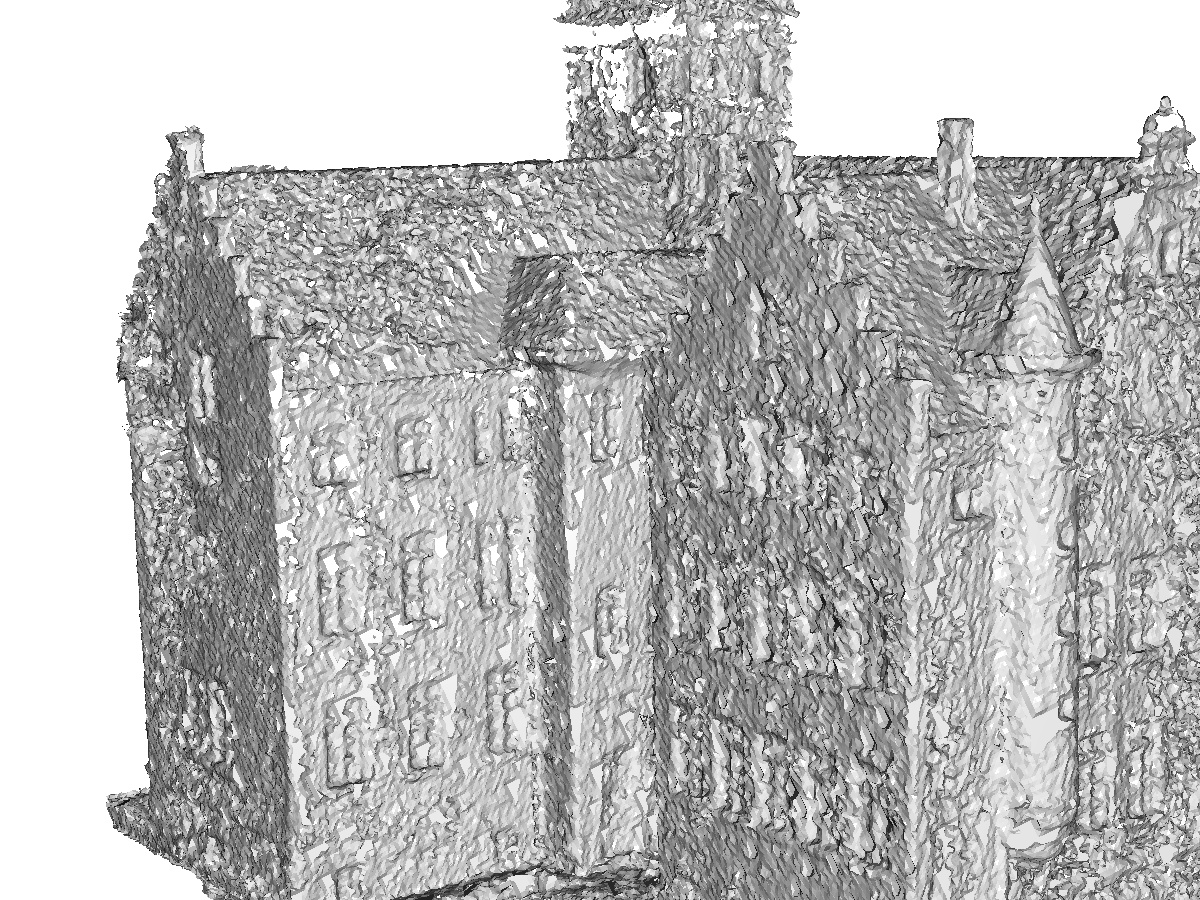}
\end{subfigure}
\begin{subfigure}{0.50\columnwidth}
  \centering
  \includegraphics[width=1\columnwidth, trim={6cm 0cm 0cm 3cm}, clip]{figs/compare_vol/ngp/mesh_30_42.jpg}
\end{subfigure}

\begin{subfigure}{0.04\columnwidth}
    \small{\rotatebox{90}{~~~~~~~NeuS+Hash~Image}}
\end{subfigure}
\begin{subfigure}{0.50\columnwidth}
  \centering
  \includegraphics[width=1\columnwidth, trim={6cm 0cm 0cm 3cm}, clip]{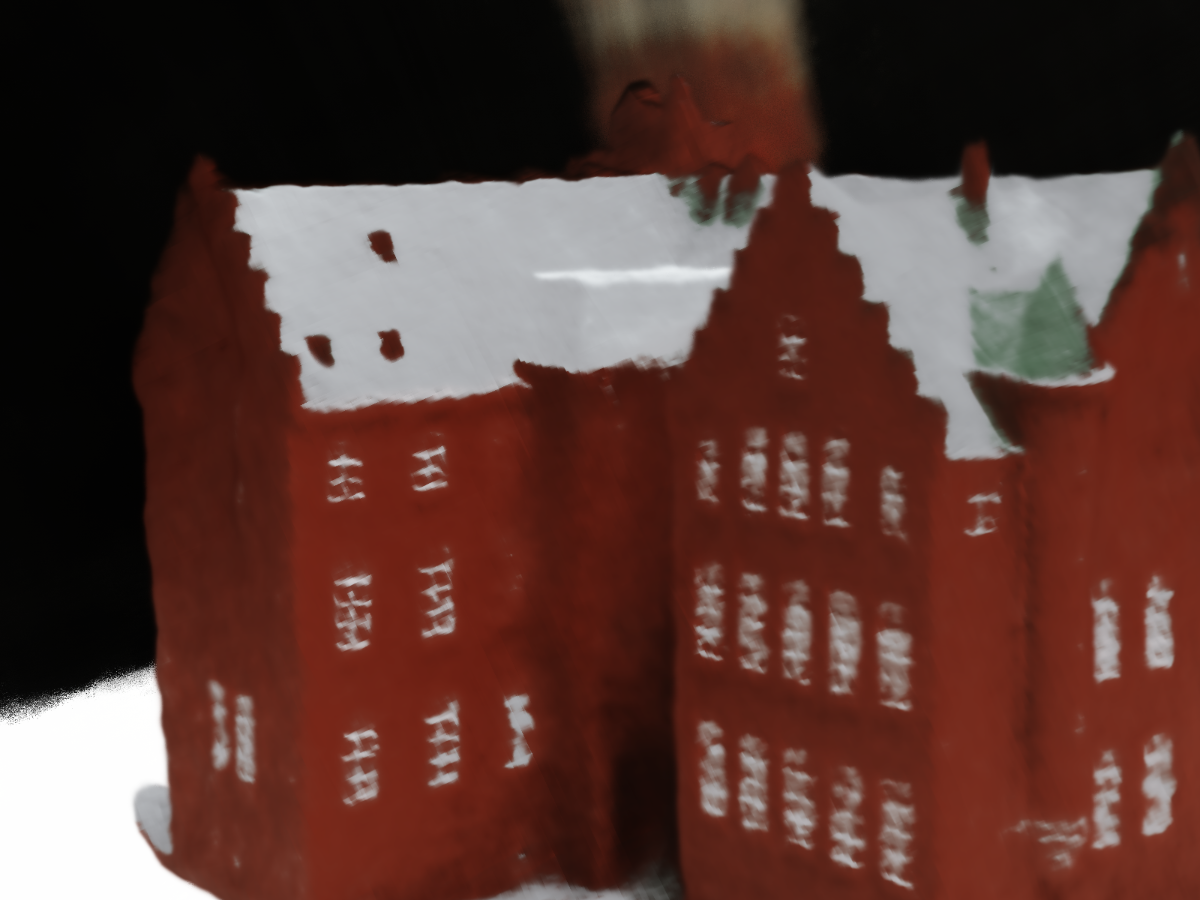}
  \caption*{PSNR: $20.26$}
\end{subfigure}
\begin{subfigure}{0.50\columnwidth}
  \centering
  \includegraphics[width=1\columnwidth, trim={6cm 0cm 0cm 3cm}, clip]{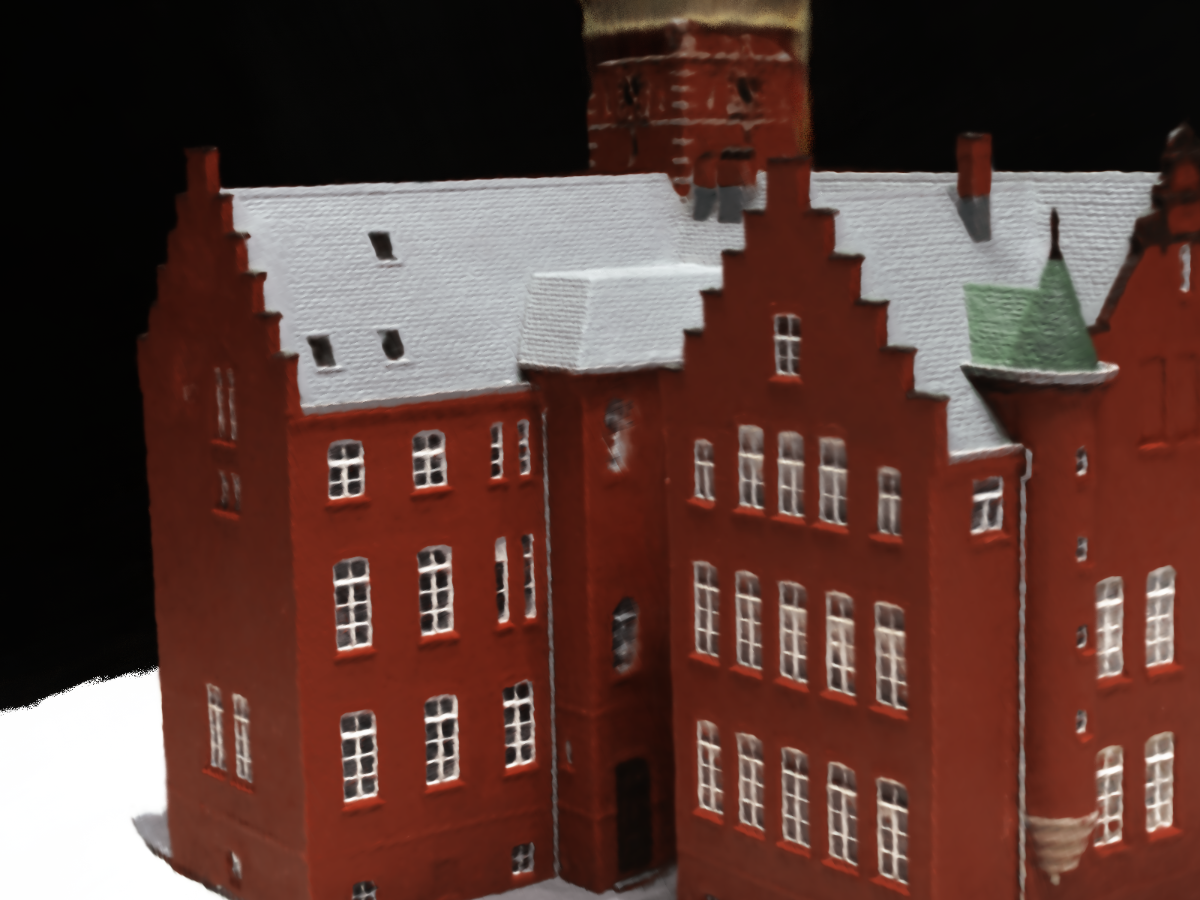}
    \caption*{PSNR: $24.67$}
\end{subfigure}
\begin{subfigure}{0.50\columnwidth}
  \centering
  \includegraphics[width=1\columnwidth, trim={6cm 0cm 0cm 3cm}, clip]{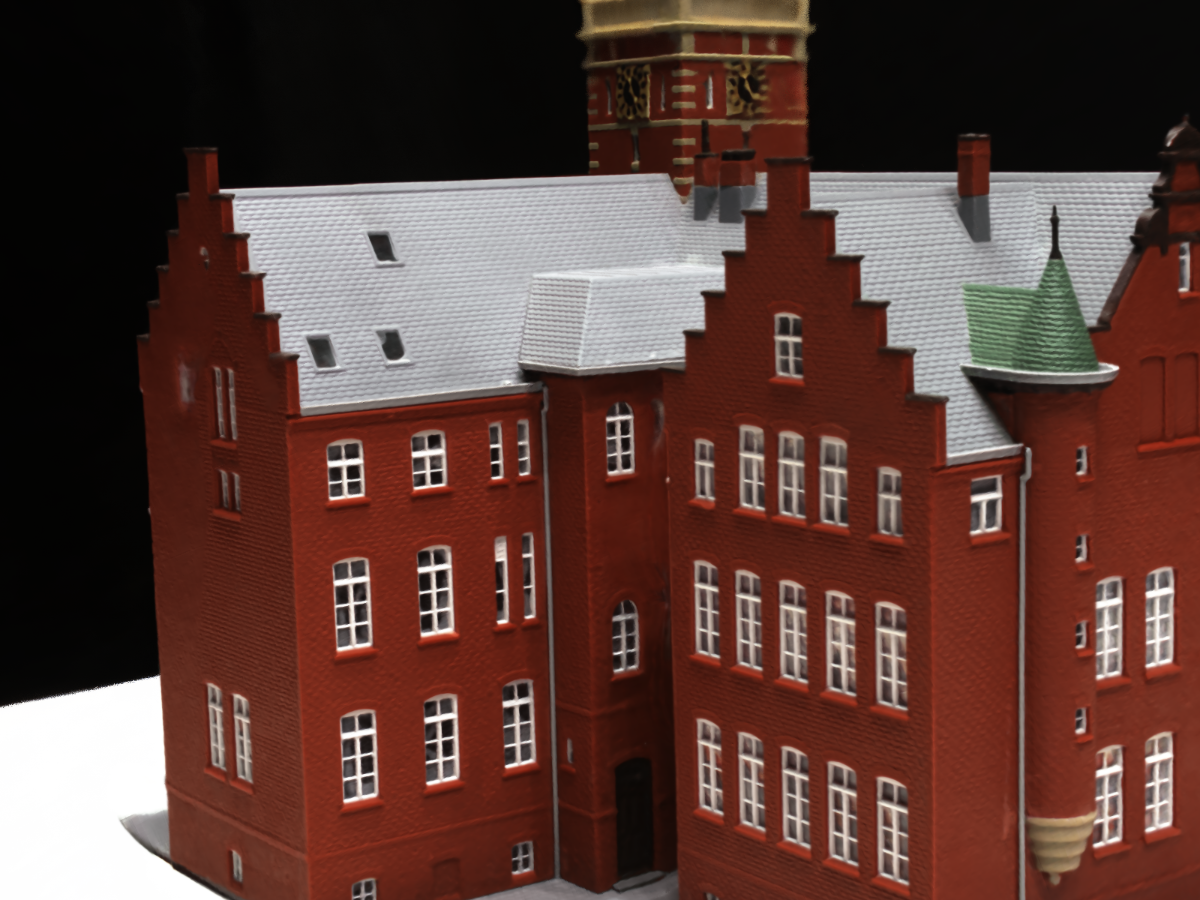}
    \caption*{PSNR: $25.95$}
\end{subfigure}

\begin{subfigure}{0.04\columnwidth}
    \small{\rotatebox{90}{~~~~~NeuS+Hash~Surface}}
\end{subfigure}
\begin{subfigure}{0.50\columnwidth}
  \centering
  \includegraphics[width=1\columnwidth, trim={6cm 0cm 0cm 3cm}, clip]{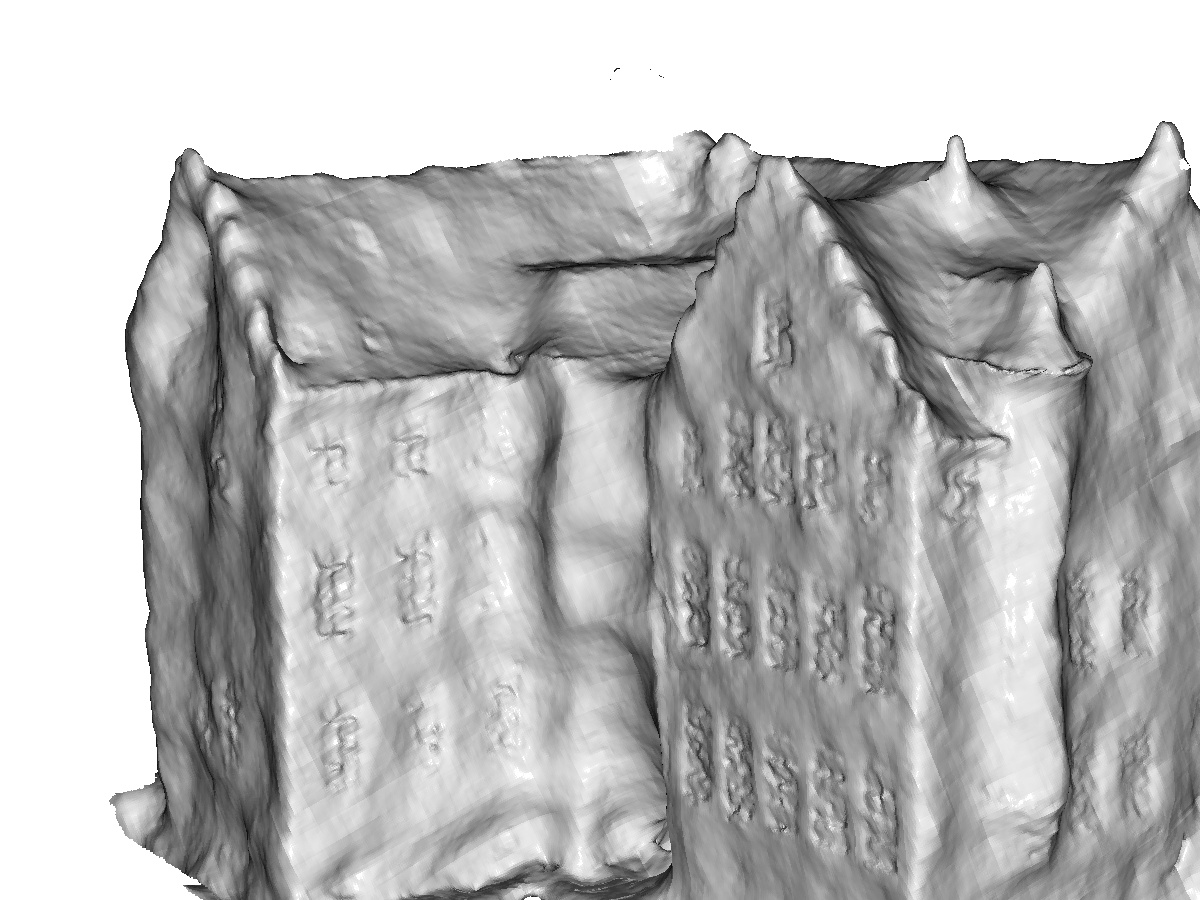}
\end{subfigure}
\begin{subfigure}{0.50\columnwidth}
  \centering
  \includegraphics[width=1\columnwidth, trim={6cm 0cm 0cm 3cm}, clip]{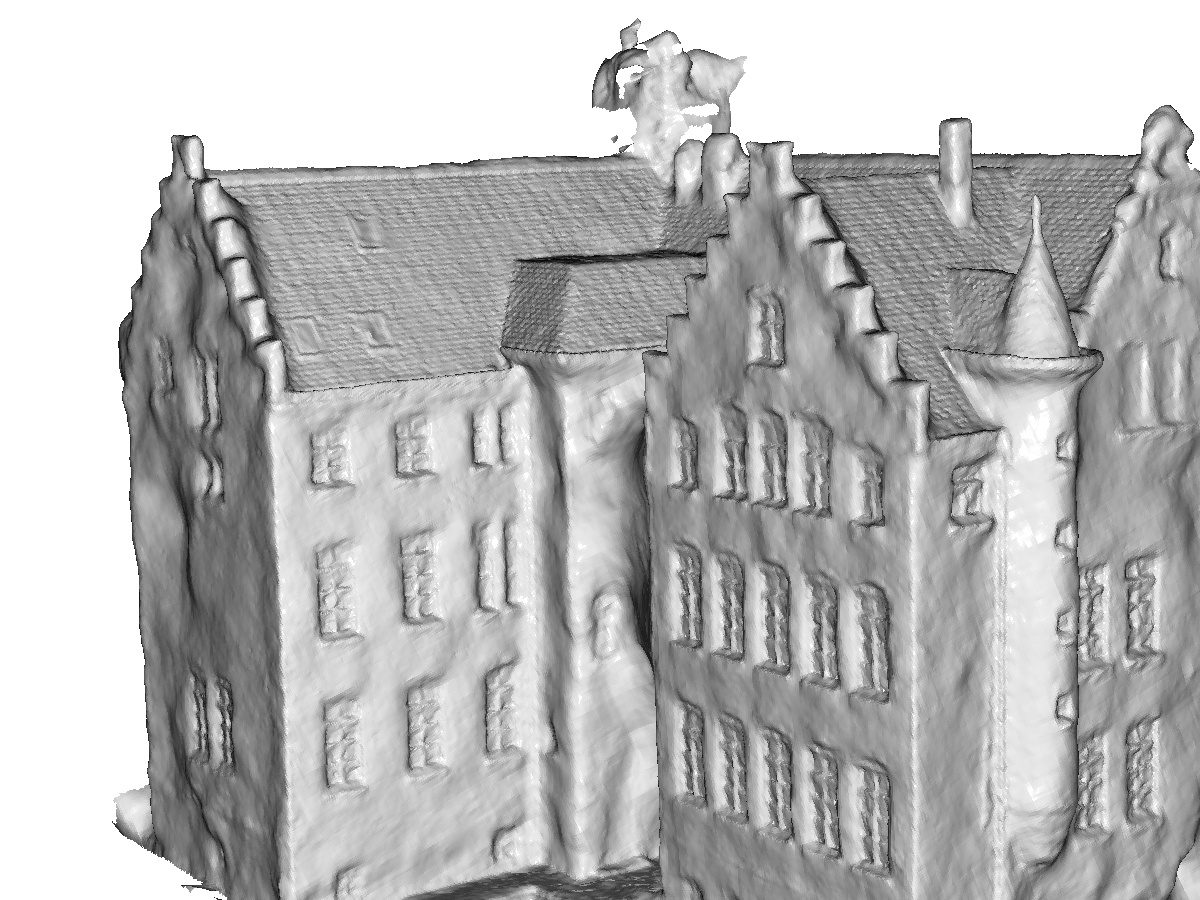}
\end{subfigure}
\begin{subfigure}{0.50\columnwidth}
  \centering
  \includegraphics[width=1\columnwidth, trim={6cm 0cm 0cm 3cm}, clip]{figs/compare_vol/ngp_neus/mesh_30_42.jpg}
\end{subfigure}

\begin{subfigure}{0.04\columnwidth}
    \small{\rotatebox{90}{~~~~~NeuS+Ours~~~Image}}
\end{subfigure}
\begin{subfigure}{0.50\columnwidth}
  \centering
  \includegraphics[width=1\columnwidth, trim={6cm 0cm 0cm 3cm}, clip]{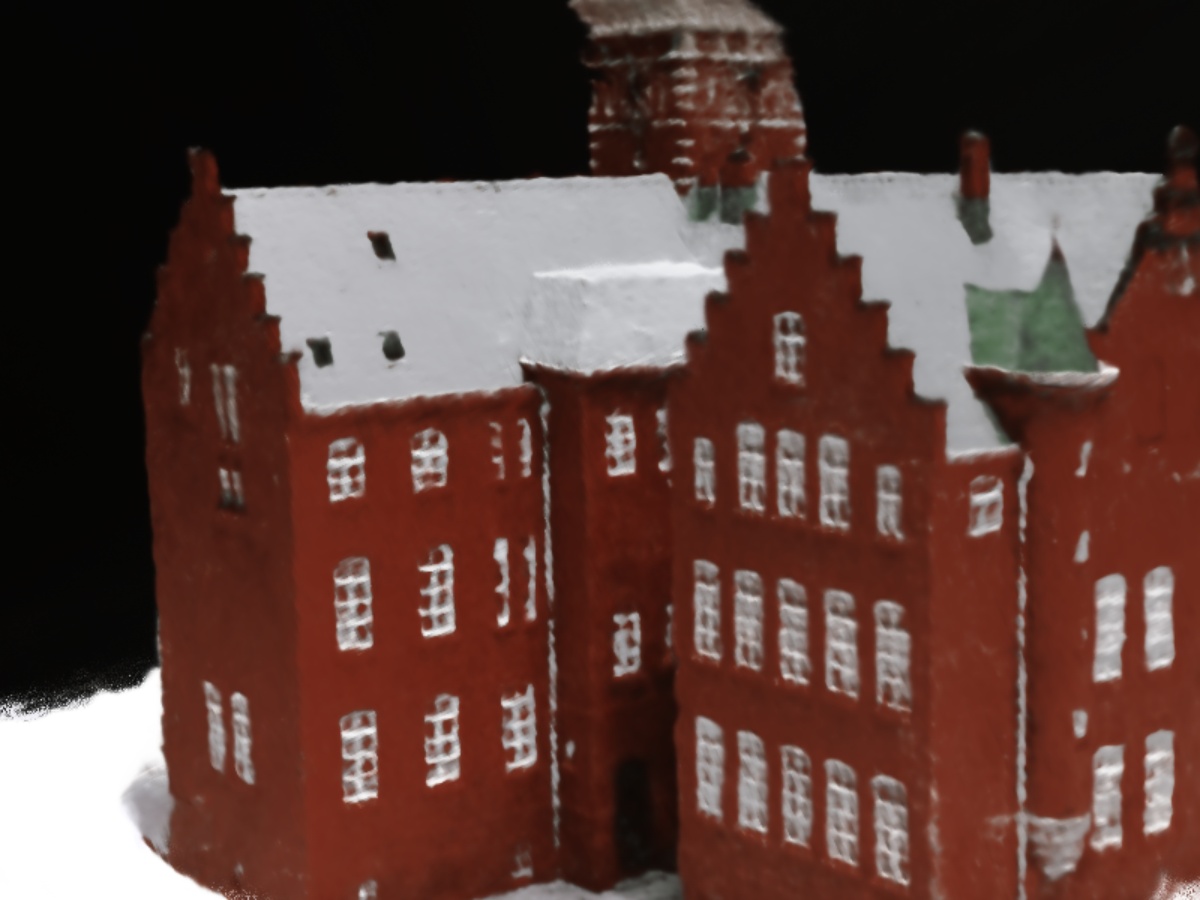}
    \caption*{PSNR: $21.60$}
\end{subfigure}
\begin{subfigure}{0.50\columnwidth}
  \centering
  \includegraphics[width=1\columnwidth, trim={6cm 0cm 0cm 3cm}, clip]{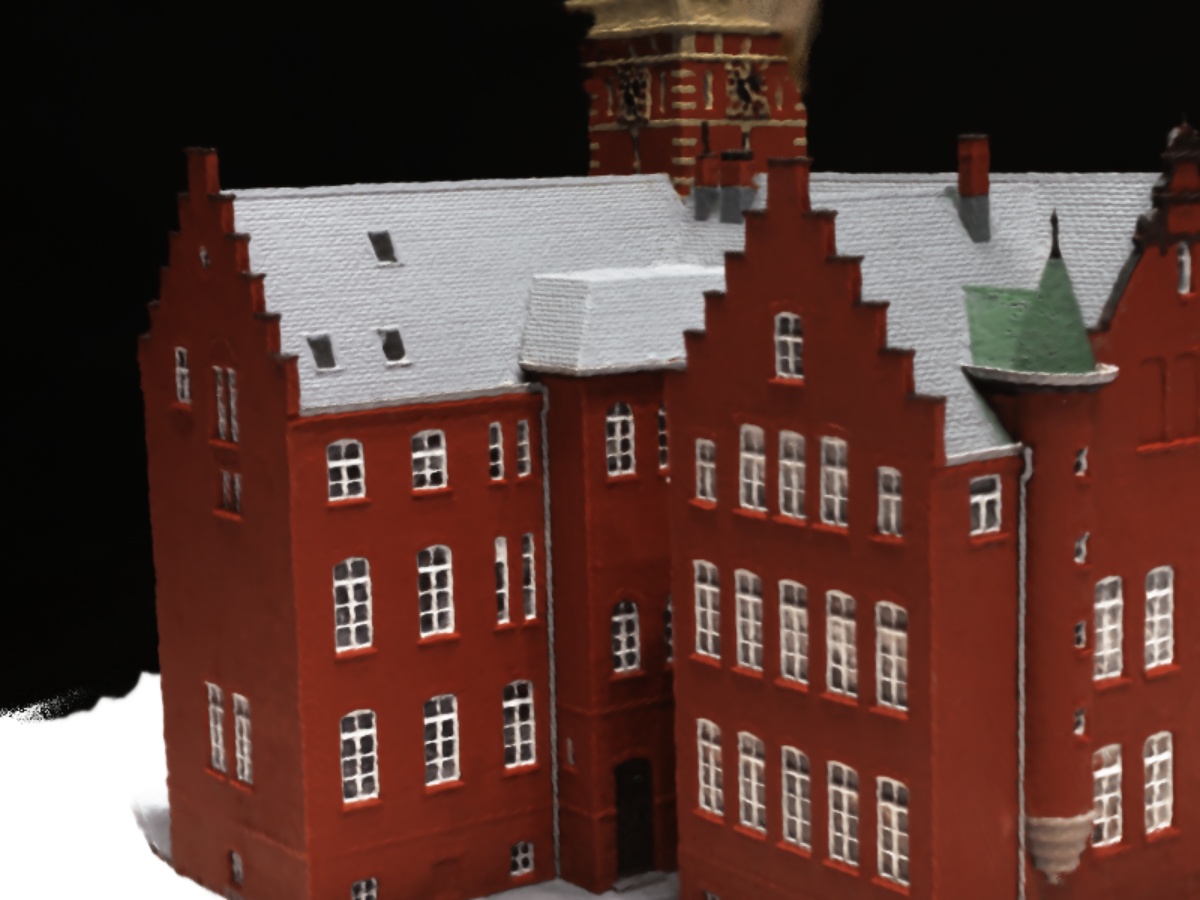}
    \caption*{PSNR: $24.20$}
\end{subfigure}
\begin{subfigure}{0.50\columnwidth}
  \centering
  \includegraphics[width=1\columnwidth, trim={6cm 0cm 0cm 3cm}, clip]{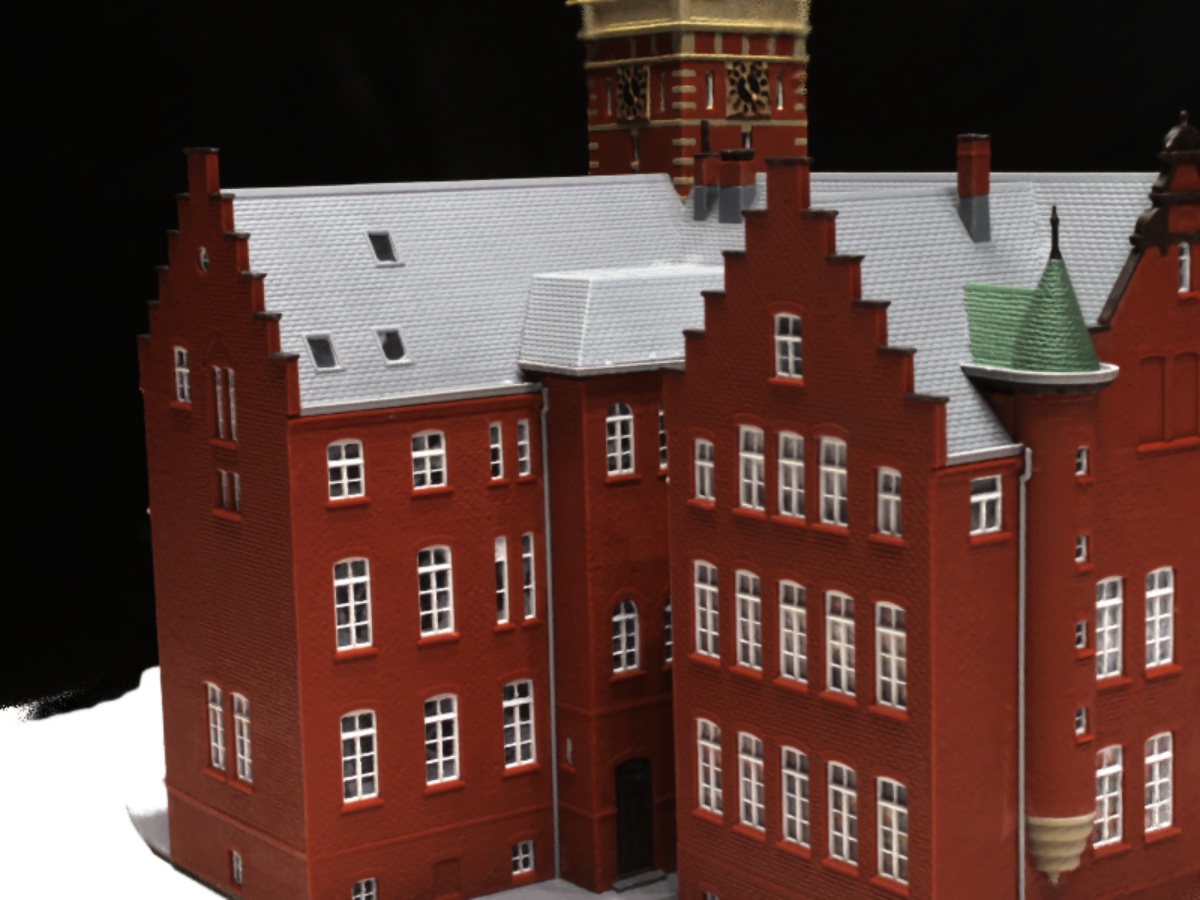}
    \caption*{PSNR: $26.09$}
\end{subfigure}

\begin{subfigure}{0.04\columnwidth}
    \small{\rotatebox{90}{~~~~~NeuS+Ours~~~Surface}}
\end{subfigure}
\begin{subfigure}{0.50\columnwidth}
  \centering
  \includegraphics[width=1\columnwidth, trim={6cm 0cm 0cm 3cm}, clip]{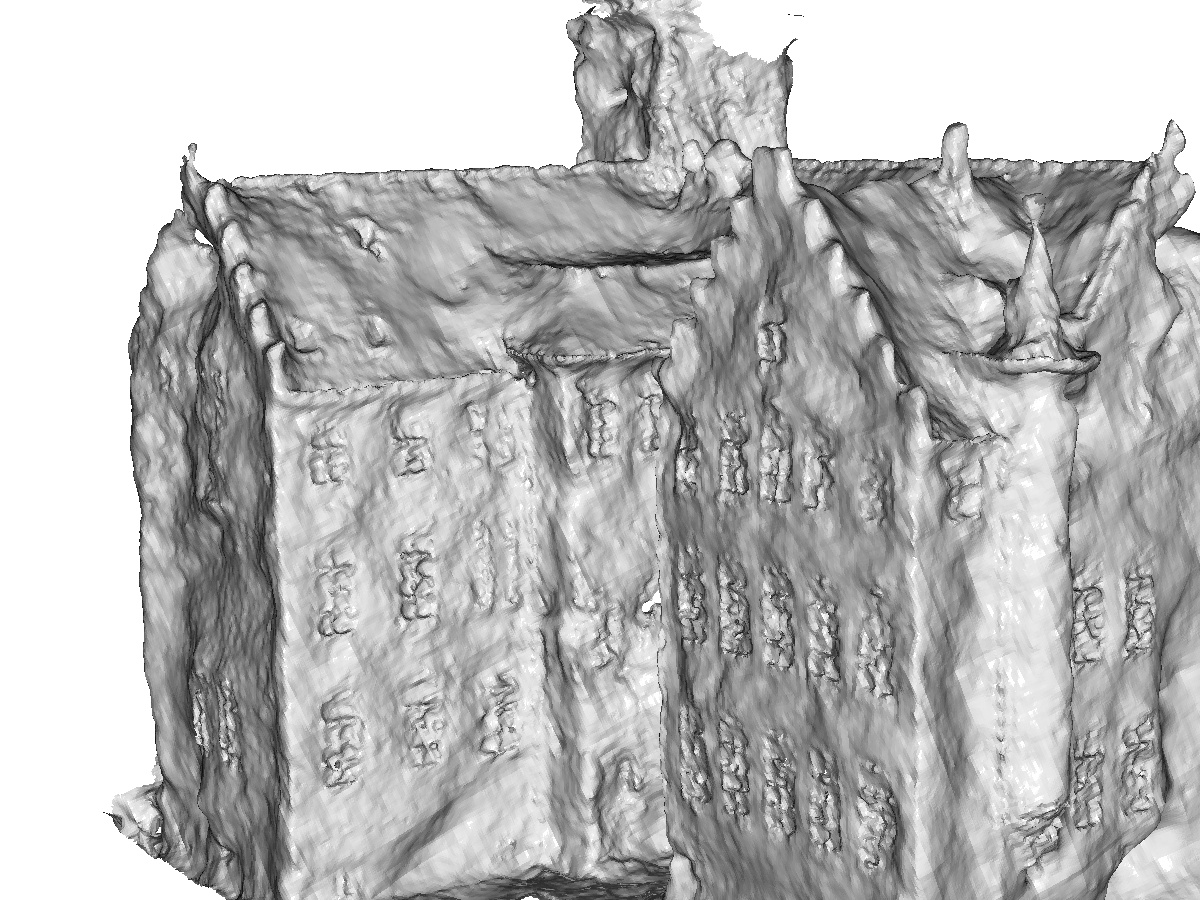}
  \caption*{10K}
\end{subfigure}
\begin{subfigure}{0.50\columnwidth}
  \centering
  \includegraphics[width=1\columnwidth, trim={6cm 0cm 0cm 3cm}, clip]{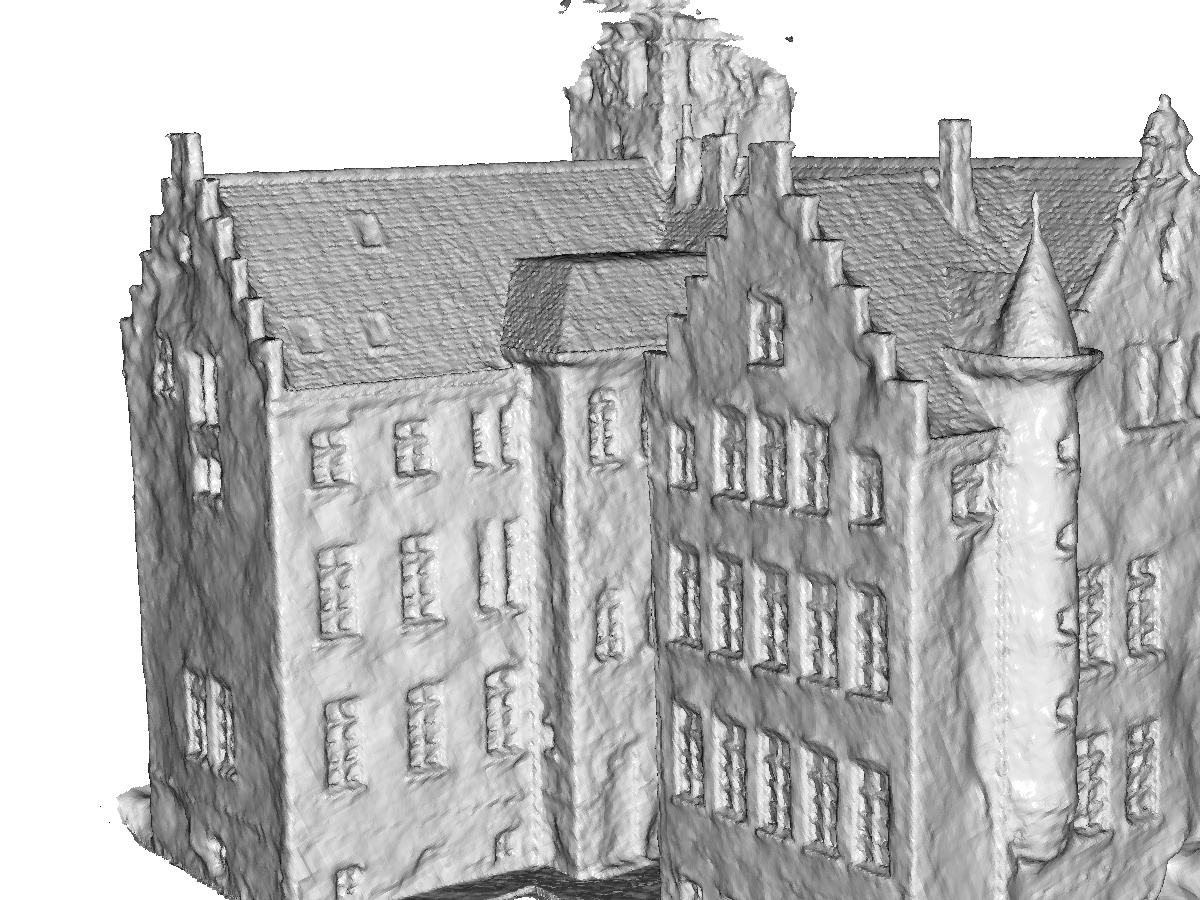}
  \caption*{50K}
\end{subfigure}
\begin{subfigure}{0.50\columnwidth}
  \centering
  \includegraphics[width=1\columnwidth, trim={6cm 0cm 0cm 3cm}, clip]{figs/compare_vol/ours/mesh_30_42.jpg}
  \caption*{300K}
\end{subfigure}

 \caption{Visual comparison to Instant-NGP on DTU scan-24. The rendered novel view and the extracted mesh are displayed with respect to the iteration number.}
\label{fig:vis_ngp}
\end{figure*}

%% file: tab_fig/vis_encoding.tex
\begin{figure*}[!h]
\centering

\begin{subfigure}{0.49\columnwidth}
  \centering
  \includegraphics[width=1\columnwidth, trim={0cm 0cm 0cm 0.1cm}, clip]{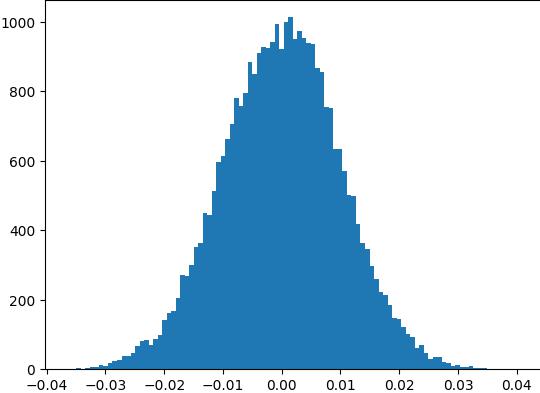}
  \caption*{\#Iter = 0, Resolution = $32^3$}
\end{subfigure}
\begin{subfigure}{0.49\columnwidth}
  \centering
  \includegraphics[width=1\columnwidth, trim={0cm 0cm 0cm 0.1cm}, clip]{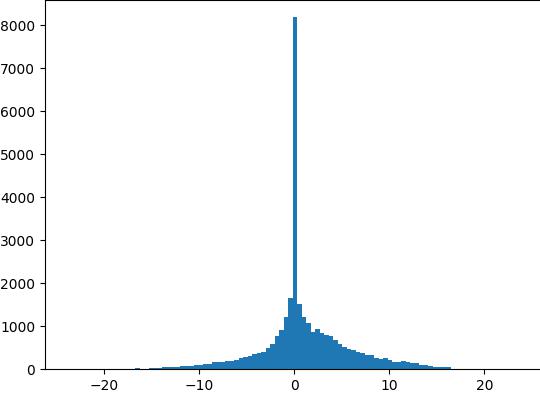}
  \caption*{\#Iter = 300K, Resolution = $32^3$}
\end{subfigure}
\begin{subfigure}{0.49\columnwidth}
  \centering
  \includegraphics[width=1\columnwidth, trim={10cm 0cm 10cm 0cm}, clip]{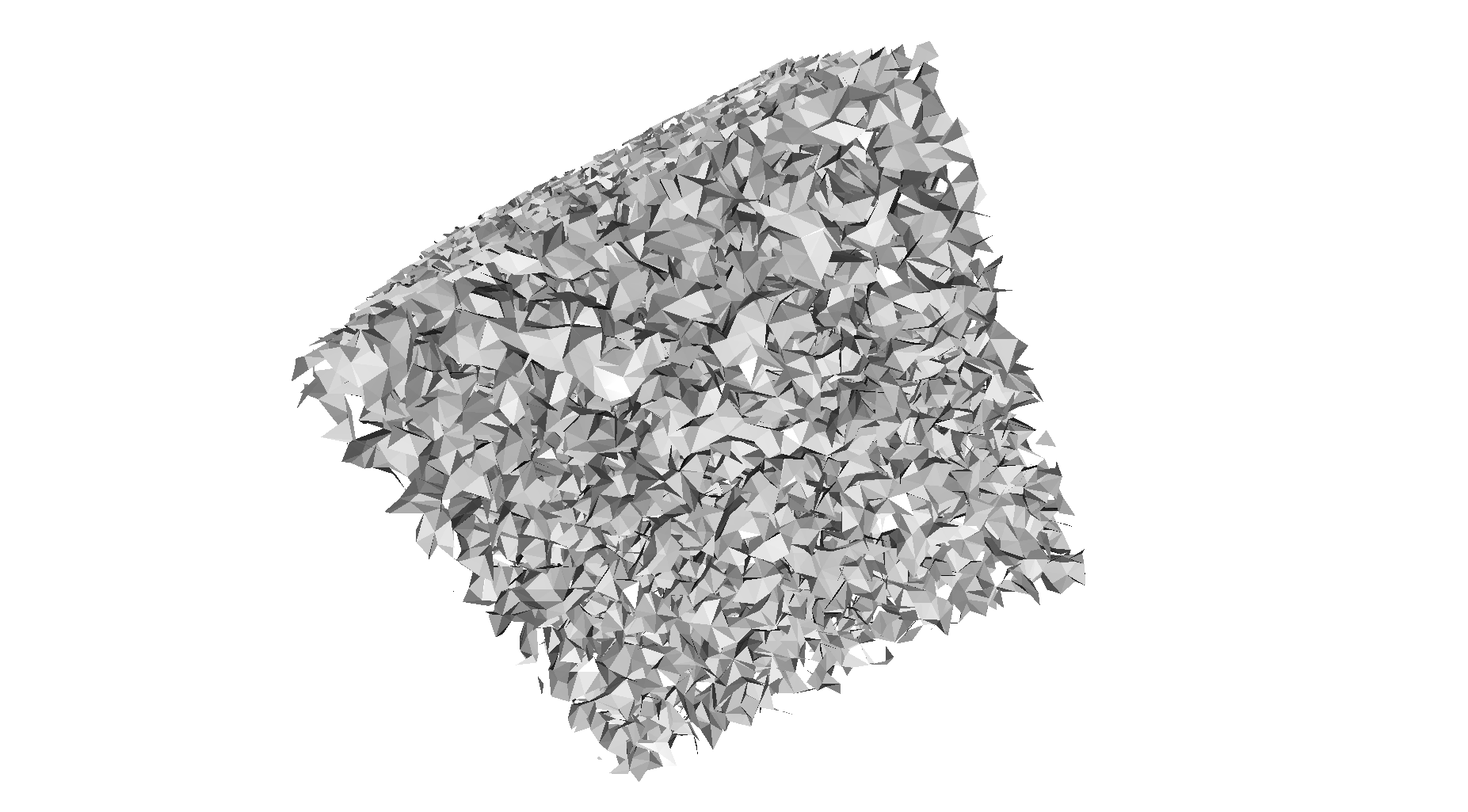}
  \caption*{\#Iter = 0, Resolution = $32^3$}
\end{subfigure}
\begin{subfigure}{0.49\columnwidth}
  \centering
  \includegraphics[width=1\columnwidth, trim={10cm 0cm 10cm 0cm}, clip]{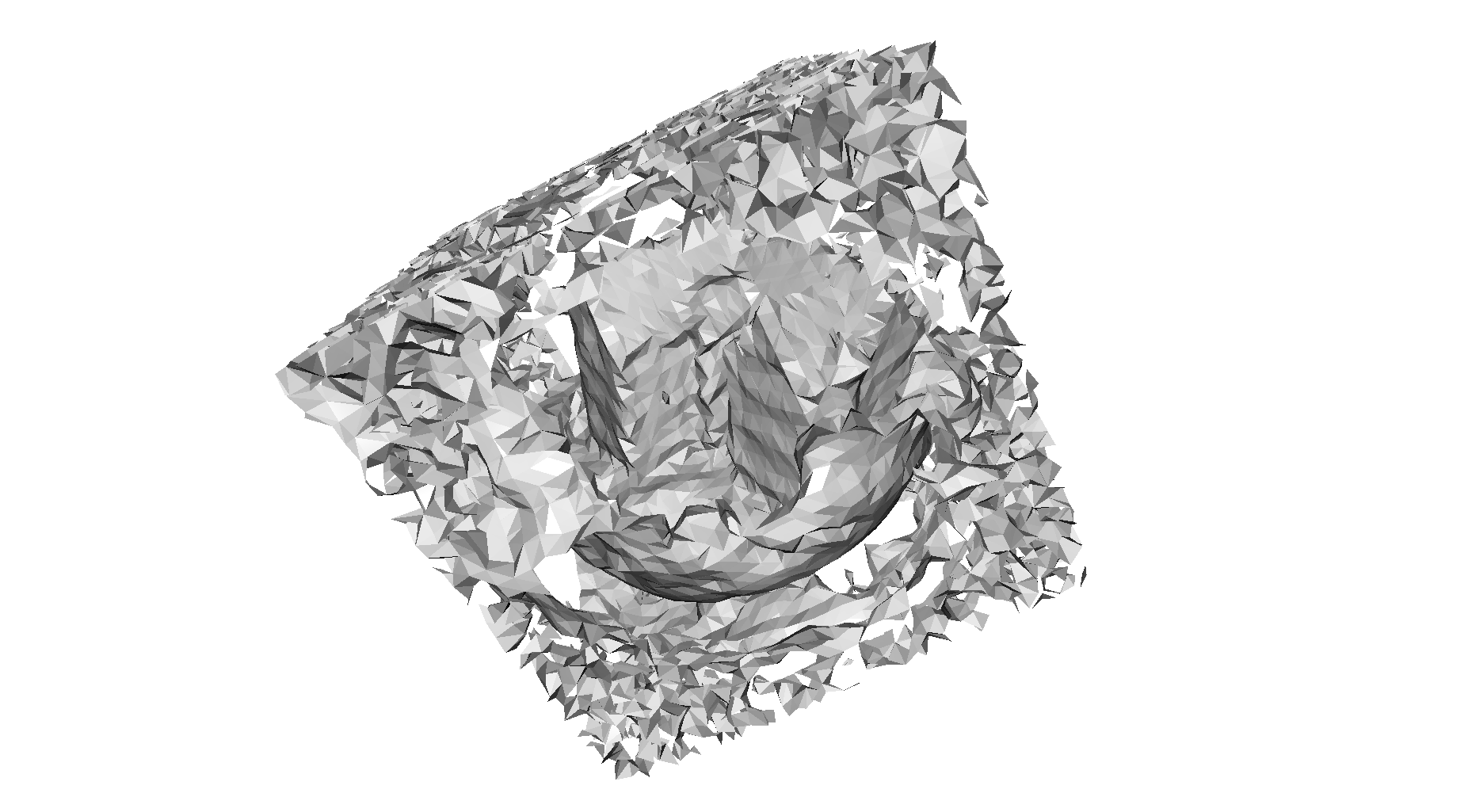}
  \caption*{\#Iter = 300K, Resolution = $32^3$}
\end{subfigure}

\begin{subfigure}{0.6\columnwidth}
  \centering
  \includegraphics[width=1\columnwidth, trim={10cm 0cm 10cm 0cm}, clip]{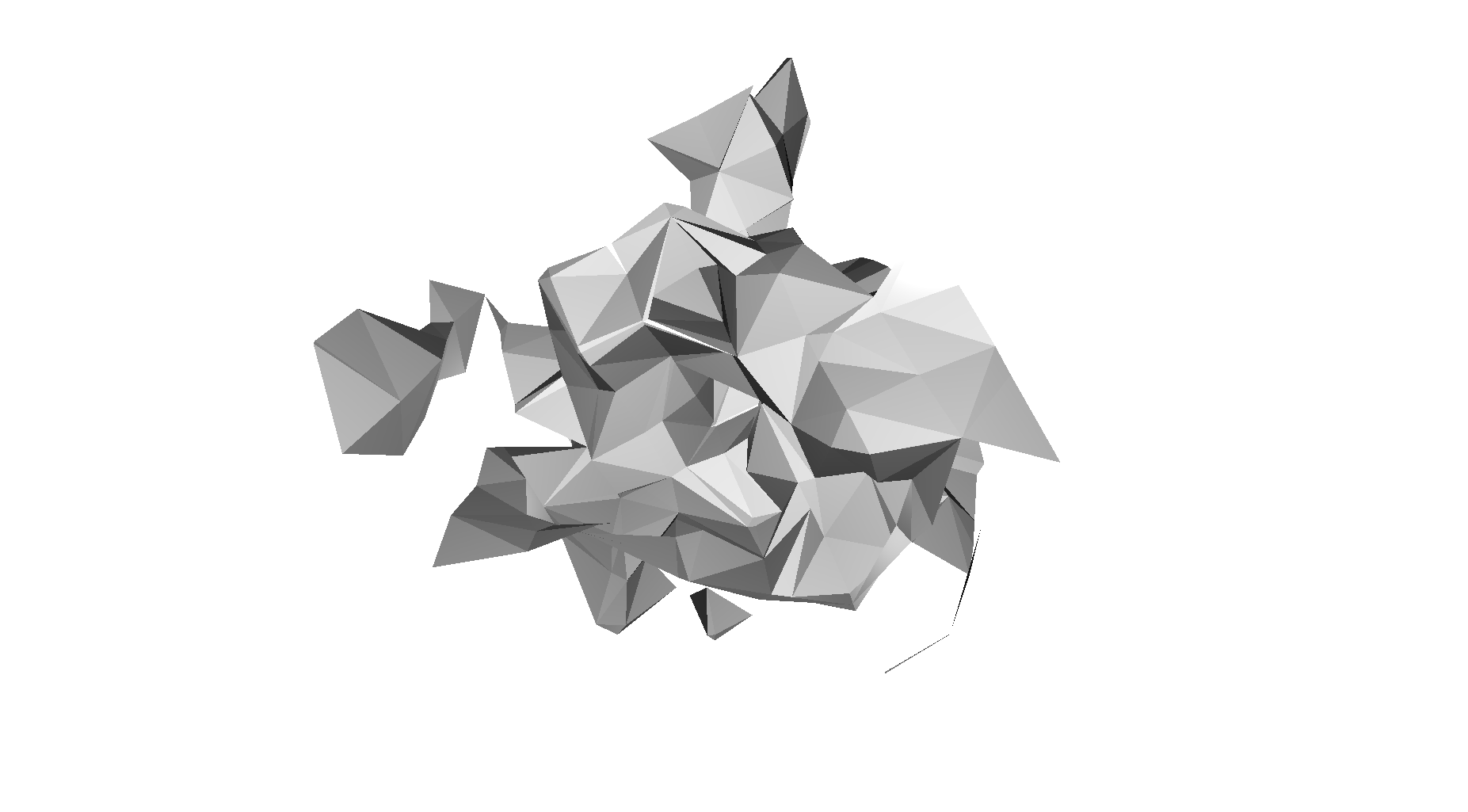}
  \caption*{\#Iter = 300K, Resolution = $8^3$}
\end{subfigure}
\begin{subfigure}{0.6\columnwidth}
  \centering
  \includegraphics[width=1\columnwidth, trim={10cm 0cm 10cm 0cm}, clip]{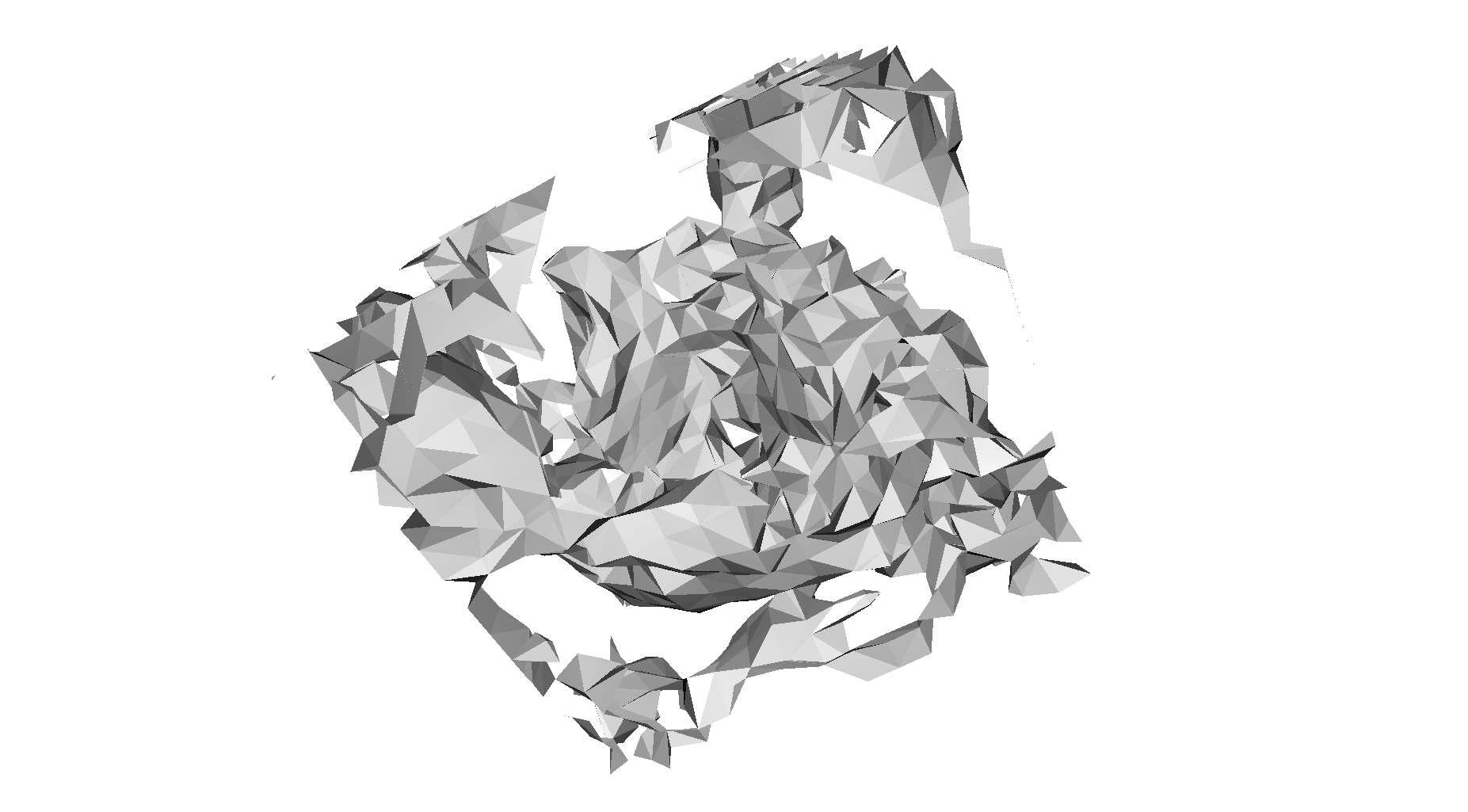}
  \caption*{\#Iter = 300K, Resolution = $16^3$}
\end{subfigure}
%
\begin{subfigure}{0.6\columnwidth}
  \centering
  \includegraphics[width=1\columnwidth, trim={10cm 0cm 10cm 0cm}, clip]{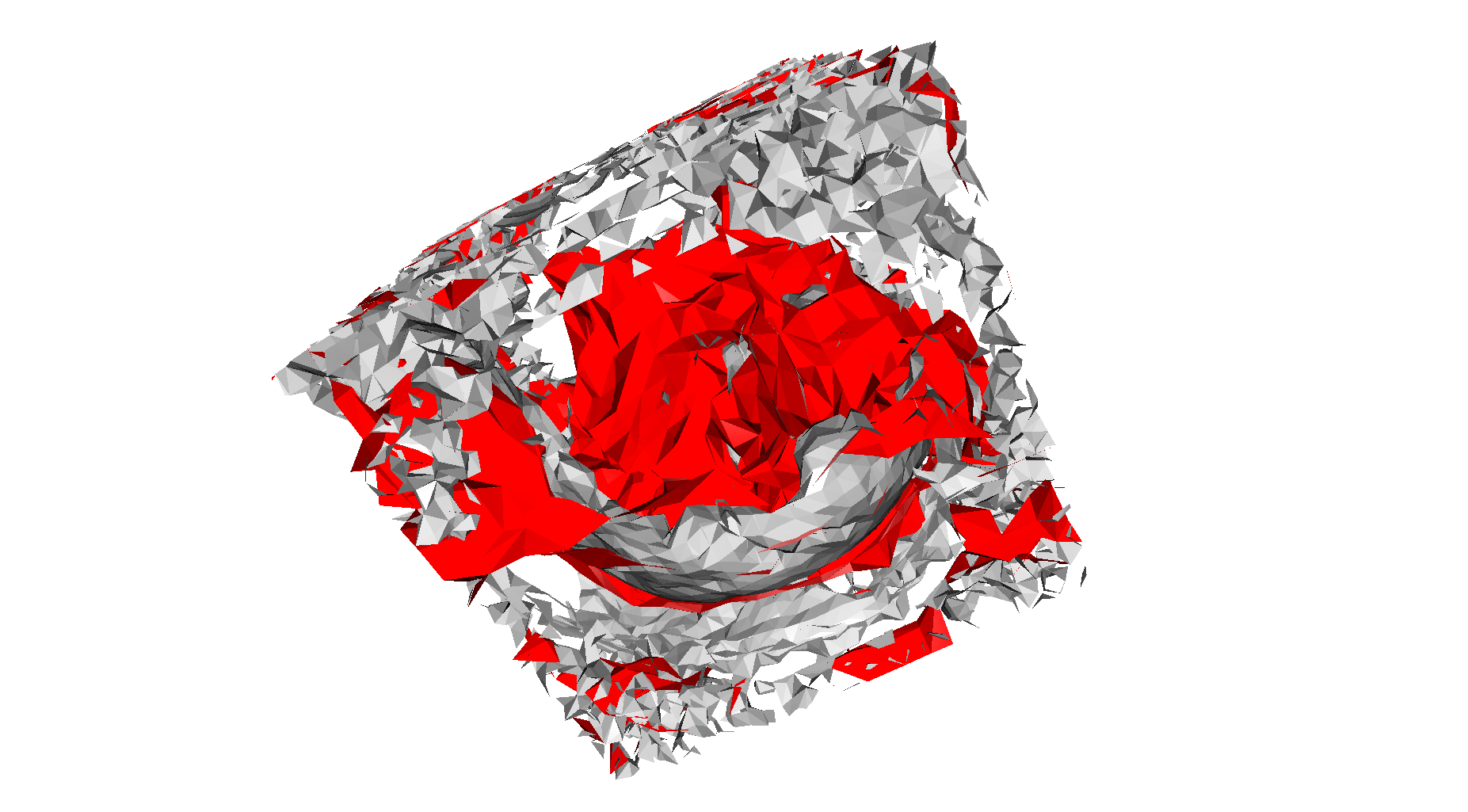}
  \caption*{Blended with Res. $16^{3}$ and $32^{3}$}
\end{subfigure}

\caption{Visualization of the volume encodings. The top row presents the histograms of the values of the voxels in the volume with the resolution of $32^3$, of which the first one is the initialized volume before the training, and the second one is the final volume after the training. The corresponding meshes are displayed in the second row, and more meshes are displayed in the third row.}
\label{fig:vis_encoding}
\vspace{-5mm}
\end{figure*}

%% file: includes_v2/conclusion.tex
\section{CONCLUSION}
We propose to explicitly encode the 3D shape surface by hierarchical volumes to facilitate MLPs in neural implicit surface reconstruction methods. 
The spatially varying features can be obtained from the high-resolution volume to reason more details for each query 3D point, while the feature from low-resolution volumes could reason spatial consistency to keep shapes smooth.
We also design a sparse structure to reduce the memory consumption of high-resolution volumes, and two regularization terms to enhance surface smoothness.
Our hierarchical volume encoding could be appended to any implicit surface reconstruction method as a plug-and-play module to significantly boost their performance
